\pdfoutput=1

\documentclass{article}

\usepackage{arxiv}
\usepackage[utf8]{inputenc} 
\usepackage[T1]{fontenc}    
\usepackage{hyperref}       
\usepackage{url}            
\usepackage{booktabs}       
\usepackage{amsfonts}       
\usepackage{nicefrac}       
\usepackage{microtype}      
\usepackage{xcolor}         
\usepackage{dirtree}
\usepackage{bbding}
\usepackage{multirow}
\usepackage{graphicx}
\usepackage{framed}
\usepackage{tcolorbox}
\usepackage{amssymb}
\usepackage{array,longtable}
\usepackage{caption}
\usepackage{xcolor}
\definecolor{forestgreen}{RGB}{34, 139, 34}
\usepackage{caption}
\usepackage{subcaption}
\usepackage{lscape}

\def \thin {.18\textwidth}
\def \cbox {$\bigcirc$}
\newcommand{\st}[1]{
    \cbox{} & \cbox{} & \cbox{} & \cbox{} & \cbox{}
}

\usepackage{url}
\usepackage{rotating}
\usepackage{booktabs}
\usepackage{amsmath,bm}
\usepackage{bbm}
\usepackage{cite}
\usepackage{enumerate}
\usepackage{algorithm}
\usepackage{algorithmic}
\usepackage{colortbl}
\usepackage{psfrag}
\usepackage{adjustbox}
\usepackage{wrapfig}
\usepackage{xspace}
\usepackage{amssymb}
\usepackage{multirow}
\usepackage{afterpage}
\usepackage{float}
\usepackage{cleveref}
\usepackage[toc,page,header]{appendix}
\usepackage{minitoc}

\newcommand{\answerTODO}[1][]{\textcolor{red}{\bf [TODO]}}

\usepackage{xcolor,colortbl}

\definecolor{Gray}{gray}{0.85}
\definecolor{LightCyan}{rgb}{0.88,1,1}

\newcolumntype{a}{>{\columncolor{Gray}}c}

\usepackage{svg}
\usepackage{xcolor}
\colorlet{darkgreen}{green!65!black}
\colorlet{darkblue}{blue!75!black}
\colorlet{darkred}{red!80!black}
\definecolor{lightblue}{HTML}{0071bc}
\definecolor{lightgreen}{HTML}{39b54a}
\definecolor{manyshot}{HTML}{6969ff}
\definecolor{medshot}{HTML}{f7c600}
\definecolor{fewshot}{HTML}{ff6969}
\definecolor{mypurple}{HTML}{412F8A}
\definecolor{myorange}{HTML}{fc8e62}
\definecolor{deemph}{gray}{0.55}

\definecolor{linkcolor}{HTML}{ED1C24}

\usepackage[colorinlistoftodos]{todonotes}

\renewcommand{\paragraph}[1]{\vspace{1.25mm}\noindent\textbf{#1}}

\definecolor{baselinecolor}{gray}{.95}

\renewcommand \thepart{}
\renewcommand \partname{}

\newcommand\cmt[1]{}  
\newcommand\TODO[1]{}  

\newcommand{\numcasestudies}{{857}\xspace}
\newcommand{\basemodel}{{Gemini Ultra 1.0}\xspace}
\newcommand{\ourmodel}{{PH-LLM}\xspace}
\newcommand{\ourmodellong}{{Personal Health Large Language Model}\xspace}

\newcommand{\numsleepmcq}{629\xspace}

\newcommand{\numfitnessmcq}{99\xspace}

\DeclareCaptionLabelSeparator{pipe}{ $|$ }

\title{Towards a Personal Health Large Language Model}

\usepackage[]{authblk}
\author[*]{Justin Cosentino}
\author[*$\S$]{Anastasiya Belyaeva}
\author[*$\S$]{Xin Liu}
\author[*]{Nicholas A. Furlotte}
\author[$\ddagger$]{Zhun Yang}
\author[$\ddagger$]{Chace Lee}
\author[$\ddagger$]{Erik Schenck}
\author[$\ddagger$]{Yojan Patel}
\author[$\ddagger$]{Jian Cui}
\author[$\ddagger$]{Logan Douglas Schneider}
\author[ ]{Robby Bryant}
\author[ ]{Ryan G. Gomes}
\author[ ]{Allen Jiang}
\author[ ]{Roy Lee}
\author[ ]{Yun Liu}
\author[ ]{Javier Perez}
\author[ ]{Jameson K. Rogers}
\author[ ]{Cathy Speed}
\author[ ]{Shyam Tailor}
\author[ ]{Megan Walker}
\author[ ]{Jeffrey Yu}
\author[ ]{Tim Althoff}
\author[ ]{Conor Heneghan}
\author[ ]{John Hernandez}
\author[ ]{Mark Malhotra}
\author[ ]{Leor Stern}
\author[ ]{Yossi Matias}
\author[ ]{Greg S. Corrado}
\author[ ]{Shwetak Patel}
\author[ ]{Shravya Shetty}
\author[ ]{Jiening Zhan}
\author[ ]{Shruthi Prabhakara}
\author[$\dagger\S$]{Daniel McDuff}
\author[$\dagger\S$]{Cory Y. McLean}
\affil[ ]{Google LLC}
\affil[*]{Co-first author.}
\affil[$\ddagger$]{Core contributor.}
\affil[$\dagger$]{Co-last author.}
\affil[$\S$]{Corresponding authors: \texttt{\{belyaeva,xliucs,dmcduff,cym\}@google.com}.}

\begin{document}
\doparttoc 
\faketableofcontents

\maketitle

\begin{abstract}
Large language models (LLMs) can retrieve, reason over, and make inferences about a wide range of information. In health, most LLM efforts to date have focused on clinical tasks. However, mobile and wearable devices, which are rarely integrated into clinical tasks, provide a rich, continuous, and longitudinal source of data relevant for personal health monitoring.
Here we present a new model, \ourmodellong (\ourmodel), a version of Gemini fine-tuned for text understanding and reasoning over numerical time-series personal health data for applications in sleep and fitness. To systematically evaluate \ourmodel, we created and curated three novel benchmark datasets that test 1) production of personalized insights and recommendations from measured sleep patterns, physical activity, and physiological responses, 2) expert domain knowledge, and 3) prediction of self-reported sleep quality outcomes. For the insights and recommendations tasks we created \numcasestudies case studies in sleep and fitness. These case studies, designed in collaboration with domain experts, represent real-world scenarios and highlight the model's capabilities in understanding and coaching. Through comprehensive human and automatic evaluation of domain-specific rubrics, we observed that both \basemodel and \ourmodel are not statistically different from expert performance in fitness and, while experts remain superior for sleep, fine-tuning \ourmodel provided significant improvements in using relevant domain knowledge and personalizing information for sleep insights. To further assess expert domain knowledge, we evaluated \ourmodel performance on multiple choice question examinations in sleep medicine and fitness. \ourmodel achieved 79\% on sleep (N=629 questions) and 88\% on fitness (N=99 questions), both of which exceed average scores from a sample of human experts as well as benchmarks for receiving continuing credit in those domains. To enable \ourmodel to predict self-reported assessments of sleep quality, we trained the model to predict self-reported sleep disruption and sleep impairment outcomes from textual and multimodal encoding representations of wearable sensor data. We demonstrate that multimodal encoding is both necessary and sufficient to match performance of a suite of discriminative models to predict these outcomes. Although further development and evaluation are necessary in the safety-critical personal health domain, these results demonstrate both the broad knowledge base and capabilities of Gemini models and the benefit of contextualizing physiological data for personal health applications as done with \ourmodel.

\end{abstract}

\section{Introduction}

Large language models (LLMs) are versatile tools for generating language and have shown strong performance across a range of diverse domains. LLMs have achieved passing grades on the US legal bar exam~\cite{katz2024gpt} and second year medical school exams~\cite{singhal2023towards,nori2023can,saab2024capabilities}. In medicine in particular, natural language as an interface has shown potential to influence clinical practice~\cite{mcduff2023towards}, education, and research~\cite{meyer2023chatgpt}. When enriched with healthcare-specific data, LLMs attain impressive performance in medical question-answering~\cite{singhal2023towards,saab2024capabilities}, nuanced analysis of electronic health records~\cite{wornow2023shaky}, differential diagnosis from medical images~\cite{tu2023towards}, assessment of psychiatric functioning based on standardized assessments~\cite{galatzer2023capability}, and the delivery of psychological interventions~\cite{sharma2023cognitive,sharma2023facilitating,lin2024imbue}. High levels of performance on these tasks shows that LLMs have been able to effectively capture signal from  ``clinical data'', collected within a clinical setting.

Due to their sporadic nature, conventional clinical visits often fail to capture key aspects of human health and well-being that are measurable with wearable devices including sleep, physical activity, stress, and cardiometabolic health measured through physiological response and behavior. These continuous, longitudinal measures have significant advantages for health monitoring in that they are passively and continuously acquired, and provide direct physiological and behavioral signals. However, they have neither been deeply integrated into clinical practice nor incorporated into standard datasets used for medical question-answering~\cite{kao2017consumer, gordon2020beyond}, despite statistics on adverse health outcomes, morbidity, and Disability-Adjusted Life Years that underscore the profound impact these factors have on overall health~\cite{althoff2017large,chaput2020sleep,fogelholm2010physical,wannamethee1998changes,ohlin2004chronic}. This limited adoption is likely because these data are typically captured without context, are computationally demanding to store and analyze, and can be difficult to interpret. As a result, general foundation LLMs or even medically-tuned LLMs may lack the ability to use these data effectively to reason about and recommend interventions based on personalized individual health behaviors.

Mobile devices, including smart wearables and smartphones, have become instrumental tools for monitoring personal health metrics and gathering longitudinal data that cannot be obtained in traditional clinical settings~\cite{steinhubl2015emerging}. Unlike structured clinical data, personal health data is heterogeneous across data types, sources, and timescales~\cite{mcduff2024google}, ranging from continuous streams of biometric data from wearables, such as sleep patterns or heart rate, to sporadic and qualitative inputs like exercise logs, dietary logs, mood journals, and even social media activity.

\begin{figure}[t!]
    \centering
    \includegraphics[width=\textwidth]{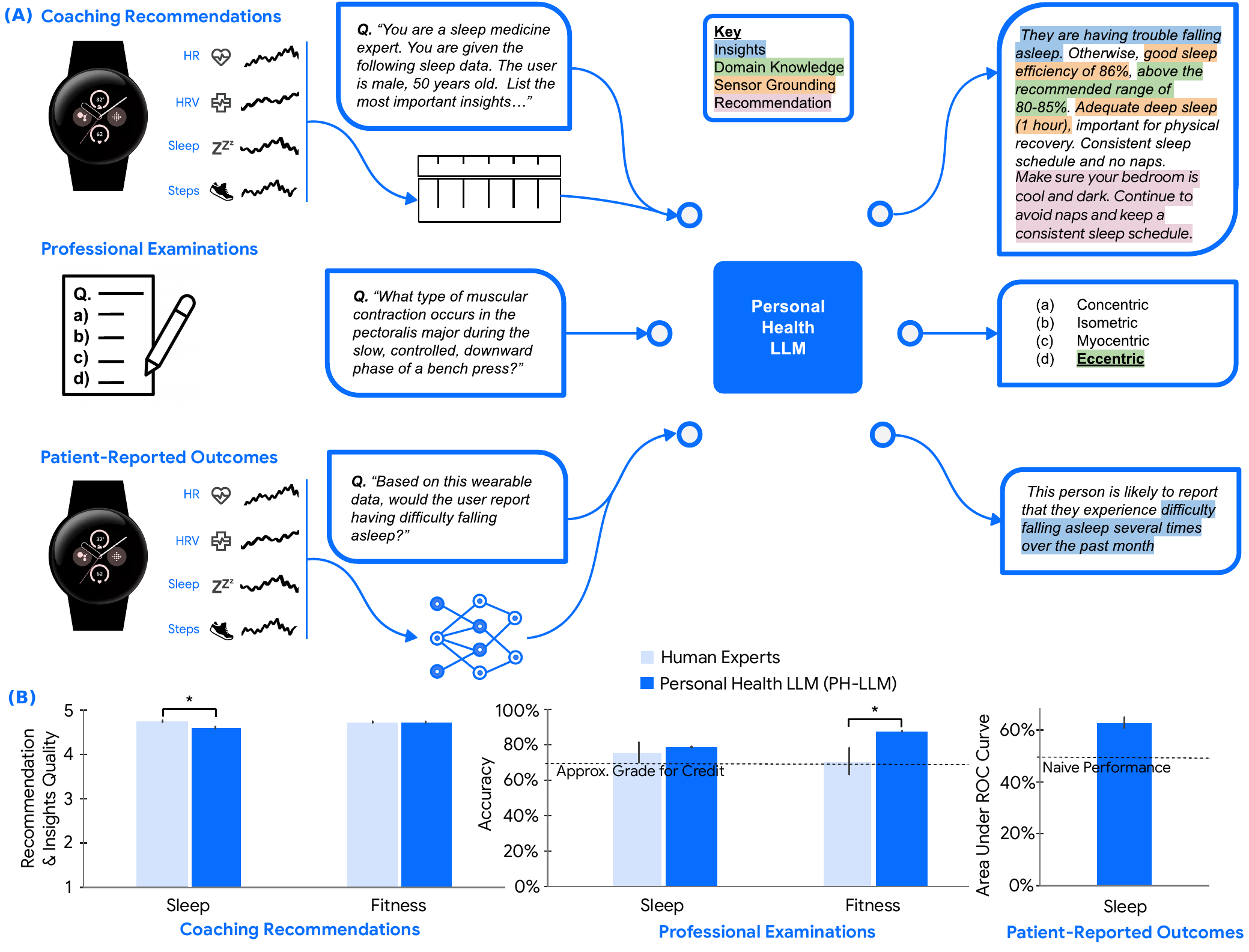}
    \caption{\textbf{\ourmodel: A Personal Health Large Language Model.} \textbf{(A)} We present \ourmodel, a version of Gemini fine-tuned for personal health and wellness. We evaluated \ourmodel on three aspects of personal health: generating personalized insights and recommendations for user goals in the domains of sleep and fitness, assessing levels of expert knowledge from certification examination style multiple choice questions, and predicting patient-reported outcomes in sleep quality from detailed sensor information. \textbf{(B)} Performance of \ourmodel contextualized with expert human responses. Error bars represent 95\% confidence intervals. ``$\ast$'' indicates a statistically significant difference between two response types. ``Naive Performance'' is that achieved by a random classifier. Human expert performance is not available for patient-reported outcome prediction from sensor features as this is not commonly performed, and no fitness-related outcomes were measured in the study assessing patient-reported outcomes~\cite{mcduff2024google}.}
    \label{fig:data_examples}
  \end{figure}

In this paper, we introduce \ourmodellong (\ourmodel), a version of Gemini fine-tuned to generate both insights about and recommendations to improve personal health behaviors related to sleep and fitness patterns. We evaluate the performance of \ourmodel~across three tasks: coaching recommendations, multiple choice exams assessing expert knowledge, and prediction of subjective patient-reported outcomes (PROs). The coaching recommendations tasks are tailored to two verticals of high personal health interest: sleep and fitness. The sleep tasks leverage individual's sleep metrics to derive insights, possible etiological factors, and provide personalized recommendations to improve sleep quality. The fitness tasks integrate information from training load, sleep, health metrics, and subjective feedback to provide personalized recommendations for the intensity of a physical activity that day. For the coaching recommendations tasks, we create the first personal health case study dataset to benchmark LLM performance on reasoning and understanding of personal health behaviors. This dataset consists of long-form questions grounded in summarized personal health behavior data, vertical-specific evaluation rubrics, and expert human responses for \numcasestudies case studies across sleep and fitness. Through rigorous human and automatic evaluation, we demonstrate that \basemodel approaches expert performance in fitness while fine-tuning \ourmodel enables it reduce the gap with experts in sleep coaching experiences, explore the breadth of personal health knowledge encoded within Gemini models, and apply multimodal capabilities to prediction of PROs in sleep (Figure~\ref{fig:data_examples}). Our key contributions are summarized as follows:

\begin{itemize}
    \item \textbf{\ourmodel:} We introduce a new model fine-tuned from Gemini for applications in personal health, capable of performing interpretation of time-series sensor data from wearables (i.e., Fitbit and Pixel Watch) for analysis and recommendations in sleep and fitness.
    \item \textbf{Long-form case studies from domain experts:} We create the first dataset of detailed personal health case studies in sleep and fitness curated by multiple experts in the associated domains. The dataset contains individual wearable sensor data and corresponding long-form insights and recommendations. We present rubrics for evaluation of long-form responses that span domain knowledge, use of user data, personalization, and potential for harm, and provide insights on training experts for accurate evaluation.
    \item \textbf{Benchmark and contextualize personal health question-answering:} We curate a set of validated domain-specific multiple choice examination questions on sleep and fitness, establish strong benchmarks based on continuing medical education requirements, and provide context for the scores through a set of human experts who completed a representative set of exam questions.
    \item \textbf{Multimodal sensor interpretation of self-reported outcomes:} We successfully integrate longitudinal time-series sensor features to interpret a user's subjective experience. To do so, we evaluate the capabilities of \ourmodel in predicting sleep disturbance and impairment PROs (acquired through validated survey instruments) from passive sensor readouts and show that accurate model performance requires native multimodal data integration.
\end{itemize}
\section{Personal Health Dataset Creation}

Owing to the absence of clearly defined language and multimodal datasets in the domain of personal health, we created datasets and associated tasks to evaluate different capabilities of \ourmodel. These datasets include case studies about real-world coaching recommendations, professional examinations that test domain knowledge about sleep medicine and fitness, and patient-reported outcomes about sleep. 

\subsection{Coaching recommendations} 
Many real-world applications of LLMs for personal health require realistic long-form text generation, which is challenging to evaluate automatically. As previously observed in the medical domain, strong performance on question-answering tasks does not necessarily transfer to the complexity of real-world tasks~\cite{fleming2024medalign}. To address the absence of rich long-form tasks for personal health data, in conjunction with domain experts and overseen by clinical leads we created detailed case studies that span two key personal health domains: sleep and fitness. Each case study was designed to interpret a range of physiological sensor information toward deriving insights, potential causes, or recommendations for future behaviors, and was sampled from high-volume anonymized production data from individuals who provided consent for research purposes.

\begin{figure*}[htp]
  \centering
    \includegraphics[width=0.65\textwidth]{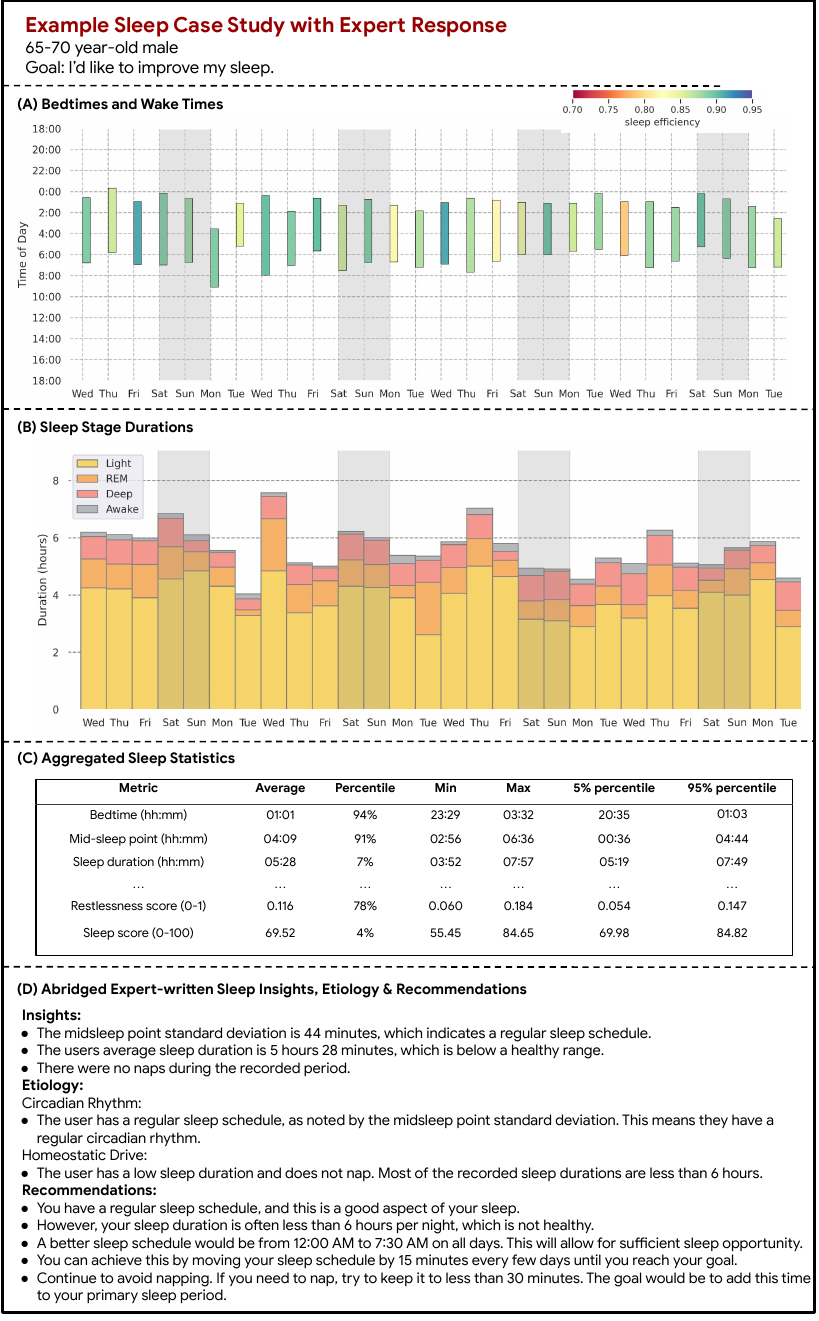}
    \caption{\textbf{Sleep case study example}: wearable sensor data used as input and corresponding expert analysis and recommendations for improving sleep quality. The experts considered individual's demographics and wearable sensor data for up to 29 days including daily metrics of \textbf{(A)} bedtimes and wake times and \textbf{(B)} time spent in various sleep stages and awake. For all daily metrics considered, see Table \ref{table:prompt_sleep_case_studies_sleep_logs}. The experts also analyzed \textbf{(C)} aggregated statistics of various sleep metrics. For a full list of aggregated statistics, see Table \ref{table:prompt_sleep_case_studies_sleep_summary}. The experts composed responses based on the data including \textbf{(D)} insights about individual's sleep, potential etiology, and recommendations for improving sleep quality.
    }
      \label{fig:casestudyexample_sleep}
  \end{figure*}

The sleep case studies aimed to enhance understanding of sleep patterns, identify causes of irregular sleep, and offer actionable recommendations based on these findings. Each case study incorporated wearable sensor data for up to 29 days, demographic information, and an expert analysis (Figure~\ref{fig:casestudyexample_sleep}). This comprehensive approach both facilitates a deeper understanding of health-related behaviors and also guides the development of personalized interventions to improve individual outcomes.

\begin{figure*}[htp]
  \centering
    \includegraphics[width=0.6\textwidth]{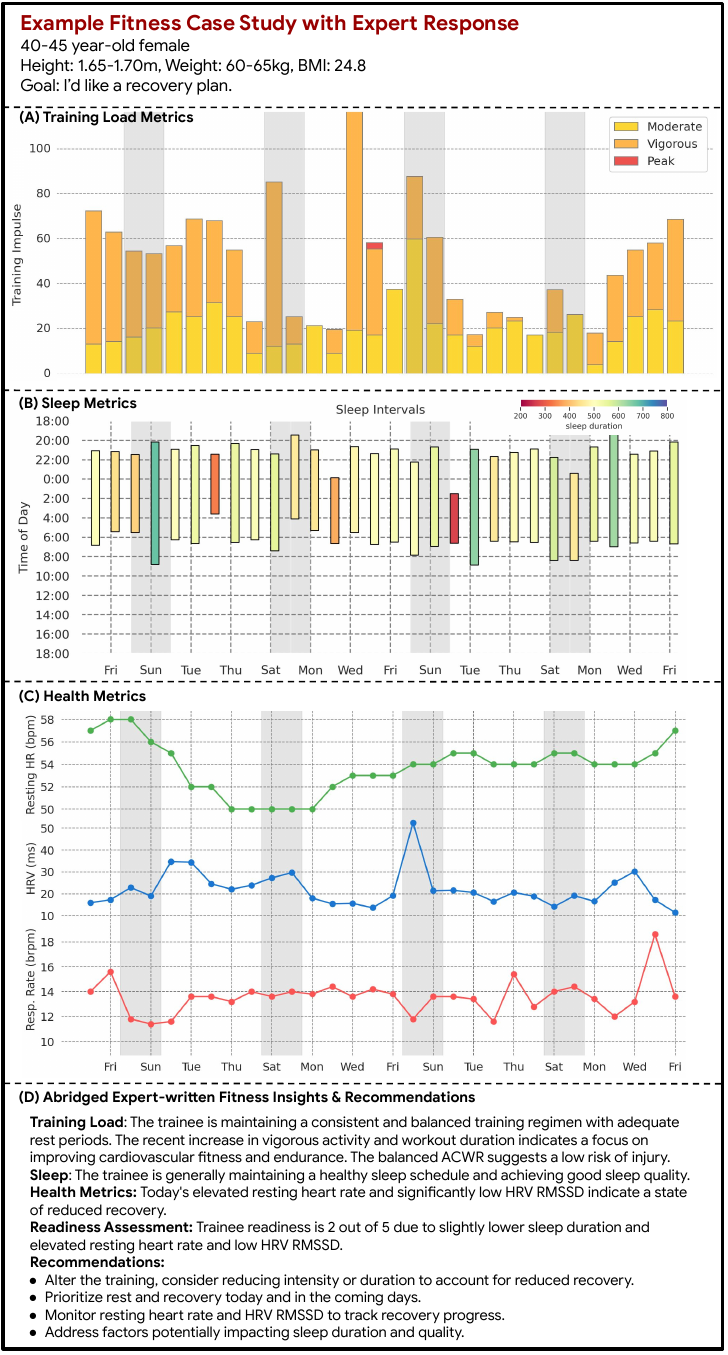}
    \caption{\textbf{Fitness case study example}: wearable sensor data used as input and corresponding expert analysis and recommendations. 
    The experts considered individual's demographics and wearable sensor data over a 30-day period including daily metrics of \textbf{(A)} cardiovascular training load such as training impulse, \textbf{(B)} sleep metrics such as the bedtimes and wake times, and \textbf{(C)} health metrics such as resting heart rate, heart rate variation, and respiratory rate. For all daily and aggregated metrics considered, see Tables \ref{table:prompt_fitness_case_studies_daily_activity}-\ref{table:prompt_fitness_case_studies_agg_health_metrics}. The experts composed responses based on the data including \textbf{(D)} insights about individual's training load, sleep, health metrics, and provided a workout readiness assessment and fitness recommendations. 
    }
      \label{fig:casestudyexample_fitness}
  \end{figure*}

The fitness case studies were designed to provide a comprehensive analysis of an individual's training load, sleep patterns, and health metrics, and were similarly based on wearable sensor data over 30 days, demographic information, and expert analysis (Figure~\ref{fig:casestudyexample_fitness}). Based on these inputs, the goal of the fitness-related task was to synthesize the suite of metrics into a data-driven assessment of the extent to which the individual is prepared for physical activity today and provide associated recommendations.

Each set of domain experts consisted of ``primary'' and ``secondary'' contributors to case study response creation and evaluation. This categorization was based on an expert’s general availability to contribute to the research project on a weekly basis throughout its duration; ``primary'' contributors had more involvement and higher volumes of case study response creation and evaluation than ``secondary'' contributors. The grouping was primarily used for research project operations planning and scheduling. The level of domain expertise was similar across the two groups. Each vertical also included a clinical lead with extensive background in sleep medicine for the sleep vertical and sport and exercise medicine for the fitness vertical. The clinical lead oversaw case study development and provided feedback and quality control to the set of domain experts.

\textbf{Sleep Case Study Creation.} In the development of sleep case studies, we recruited six domain experts in sleep medicine to craft guidance in the second person narrative, fostering a direct and personalized dialogue with the user. The six sleep experts all possessed advanced degrees (M.D., D.O., or Psy.D.) in sleep medicine and professional sleep medicine work experience ranging from 4 to 46 years. All experts were trained to read and interpret wearable data and map outputs to their corresponding sleep medicine literature counterparts. Experts were instructed to use best practices in goal-setting, emphasizing the creation of recommendations that are Specific, Measurable, Achievable, Relevant, and Time-bound (SMART)~\cite{doran1981there}. The data was sampled to achieve a representative group across age and gender (see Appendix \ref{apendix: additional_sleep_case_study} for details, \Cref{figure:case_study_stats}).

As illustrated in Figure \ref{fig:casestudyexample_sleep}, demographics information (age and gender), daily sleep metrics (e.g., bedtimes, wake times, and sleep stage durations), and aggregated sleep statistics (e.g., average bedtime) were selected collaboratively with the experts (Appendix \ref{apendix: additional_sleep_case_study}). These data were presented to the experts in both graphical and tabular formats for ease of analysis (Figure \ref{fig:casestudyexample_sleep}A-C, Table \ref{table:prompt_sleep_case_studies_sleep_logs}). Finally, the experts were tasked with composing responses across the following sections, aimed at analyzing the data with the objective of enhancing the sleep quality of the individual under consideration.

\texttt{Insights:} Implicitly this section was aimed at answering the question of ``What are some sleep-related insights based on my data?'' The sleep medicine expert examined the data and provided an interpretation of whether a data point might represent an atypical sleep pattern. The experts were asked to systematically review each case to provide a holistic assessment of the user's sleep patterns. To do so, Fitbit sleep metrics were assessed according to the validated RU-SATED framework (Routine, Sleep Quality, Alertness, Timing, Efficiency, and Duration) to generate sleep insights~\cite{buysse2014sleep}.

\texttt{Etiology:} Implicitly this section answered the question of ``What are the possible underlying causes that could explain the observed data?'' The experts generally considered the contribution of circadian rhythm, homeostatic drive, psychophysiologic hyperarousal, and extrinsic factors and indicated their likelihood.

\texttt{Recommendations:} This section was generally designed to answer the question of ``What can I do to improve my sleep?'' The experts were asked to provide personalized recommendations to the individual that can help them improve their sleep by addressing potential causes identified in the etiology section. The experts were instructed to utilize best practices in goal-setting using the SMART framework.

\textbf{Fitness Case Study Creation.} To construct fitness case studies (Figure~\ref{fig:casestudyexample_fitness}), we recruited seven domain experts in fitness to analyze an individual's quantitative fitness data. The seven fitness experts all possessed advanced degrees (M.S., M.A., M.Ed., or D.A.T.) related to the athletic training field and professional athletic training work experience ranging from 4 to 25 years. The experts were directed to formulate insights, assessments, and recommendations in the second person narrative. The data for fitness case studies were sampled to produce a variety of different fitness assessments (see Appendix \ref{apendix: additional_fitness_case_study} for details). The quantitative fitness data included a comprehensive array of metrics encompassing daily cardiovascular training load, sleep patterns, and health metrics spanning the preceding 30-day period (see Appendix \ref{apendix: additional_fitness_case_study} for details). These data were presented in tabular, text, and graphical formats. The experts were tasked with providing responses to the following sections, with the objective of facilitating a personalized approach to improving individual fitness levels by guiding on the intensity and duration of fitness sessions.

\texttt{Training Load:} The daily metrics over the past 30 days covered steps, time spent in different heart rate zones, and training impulse (TRIMP), which is a training load measure derived from heart rate and exercise duration (Figure \ref{fig:casestudyexample_fitness}A, Table \ref{table:prompt_fitness_case_studies_daily_activity}). Recent exercise logs and aggregate analyses were also provided, including acute and chronic training loads, and the Acute-Chronic Workload Ratio (ACWR), aiding the assessment of training stress (Table \ref{table:prompt_fitness_case_studies_agg_daily_metrics}).

\texttt{Sleep Metrics:} The data included daily sleep measurements such as bedtime, wake time, duration of different sleep stages, and the sleep score (Table \ref{table:prompt_fitness_case_studies_sleep_metrics}), along with aggregated statistics for comparison (Table \ref{table:prompt_fitness_case_studies_agg_sleep_metrics}). Visual aids facilitated the analysis, with select sleep metrics graphically represented over time (Figure \ref{fig:casestudyexample_fitness}B).

\texttt{Health Metrics:} Daily data on resting heart rate, heart rate variability (HRV), and respiratory rate were provided (Table \ref{table:prompt_fitness_case_studies_health_metrics_month}), along with graphical representations (Figure \ref{fig:casestudyexample_fitness}C) for easier interpretation. Aggregated metrics (Table \ref{table:prompt_fitness_case_studies_agg_health_metrics}) allowed experts to gauge changes over time and assess recovery and stress levels effectively.

\texttt{Assessment \& Recommendation:} The information from the previous sections was used to provide a summary of the most important insights. These insights along with synthetically generated user input on subjective readiness and muscle soreness (e.g., Tables \ref{table:prompt_fitness_case_studies_subjective_readiness} and \ref{table:prompt_fitness_case_studies_muscle_soreness}) were used to inform an assessment of how ready the individual is to perform a workout today on the scale of 1 to 5. The experts also provided fitness recommendations to the individual (Figure \ref{fig:casestudyexample_fitness}D).

\textbf{Holistic View of Case Study Creation.} For both the sleep and fitness verticals, we generated two sets of data: a dataset used for model training, validation, and testing and a holdout dataset that was only used for final evaluation of the model by experts (\Cref{figure:casestudyworkflow}).

To generate the dataset used for training, validation, and testing, we first prompted the Gemini family of models with the data for each section in order to generate baseline model (\basemodel) responses (\Cref{figure:casestudyworkflow}A). The experts then reviewed the responses and rewrote them as needed. The dataset also underwent multiple rounds of quality control engaging the experts and clinical leads. Separately, to generate the holdout dataset, the experts wrote the responses from scratch (without any LLM assistance). This was done to ensure a more clear comparison between experts and the model during evaluation.

In total, we created 350 case studies for fitness (300 case studies for the training, validation, and test set and 50 case studies for the holdout set) and 507 case studies for sleep (457 case studies for the training, validation, and test set and 50 case studies for the holdout set).
  
\subsection{Professional examinations} 

\textbf{Sleep Medicine Exams.} We compiled a set of \numsleepmcq multiple choice questions (MCQs) from BoardVitals\footnote{https://www.boardvitals.com/} sleep medicine board review question banks. We used text exam questions from the American Medical Association (AMA) Physician's Recognition Award (PRA) ``Category 1 - Sleep Medicine'' question bank, which emulates exam content for the American Board of Internal Medicine (ABIM) Sleep Medicine Certification Exam. We also used text exam questions from the Sleep Medicine Maintenance of Certification (MOC) Exam and Longitudinal Knowledge Assessment Review question bank, which emulates exam content for the ABIM Sleep Medicine MOC Exam and ABIM Longitudinal Knowledge Assessment. This compiled set of MCQs spanned a wide range of sleep-related topics: Normal Sleep and Variants (N=127), Breathing Disorders (N=84), Hypersomnolence (N=60), Insomnias (N=85), Movement Disorders (N=23), Parasomnias (N=57), Sleep in Other Disorders (N=112), and Sleep-Wake Timing (N=81). 

\textbf{Fitness Exams.} We compiled a set of \numfitnessmcq multiple choice questions sourced from multiple question banks that emulate exam content for the Certified Strength and Conditioning Specialists (CSCS) exam preparation book provided by the National Strength and Conditioning Association (NSCA)\footnote{https://www.nsca.com/certification/cscs/certified-strength-and-conditioning-specialist-exam-description}. We used the test exam questions from the NSCA-CSCS textbook “Essentials of Strength Training and Conditioning”.

Each exam question presents up to five possible answers, with a single correct answer, facilitating automated and quantitative assessment of performance.

\subsection{Patient-reported outcomes} 
\label{methods:pros}

To evaluate the ability of \ourmodel to predict patient-reported outcomes (PROs) from longitudinal passive sensor data, we used a large IRB-approved study in which wearable data was collected for a population of 4,759 consented individuals for a four-week period~\cite{mcduff2024google}. At both intake and completion, participants completed the Patient-Reported Outcomes Measurement Information System (PROMIS)\footnote{https://commonfund.nih.gov/promis/index} short-form Sleep Disruption and Sleep Impairment surveys~\cite{yu2012development}. Both surveys contained eight items with answers on a 5-point Likert scale (Appendix~\ref{sec:pro_surveys}). The study thus linked individuals' perceived sleep quality and its impact on their functioning with longitudinal observed physiological (e.g., heart rate, sleep duration) and behavioral (activity) measurements.

To maximize sample size, we used the intake survey responses as the basis for prediction. For each question, we defined a binary outcome that compared the highest answer (e.g., ``strongly agree'') against all others (Supplementary Figure~\ref{fig:supp:survey_distributions}).
Features used to predict each binary outcome included 20 time-varying wearable measurements (Table~\ref{table:pro_sensor_features}), each of which was collected from study participants over a four-week span. While most individuals had sensor data for over 21 days, distributions were heavily left-skewed (Supplementary Figure~\ref{fig:supp:sensor_distributions}).
To obtain a rectangular dataset we retained only individuals with at least 15 days of sensor data (N=7,114) and downsampled all individuals to a set of 15 contiguous days. We imputed all remaining missing values with the population median computed using all available data from training set individuals, resulting in a $20 \times 15$ matrix that represents the wearable sensor data for each research participant over 15 days. Furthermore, we performed standard filtering for data quality by removing any data points that were more than four standard deviations from the population median for each sensor value. No imputation was performed for survey answers.

\section{Methods}

\subsection{Base model selection} In order to start from the most capable base model, we performed automated evaluation of several Gemini candidate model sizes and a medical LLM on the professional exam questions. The candidate models were Gemini Nano 1.0, Gemini Pro 1.0, \basemodel~\cite{team2023gemini}, and MedPaLM-2. \basemodel consistently produced the best accuracy on professional examinations (\Cref{fig:mcq_scaling_comp,fig:mcq_scaling_comp_stratified}).

\subsection{Base model prompting on case studies}

Since \basemodel was the most accurate model on professional examinations, suggesting it has appropriate domain knowledge in the areas of sleep and fitness, we explored the performance of this model on case studies. We prompted \basemodel by summarizing guidelines given to the experts for dataset creation. For example, the sleep experts generally were asked to follow the RU-SATED format (Routine, Sleep Quality, Alertness, Timing, Efficiency, and Duration)~\cite{buysse2014sleep} to generate sleep insights. In order to give \basemodel the best shot at answering case studies, we similarly prompt it to follow the RU-SATED format and provide an explanation of what metrics should be used to assess each dimension (see \Cref{table:prompt_sleep_case_studies_insights}-\ref{table:prompt_fitness_case_studies_readiness_assessment} for details). We note that each case study consisted of multiple sections representing different queries and responses: three sections for sleep case studies (insights, etiology, recommendations) and five sections for fitness case studies (demographics, training load, sleep, health metrics, and the assessment). Since each section represented a different aspect of the case study, we developed prompts specifically for each section. Tables \ref{table:prompt_sleep_case_studies_insights}-\ref{table:prompt_sleep_case_studies_sleep_summary} show the prompts for sleep case studies and Tables \ref{table:prompt_fitness_case_studies_demographics}-\ref{table:prompt_fitness_case_studies_readiness_assessment} show the prompts for fitness case studies. For sections that synthesized results from previous sections, i.e., the etiology and recommendation sections in sleep case studies, and the assessment section in fitness case studies, we substituted the model answers from previous sections into the prompt (see Table \ref{table:prompt_fitness_case_studies_readiness_assessment} for an example).

\subsection{Training \ourmodel on case studies}
\label{section:methods:phllm_training_text}

We fine-tuned \basemodel on the dataset of coaching recommendations and call this model \ourmodel. We use the case studies from the training, validation, and test sets for model training and selection (457 case studies for sleep and 300 case studies for fitness). For each of the sleep and fitness domains, we randomly split the dataset into separate training, validation, and test splits using a 70:15:15 ratio. We used the same prompts that were given to the baseline model to form prompt-response pairs for model tuning. Since each section was treated as a separate example, this resulted in 1,371 prompt-response pairs for sleep and 1,500 prompt-response pairs for fitness across the training, validation, and test sets (\Cref{figure:casestudyworkflow}A,B).

Typically, LLMs are trained on mixture of tasks \cite{wei2021instructiontune}. Here we fine-tuned the model on a 1:1 mixture of sleep and fitness prompt-response pairs. Within the fitness prompt-response pairs, we chose to upsample higher quality case studies by a 2:1 ratio, where higher quality case studies were defined as those that underwent additional rounds of quality control by the fitness experts.

 The model was fine-tuned for a maximum of 1500 steps with a global batch size of 4 using linear warm-up over 50 steps and cosine decay. We used a learning rate of $2.5 \times 10^{-7}$, weight decay of $1 \times 10^{-2}$, and a learning rate decay minimum ratio of 0.1. We saved model checkpoints every 50 steps. For our final model candidate, we chose the first checkpoint after the model had been trained for at least one epoch (this checkpoint also had a relatively low log perplexity).

\subsection{Training \ourmodel for patient-reported outcomes}

To train \ourmodel to predict PROs from wearable data, we followed the methodology developed in HeLM~\cite{belyaeva2023multimodal}. Wearable data for each user was stored as a matrix in which the rows represent wearable measurement devices and the columns represent measurements at a specific time. In our case, we had 20 device measurements measured once over 15 days for each sample in the dataset. Next, we encoded this data by computing the mean and variance across days, and z-scoring the results using the training data as a reference. This yielded a new ``encoded'' matrix of $20 \times 2$ where columns correspond to a measure's mean and variance. The encoded data matrix was projected into the token embedding space of \ourmodel via a multilayer perceptron (MLP) adapter with three hidden layers (sizes 1,024, 4,096, and 1,024) and an output of $2$ tokens. The resulting set of tokens were provided to \ourmodel as a prefix to the text input, which included a text representation of all input fields in their native form (e.g., steps per day; not z-scored). We prompted the model to predict a specific binary outcome (e.g., ``I am satisfied with my sleep - `yes' or `no'''). An example of the corresponding text prompt is shown in \Cref{table:prompt_pro}. The adapter was trained via backpropagation while keeping \ourmodel weights frozen.

We compared these adapter-based predictions to text-only predictions using both zero-shot and few-shot prompting. For zero-shot, the prompt format was identical to the adapter-based prediction except the adapter token prefix was omitted. For few-shot, as many complete examples as could fit within the context window (up to seven) were included as exemplars. For all three models, the positive and negative outcomes were scored by computing the log likelihood for each outcome.

Text prompts that included using only mean results, and both mean and variance, were explored (while always including both mean and variance in the input to the MLP adapter). Since performance was not appreciably different (data not shown), we omitted the variance encoding to enable more in-context examples to be passed as textual context to \ourmodel.

As a separate comparison, we fitted logistic regression models separately for each binary outcome, in which the predictors were the same mean and variance computed across 15 days of sensor data. For training both the logistic regression and MLP adapter we used a set of 4,978 training examples. For the MLP adapter, we selected the best model using a validation set of 703 individuals and we produced all results presented here in a holdout set of 1,433 samples.

\subsection{Expert grading of case study responses}
\label{section:methods:expert_rating}

While evaluation against MCQs and PROs can be performed by comparing model predictions to gold-standard structured responses and numerical values, respectively, the case studies involve longer-form outputs.  

In order to evaluate these longer-form case study responses, the domain experts (including all individuals involved in creating the case study responses) were asked to evaluate three responses written to each case study: one by \basemodel, one by \ourmodel, and one by a domain expert. Each domain expert was assigned evaluations randomly to case studies for which they did not write the expert response. The domain experts evaluated each case study response based on a custom rubric that quantifies incorporation of user data, appropriate personalization based on user data, use of expert domain knowledge, evidence of confabulations or unwarranted assumptions, potential for harm, readability, and overall quality. The complete set of evaluation questions for the case studies is provided in Appendix~\ref{app:rubrics}.

Evaluation cases were fully distributed across the primary group of experts based on availability during the research project’s evaluation period. A portion of the evaluation case studies were additionally assigned to the rest of the available domain experts to ensure on-schedule, thorough completion of the evaluation dataset.

Both the creation of expert written case study responses and the evaluation of all 3 types of responses were performed on an internal health data labeling platform that adheres to data privacy and security best practices and design principles. It handles labeling task creation, scheduling and assignment, answer storage as well as front-end visualization and labeling through its web application. It supports highly customizable viewers for multiple data modalities including medical images and text reports. We customized the HTML viewer to display long-form case studies comprising figures, tables, and text, in an effective and intuitive manner.

\subsection{Automatic evaluation of case study responses}
\label{section:methods:autoeval}

Though expert grading of case study responses was our primary mechanism for assessing model performance, it is a time-consuming process that scales poorly. This makes it challenging to iterate on model improvements since sending all checkpoints to human raters is prohibitively expensive. Automated evaluation (AutoEval) allows us to obtain a quick--though potentially less accurate than human evaluation--signal that can be used during model development by using secondary models to perform this rating task \cite{chiang2023autoeval}. In this section, we describe our approach for curating a case study response rating dataset, fine-tuning AutoEval models capable of rating candidate models, and using AutoEval to select promising models that are then sent to expert raters for human feedback.

While exploring different modeling mechanisms, we performed an initial round of expert grading using the rubrics and procedure described in \Cref{section:methods:expert_rating} for 50 expert-generated case studies from each vertical across three response sources: experts, an untuned \basemodel model, and a fine-tuned Gemini Pro 1.0 model. We then split these studies into vertical-specific training and validation splits containing roughly 80\% ($N=38$) and 20\% ($N=12$) of case studies, respectively. Splits were structured such that samples rated by a given expert were evenly distributed between sets. All ratings associated with a given case study were included in that split, resulting in $N=6,552$ total ratings across case study sections and evaluation principles for sleep ($N=4,872$ train; $N=1,596$ validation) and $N=9,331$ for fitness ($N=7,138$ train; $N=2,193$ validation). Using these ratings and the corresponding case study data and responses, we constructed LLM prompts and targets matching the format described in \Cref{table:autoeval_prompt_format} (see \Cref{table:autoeval_prompt_format_example} for a full example). Prompts included a description of the rating task objective for the given case study section, a summary of data describing the case study, the principle being assessed, and the principle's Likert scale options. Each target was the expert-generated rating followed by the rating's Likert option text description (e.g., for a ``No incorrect domain knowledge'' principle rating of 5, the target is ``5. No incorrect domain knowledge references exist.'').

We fine-tuned Gemini Pro 1.0 models using LoRA \cite{hu2022lora} across a variety of vertical-specific data mixtures, including all ratings for a vertical and all ratings from a single rater. All AutoEval modeling experiments used a fixed set of hyperparameters, varying only the training data mixture: a LoRA rank of 4 on attention heads, a constant learning rate of $2 \times 10^{-5}$, a global batch size of 32, and a maximum of 20 epochs for the given training mixture. We present results for the following data mixtures:

\begin{enumerate}
    \item All ratings in either the fitness or sleep verticals (``All'').
    \item All ratings from the lowest variance rater in the fitness (``Fitness Primary B'') or sleep (``Sleep Primary D'') verticals, where variance is calculated across all ratings from that expert.
    \item All ratings from the highest variance rater in the fitness (``Fitness Primary C'') or sleep (``Sleep Primary C'') verticals.
\end{enumerate}

An untuned Gemini Pro 1.0 model served as a baseline. We generated model predictions by scoring the likelihood of each Likert option given the input prompt, converted these scores into five-class multinomial probabilities, and chose the option with the largest probability score. We selected candidate AutoEval models using a combination of log perplexity loss and Spearman's rank correlation between predictions and the ground truth ratings in the validation dataset.

Given case study responses from candidate \ourmodel models trained using the procedure described in \Cref{section:methods:phllm_training_text}, we used the same scoring procedure above to automatically rate model outputs across case study sections and evaluation principles. We used these ratings in conjunction with non-expert feedback to filter candidate models for full human expert evaluation. We then used the resulting ratings to further evaluate the performance of our final AutoEval models.

\subsection{Qualitative interviews}

Following the study we performed a semi-structured 30-minute interviews with the experts who participated in creating, editing, and evaluating the case studies. Semi-structured interviews explored the following questions:  

\begin{tcolorbox}
\begin{enumerate}
    \item When creating case studies from scratch, how did you find the task of writing the case studies?
    \item When editing the LLM text, how did you find the task of editing the case studies? Did you feel you made substantive edits to the LLM text overall? What areas needed the most improvement?
    \item When creating/editing the case studies which numerical data was helpful? Which data was unhelpful? Why?
    \item When evaluating the case study text what was easy or difficult about the different sections?
    \item How did you rate overall quality?
    \item In what ways could the case study responses be potentially harmful?
\end{enumerate}
\end{tcolorbox}

\subsection{Statistical analyses}
Confidence intervals (95\%) were determined via bootstrapping with 1,000 iterations. Statistical significance of expert ratings was determined using Wilcoxon rank-sum test with false-discovery rate (Benjamini-Hochberg) correction when multiple sections or multiple evaluation principles were analyzed together. All p-values below refer to p-values after FDR correction.
\section{Results}

\subsection{\ourmodel approaches expert performance on long-form case studies} 

We evaluated the aggregated performance of \ourmodel and human experts on the long-form case study responses, rated by human experts using 15 questions with grading scale 1 through 5, spanning topics such as using important domain knowledge, correctly referencing relevant user data, and avoiding confabulations. A rating of 5 indicates high quality: for example, a 2 or 3 indicates many or several important data interpretations are missing, while a 4 or 5 indicates few or none missing. All 15 questions and rating descriptions are detailed in Appendix~\ref{app:rubrics}. For sleep case studies, \ourmodel received an average rating of 4.61 versus 4.75 for human experts, indicating a close match ($p=3.3 \times 10^{-11}$, $N \geq 2606$, Figure \ref{fig:data_examples}A). Although the difference is statistically significant, the effect size is small and our model responses are high quality as indicated by receiving the top rating of five 73\% of the time. Fine-tuning \ourmodel on sleep case studies significantly improved its overall performance in this task (average rating of 4.51 versus 4.61, $p=4.0 \times 10^{-6}$, $N \geq 2603$). For fitness case studies, \ourmodel aggregate performance was not statistically different from expert performance ($p=0.48$, $N \geq 3335$, Figure \ref{fig:data_examples}B). \basemodel responses were also statistically indistinguishable from human expert responses ($p=0.57$, $N \geq 3161$).

\begin{figure}[htp]
  \centering
    \includegraphics[width=\columnwidth]{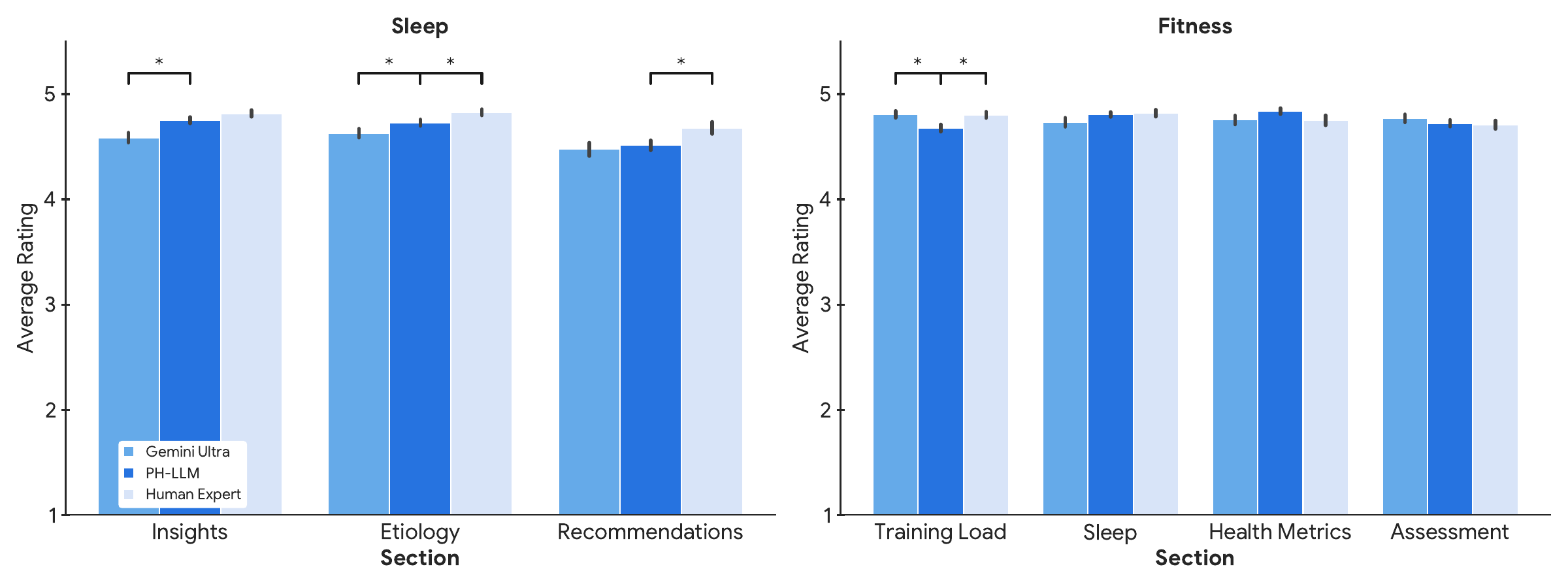}
    \caption{\textbf{Case Study Human Evaluation Results.} Mean ratings given by experts for the case study subsections across the \textbf{(A)} sleep and \textbf{(B)} fitness domains. ``$*$'' indicates a  statistically significant difference between two response types after multiple hypothesis testing correction.}
    \label{fig:case_study_human_eval_barplot}
\end{figure}

Since the case studies consist of multiple sections, we also analyzed ratings for each section separately (Figure \ref{fig:case_study_human_eval_barplot}A). For sleep case studies, fine-tuning \ourmodel improved its ability to provide insights and etiologies ($p=6.65 \times 10^{-7}$, $N \geq 800$ and $p=2.46 \times 10^{-3}$, $N \geq 801$, respectively), with recommendations showing no statistically significant difference ($p=0.44$, $N \geq 801$). We further analyzed ratings by various rubric questions. Fine-tuning \ourmodel improved its ability specifically on being able to reference important domain knowledge ($p=4.46 \times 10^{-5}$, $N \geq 201$), important interpretations ($p=4.46 \times 10^{-5}$, $N \geq 201$), important user data ($p=5.21 \times 10^{-8}$, $N \geq 201$), and no unimportant interpretations ($p=4.30 \times 10^{-2}$, $N \geq 201$), see Figure \ref{figure:case_study_principle}. Overall, these results suggest that fine-tuning improved the model's ability to mention relevant domain knowledge, relevant interpretations, and relevant user data, especially when deriving insights and etiology from the data.

For fitness case studies, \ourmodel had similar performance (no statistically significant difference detected, $N \geq 768$) to human experts on three out of four sections (Figure \ref{fig:case_study_human_eval_barplot}B). Training load was the only section in which responses from human experts were rated higher than those from \ourmodel ($p=0.005$, $N \geq 768$). When analyzing ratings by rubric questions, we observed no statistically significant differences in ratings between \ourmodel and human experts (Figure \ref{figure:case_study_principle}). 

Examples of \ourmodel model responses are presented in Tables \ref{table:model_response_sleep_insgihts}-\ref{table:model_response_sleep_recs} for a sleep case study and in Tables \ref{table:model_response_fitness_demographics}-\ref{table:model_response_fitness_recommendation} for a fitness case study.

Furthermore, our fine-tuned AutoEval models can act as strong proxies for expert annotation. The best AutoEval models ranked case study response sources similarly to human experts (compare \Cref{fig:autoeval_ratings_high_var_raters} to Figure \ref{fig:case_study_human_eval_barplot}). When measuring Spearman's rank correlation, Kendall's Coefficient of Concordance (Kendall's W), Weighted Cohen's Kappa, and Interclass correlation (ICC(3,1)) between AutoEval rating predictions and ground-truth human ratings across the validation dataset in both verticals, the best AutoEval models obtained similar prediction-rating agreement when compared to inter-rater agreement metrics (\Cref{section:appendix:case_studies:primary_vs_secondary}).

We explored different AutoEval training data mixtures and found that all mixtures produced models that significantly improved upon a Gemini Pro 1.0 rater not explicitly fine-tuned for AutoEval tasks (\Cref{table:autoeval_validation_performance}). Additionally, we observed that models tuned using only ratings from a single, high variance rater outperformed models tuned using all ratings from the given vertical and models tuned using only ratings from a single, low variance rater, though these differences were not statistically significant due to limited sample size.

Finally, we investigated variation in rating time. Interestingly, primary raters were faster than secondary raters across both verticals. As expected, however, a single AutoEval model rating case study responses sequentially is significantly faster at rating than both primary and secondary raters (\Cref{table:auto_eval_time_to_rate}). Furthermore, since rating a conversation for a given principle is embarrassingly parallel, we can split tasks across model replicas to decrease time-to-rate to that of a single principle for a single section, decreasing time-to-rate by two orders of magnitude.

\subsection{\ourmodel exceeds grades to receive CME credit on sleep medicine and pass the fitness certification practice examination question banks} 

\ourmodel correctly answered 79\% of sleep medicine and 88\% of fitness board examination questions tested, comfortably exceeding the approximate grade (70\%) to either receive CME credit for sleep or pass the practice exam for fitness (Table \ref{tab:mcqs_main_results}).  On the AMA PRA Category 1 and ABIM MOC, \ourmodel scored 79\% while \basemodel scored 77\%. On the NSCA-CSCS coaching exams both \ourmodel and \basemodel scored 88\%. The sleep medicine question bank contained additional metadata for each question including the distribution of responses from human test takers, enabling comparisons of performance by empirical question difficulty. Both \ourmodel and \basemodel performed comparably across the question difficulty strata and suggest that the performance of \ourmodel is comparable to that of humans who have prepared for or are in the process of preparing for these examinations (Table \ref{tab:mcqs_diff_index}). To further contextualize the performance of \ourmodel with experts, five professional athletic trainers (average experience: 13.8 years) and five sleep medicine experts (average experience: 25 years) with advanced degrees were recruited to take the respective exams. The experts achieved an average score of 71\% in the fitness exam and an average score of 76\% in a representative subset of the sleep medicine exam (N=204) stratified based on medical content categories\footnote{https://www.abim.org/Media/aypkdxpi/sleep-medicine.pdf} and their difficulty levels. As illustrated in Table \ref{tab:mcqs_main_results}, \ourmodel outperforms expert graders on both professional exam question banks.

We performed ablation studies on the use of self-consistency \cite{wang2022self} (N=5 rounds) and chain-of-thought (CoT) prompting \cite{wei2022chain}. Self-consistency improved performance on fitness questions for both CoT and Non-CoT prompting techniques while the performance from including CoT was mixed (\Cref{table:mcq_ablation_cot_sc}). Question prompts, along with examples of correct and incorrect answers, can be found in Appendix~\ref{app:mcqs}. Results from additional models can be found in Appendix~\ref{appendix:additional_models_mcqs_perf}.

\begin{table}[htp]
\setlength{\tabcolsep}{3pt}
\caption{\textbf{Performance on Professional Exam Question Banks.} Accuracy on the multiple choice questions from AMA PRA Category 1 - Sleep Medicine and ABIM MOC - Sleep Medicine MOC and NSCA-CSCS coach certification examination question banks.}
\label{tab:mcqs_main_results}
\small
\centering
\begin{tabular}{lcccc}
\toprule[1.5pt]
\textbf{Sleep Medicine}& Approx. CME Grade & Expert & \basemodel & \ourmodel \\
\midrule
AMA PRA Category 1 / ABIM MOC & 70\%$^{*}$& 76\% & 77\% & \textbf{\textcolor{forestgreen}{79\%}} \\[1.2pt]
\bottomrule[1.5pt]
\textbf{Fitness} & Approx. Pass Grade & Expert & \basemodel & \ourmodel \\
\midrule
NSCA-CSCS Coaching Certification & 70\%$^{\dagger}$ & 71\% & \textbf{\textcolor{forestgreen}{88\%}} & \textbf{\textcolor{forestgreen}{88\%}} \\[1.2pt]
\bottomrule[1.5pt]
\multicolumn{4}{l}{
$*$ {\tiny\url{https://www.boardvitals.com/sleep-medicine-moc-recertification}.}} \\
\multicolumn{4}{l}{
$\dagger$ {\tiny\url{https://www.nsca.com/certification/cscs/certified-strength-and-conditioning-specialist-exam-description}}} \\
\end{tabular}
\vspace{-0.2cm}
\end{table}

\begin{table}[htp]
\setlength{\tabcolsep}{3pt}
\caption{\textbf{Performance Comparison of Models and Experts Relative to Average Reported Test Takers for the Sleep Professional Exam.} Questions were classified as ``Easy'', ``Medium'', or ``Hard'' based on the percentage range of human test takers who answered the corresponding questions correctly.}
\label{tab:mcqs_diff_index}
\small
\centering
\adjustbox{max width=\textwidth}{
\begin{tabular}{lcccc}
\toprule[1.5pt]
Difficulty & Count & Expert & \basemodel & \ourmodel\\
\midrule
Easy (90\%-100\%) & 214 & 90\% & 94\% & \textbf{\textcolor{forestgreen}{95\%}} \\[1.2pt]
Medium (75\%-90\%) & 204 & \textbf{\textcolor{forestgreen}{81\%}}  & 78\% & 80\% \\[1.2pt]
Hard (0\%-75\%) & 211 & 53\% & 55\% & \textbf{\textcolor{forestgreen}{57\%}} \\[1.2pt]
\bottomrule[1.5pt]
\end{tabular}
}
\vspace{-0.2cm}
\end{table}

\subsection{Multimodal sensor integration enables \ourmodel to predict patient-reported outcomes}

\begin{figure}
\renewcommand{\thesubfigure}{\Alph{subfigure}}
\captionsetup[subfigure]{justification=justified,singlelinecheck=false}
     \centering
     \begin{subfigure}[b]{0.49\textwidth}
         \caption{}
         \centering
         \includegraphics[width=\textwidth,trim={0 -7.2cm 0 0}]{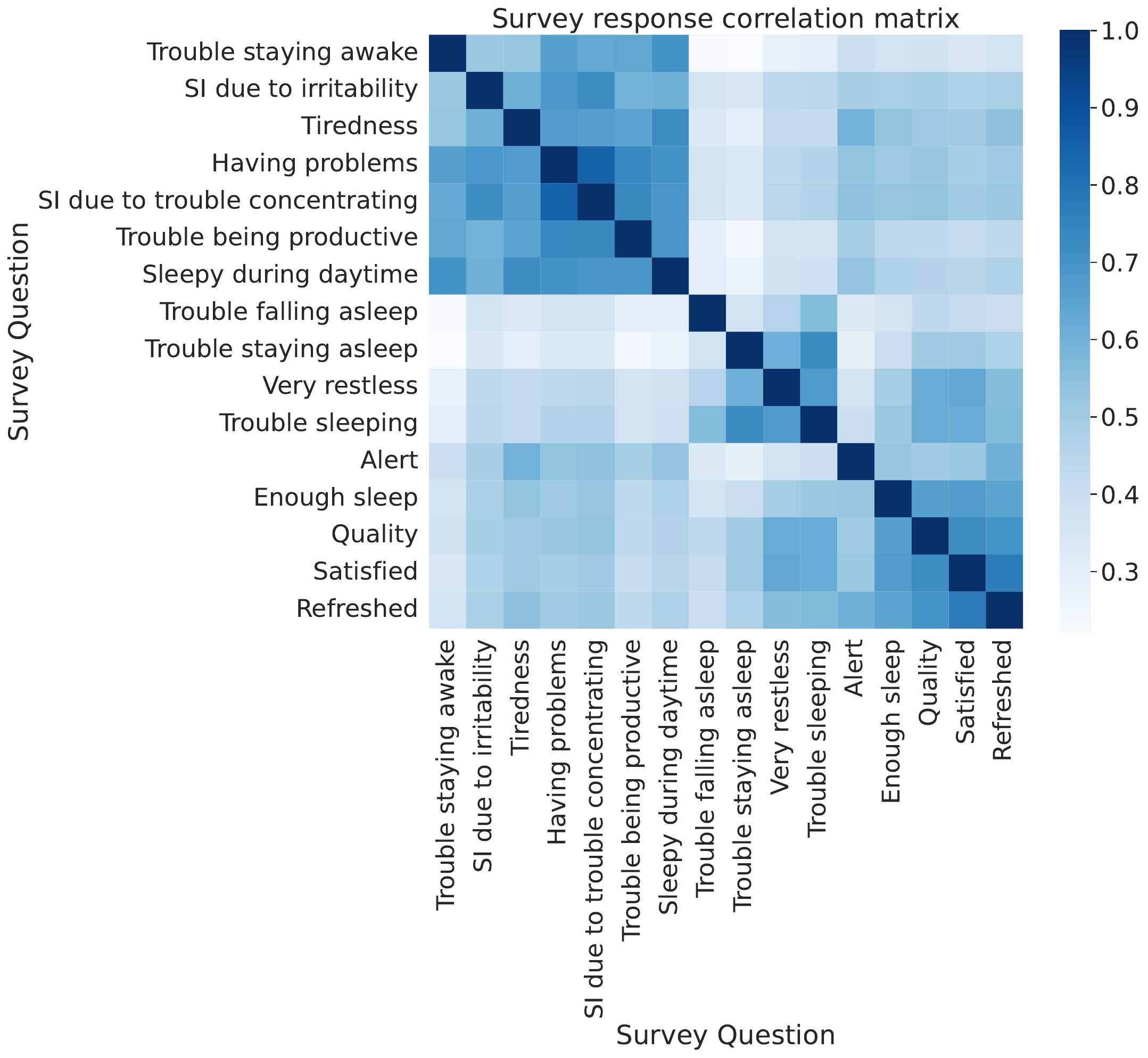}
         \label{fig:survey_response_correlations}
     \end{subfigure}
     \hfill
     \begin{subfigure}[b]{0.49\textwidth}
         \caption{}
         \centering
         \includegraphics[width=\textwidth]{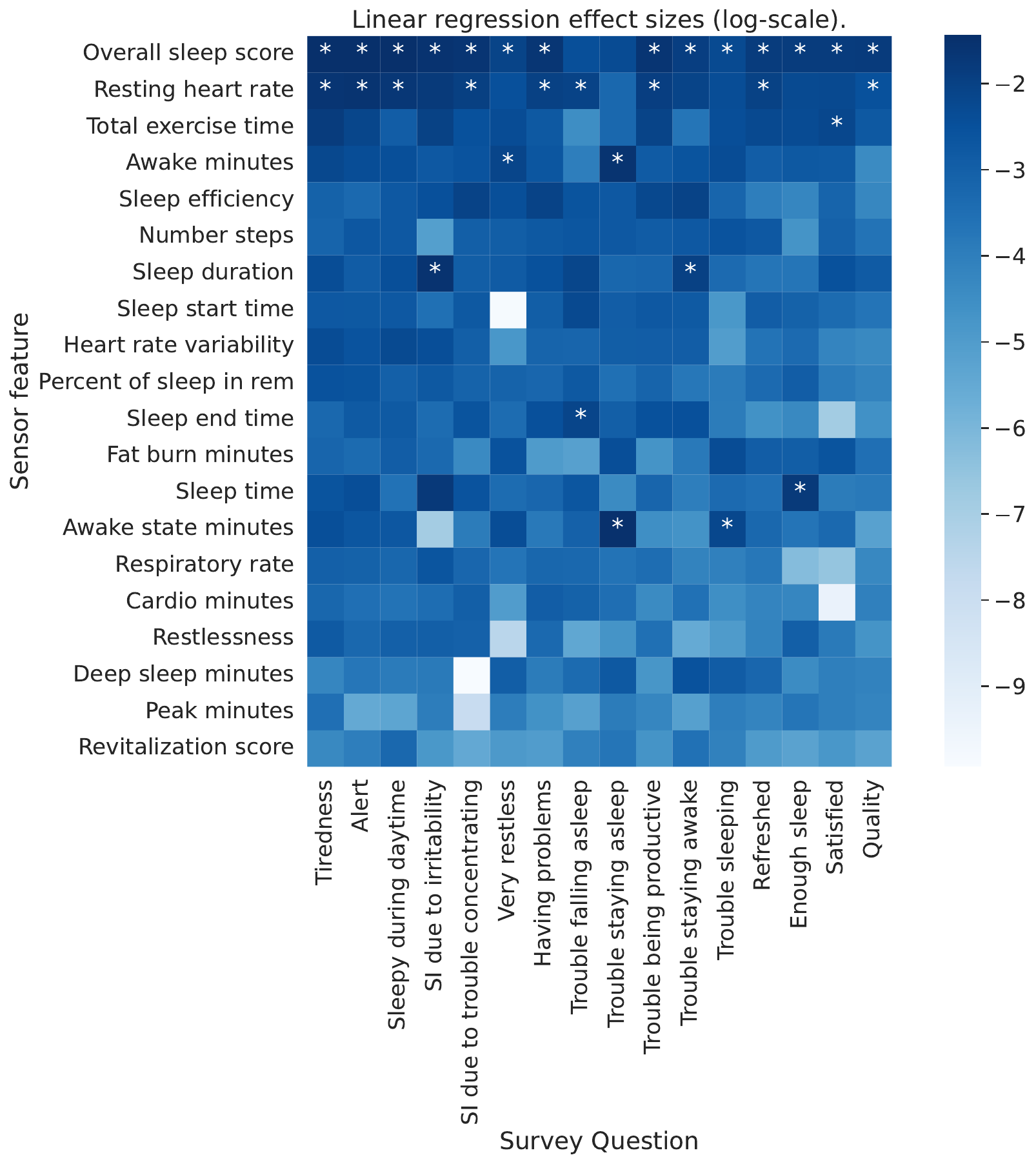}
         \label{fig:feature_importance}
     \end{subfigure}
     \par
     \begin{subfigure}[b]{0.48\textwidth}
         \caption{}
         \centering
         \includegraphics[width=\textwidth]{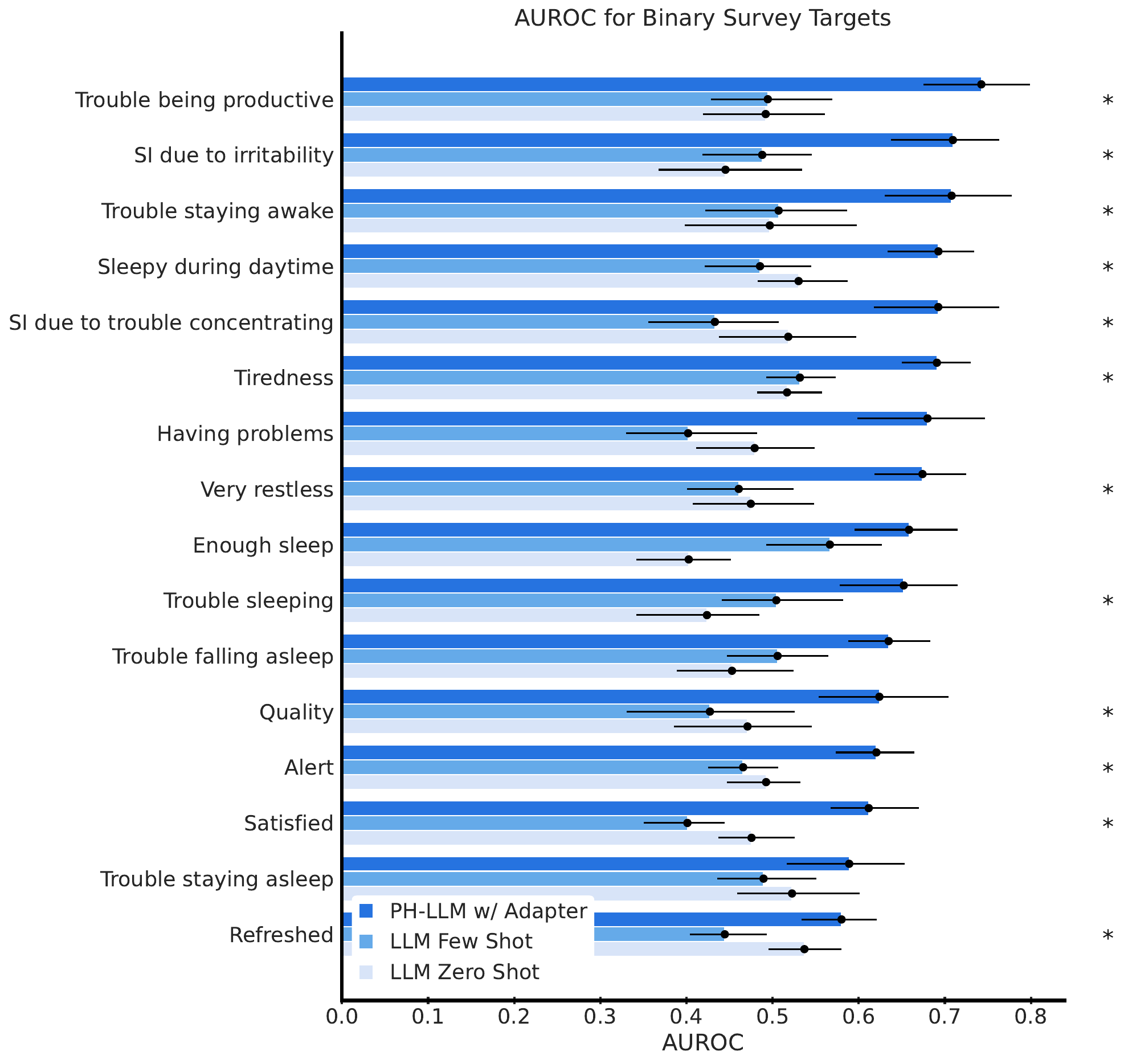}
         \label{fig:pro_prediction_auroc}
     \end{subfigure}
     \hfill
     \begin{subfigure}[b]{0.51\textwidth}
         \caption{}
         \centering
         \includegraphics[width=\textwidth]{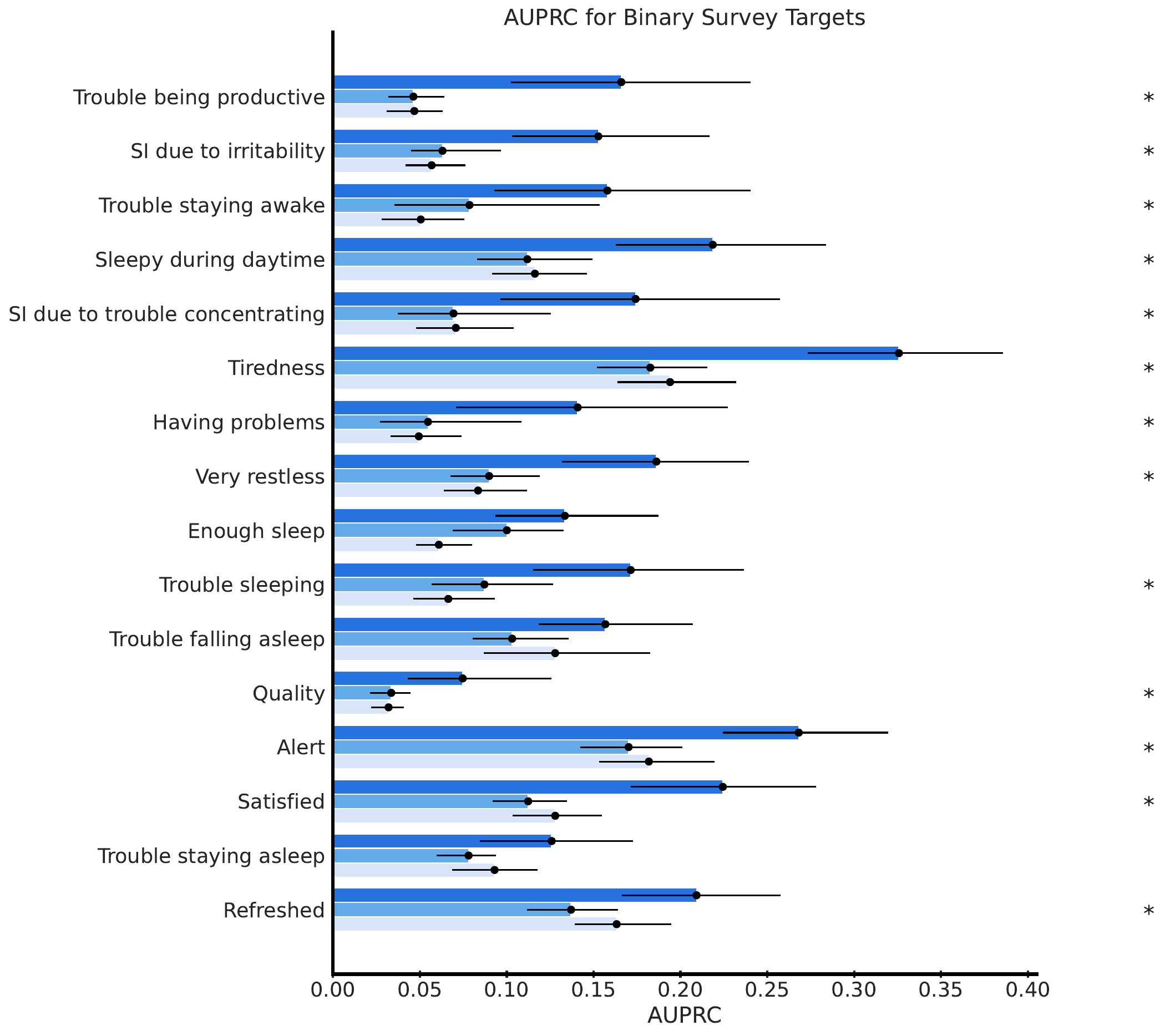}
         \label{fig:pro_prediction_auprc}
     \end{subfigure}
     
    \caption{\textbf{Prediction of Patient-Reported Outcomes by \ourmodel.}
     \textbf{(A)} Correlation between survey responses for questions that measure related but distinct sleep outcomes from the PROMIS Sleep Disturbance and Sleep Impairment surveys.
     \textbf{(B)} Feature importance for sensor features predicting survey responses in a linear regression model. The top two predictors for each target, measured based on the magnitude of the regression coefficient, are annotated with ``*''.
     \textbf{(C)} Area under the receiver operator characteristic curve for performance of \ourmodel, zero-shot, and few-shot prompting approaches when predicting binary outcomes derived from survey responses. Cases where performance of PH-LLM w/ Adapter is significantly better than both Zero and Few Shot are annotated with ``*''.
     \textbf{(D)} Area under the precision-recall curve for performance of \ourmodel, zero-shot, and few-shot prompting approaches when predicting binary outcomes derived from survey responses. Survey response names are mapped to their corresponding questions in~\Cref{sec:pro_surveys}. ``SI'', sleep impairment.
    }
 \hfill
\end{figure}

We evaluated the ability of \ourmodel to predict self-reported outcomes in sleep disturbance and sleep impairment. Using a dataset of 4,759 individuals with 20 wearable device measurements, a subset of the Google Digital Wellbeing Study dataset~\cite{mcduff2024google}, and 16 derived binary outcomes, we trained a multilayer perceptron (MLP) adapter to map the 20 wearable measurements into \ourmodel's latent token space (Section \ref{methods:pros}). We then provided the latent tokens to \ourmodel as context and prompted it to predict each patient outcome. Given the trained MLP adapter, we evaluated the likelihood of each binary outcome for each sample in the evaluation set and compared its predictive power to baseline approaches using in-context learning of textual sensor data representations.

We evaluated the correlation between the survey answers obtained from research participants and observed that the questions measure related but distinct sleep outcomes (Figure~\ref{fig:survey_response_correlations}). We then computed individual sensor feature importance for predicting PROs by fitting linear models taking all sensors as input to predict each survey question independently and assessing importance as the magnitude of the absolute value of the regression coefficient. We observed that there is no single feature that dominated predictive power, but rather that the overall predictive signal was spread across many sensors for the outcomes of interest (Figure~\ref{fig:feature_importance}). However, we note that interpreting the relationships between sensor features and outcomes is complicated because sensor features may capture demographic effects such as those from gender differences.

We then compared the area under the receiver operator characteristic curve (AUROC, Figure~\ref{fig:pro_prediction_auroc}) and area under the precision-recall curve (AUPRC, Figure~\ref{fig:pro_prediction_auprc}) for each binary trait, computed in the holdout set, for \ourmodel using the multimodal adapter and the zero- and few-shot text approches. We note that in general, objective measurements of sleep and sleep behaviors provide only modest predictive power for perceived sleep quality metrics. However, \ourmodel using the adapter significantly outperformed both prompt-based approaches in terms of both AUROC and AUPRC for 12 of the 16 traits (Supplementary Tables~\ref{table:supp:pro_roauc_diff} and~\ref{table:supp:pro_roauprc_diff}). This relative performance increase is due to adapter-enabled LLMs being able to capture more signal from the training set as compared to zero and few-shot prompting which see a very limited amount of training data~\cite{belyaeva2023multimodal}.

We do not expect an adapter-enabled language model to exceed the performance of a specialized discriminative model trained to predict the same binary traits. However, if the LLM has roughly comparable ability to a specialized model, this could be beneficial. To assess how well \ourmodel performed compared to a traditional machine learning approach, we fit a logistic regression (LR) model for each binary trait.
Supplementary Figures~\ref{fig:supp:pro_prediction_auroc} and~\ref{fig:supp:pro_prediction_auprc} compare \ourmodel to LR models trained using the same encoded vector input. We found no statistically significant differences in performance between \ourmodel and specialized LR models for either AUROC or AUPRC (Supplementary Tables~\ref{table:supp:pro_roauc_diff} and \ref{table:supp:pro_roauprc_diff}).

We also explored using convolutional neural nets on the time-series sensor data to predict outcomes, to model non-linear effects not modeled by logistic regression. However, these models underperformed the logistic regression model (data not shown). This is likely due to the limited size of the existing dataset.

\subsection{Qualitative analysis identifies areas for \ourmodel improvement}

In the following section we describe our qualitative results that provide insight into how the sleep and fitness experts viewed case studies. We identified several key themes in the experts' responses and present illustrative quotes below (S=Sleep expert, F=Fitness Expert). 

\textbf{Context Matters and is Difficult to Glean from Personal Health Data Alone.} 

Both in constructing and evaluating case studies several clinicians expressed how difficult it was to be precise and confident about the content when they did not have the full context of the individual's situation, e.g., lifestyle choices, job, family situation, etc. For the sleep case studies, the experts found it hard to identify which etiology was relevant in the absence of sufficient context.

A sleep expert summarized the following: [In generating recommendations from incomplete data]
\textit{"you are solving a problem that is vague... If it's somebody who's having difficulty falling asleep because they're napping will then they [sic] have insufficient sleep drive at the time of their primary sleep period. Conversely, if it's somebody who has insufficient sleep and therefore needs to nap, they have a high sleep drive ... it's hard to identify how to characterize the etiology correctly [from the data given]."} - S1

For certain data, only having access to a limited amount of history made the task hard - \textit{"[You] have to assume that was the only 30 days they did. If you had 6 months or 12 months it would provide more context."} - F1

The task of drawing insights, and generating recommendations, from wearable data was different to the general practice that the experts engaged in in their work. However, they highlighted that while it was not easy, they felt it was a valuable complement.

\textbf{Recommendations were the Hardest to Write and Evaluate.}

Related to the challenges of interpreting the data without context, the Recommendations section of the case study was universally described as the most challenging part to write. Identifying insights, while not simple, was reasonable. However, when it came to recommending behaviors, the experts described how they were not sure what would be realistic or achievable for the users without more context. For example, it would be unreasonable to recommend that a shift worker maintains a very regular bedtime. 

A specific example provided by one clinician was: \textit{"With my CBT-i hat on...I sort of compressed people if they had increased sleep efficiency, but in reality those are very personalized recommendations, and the model never got it. But whatever I put I'm sure somebody else [could say] ``I don't think that's good either.''"} - S2

These comments support the quantitative results from our coaching recommendation evaluations. The insight sections showed the largest overall gains from fine-tuning on the experts' case studies. 

\textbf{Model-Generated Responses were Inconsistent Across Sections.}

The model-generated responses were not always consistent from section to section.  A lack of internal consistency in certain case studies led to overall scores that were lower, despite individual sections scoring well. \textit{"There were a couple times with the cases where I got excited because [the section] was looking better...then I get to the next section and I was like scratch that not looking better [because of a contradiction]."} - F5

\textbf{Differences in Confabulations were Noticeable.}

Several experts described noticing conspicuous differences in confabulations across the case studies.  \textit{"In some cases it happens that you have to go through and if it's referencing a certain day, there'll be a [confabulation] or something like that. But I think ... more accurate and structured."} - F2. However, the different case study response sources (human expert, \basemodel, \ourmodel) did not show significant differences in the rate of confabulations, suggesting that confabulations were a function of specific case studies rather than response source.

\textbf{Some Mistakes Could be Harmful.}

The experts noted that the LLM was conservative or cautious in its responses (\textit{"It's being very cautious of things"} - F3) 

In fitness case studies, while the model was more sensitive to over-training, the experts highlighted when it failed to identify under-sleeping as a potential cause of harm. \textit{"I think most people were under sleeping. And so that's where you would want to focus on things." - S3}

Another expert noted detraining, defined as the partial or complete loss of fitness gains due to a reduction in training frequency, volume, or intensity, in a small number of case studies in which increases in resting HR were not picked up on and that the person could be at risk.  \textit{"The model did a worse job of calling out detraining [some responses] had done a better job."} - F4

\section{Related Work}

\subsection{Large language models in health}

LLMs have the ability to perform complex language comprehension and reasoning tasks, generate coherent text and thereby enable real-world applications~\cite{team2023gemini, openai2023gpt4,anil2023palm,scao2022bloom,touvron2023llama}. Explorations of LLM utility in health domains have shown their ability to answer medical questions and enable data-driven decision making~\cite{gu2021domain,singhal2023large,singhal2023towards,nori2023capabilities,saab2024capabilities,yang2024advancing}. Med-PaLM~\cite{singhal2023large} and its successor, Med-PaLM 2~\cite{singhal2023towards}, leveraged a combination of methodological advancements and domain-specific fine-tuning to increase performance, relative to previous models, on medically relevant evaluation tasks. Med-PaLM 2 achieved a score of up to 86.5\% across several medical datasets, such as MedMCQA, PubMedQA, and MMLU clinical topics, achieving physician-level performance. GPT-4 and Gemini have further improved performance on the USMLE-style examinations in MedQA, reaching 90.2\%~\cite{nori2023can} and 91.1\%~\cite{saab2024capabilities}, respectively. On complex diagnostic tasks it is even possible for LLMs to outperform clinicians (as in the case of medical internists constructing differential diagnoses~\cite{mcduff2023towards}). However, while models such as Med-PaLM 2, Med-Gemini \cite{saab2024capabilities,yang2024advancing}, GPT-4~\cite{nori2023can}, and Health-Alpaca \cite{kim2024health} excel at medical question answering and interpreting clinical data, their capabilities for interpreting personal health data is less well established.

Expanding LLMs to operate on modalities beyond just text has been a recent area of intense research, with prominent examples including but not limited to Flamingo~\cite{alayrac2022flamingo}, PaLI~\cite{chen2022pali}, GPT-4~\cite{openai2023gpt4}, GPT-4v~\cite{openai2024gpt4v}, Gemini 1.0~\cite{team2023gemini}, and Gemini 1.5~\cite{reid2024gemini}. The exploration of multimodal LLMs has also been extended to biomedical applications. Many models explore pairing one or multiple medical imaging modalities with language, including Med-Flamingo~\cite{moor2023med}, LLaVA-Med~\cite{li2024llavamed}, BiomedCLIP~\cite{zhang2023biomedclip}, MedBLIP~\cite{chen2023medblip}, ELIXR~\cite{xu2023elixr}, and others reviewed in further detail elsewhere~\cite{xiao2024comprehensive}. Other models explore support for non-imaging medical modalities, including HeLM~\cite{belyaeva2023multimodal}, Med-PaLM M~\cite{tu2023towards}, and Med-Gemini~\cite{saab2024capabilities,yang2024advancing}. While many of the earlier works focused primarily on medical question answering, there is increasing focus on report generation and other long-form responses.

Evaluation of long-form text is challenging~\cite{liang2022holistic} but is critical to ensure practical utility of LLMs in realistic settings. Similar to our efforts to generate realistic case studies of personal health coaching scenarios, MedAlign introduced a dataset for evaluating LLMs on relevant clinical tasks~\cite{fleming2024medalign} and demonstrated frequent misalignment between question answering performance and realistic task performance.

\subsection{Discriminative and Generative Personal Health}

Wearable sensors can help people realize meaningful changes in their health, such as helping to increase the amount of physical activity they engage in~\cite{ferguson2022effectiveness}. Moreover, when done thoughtfully and in an evidence-based manner, it is generally accepted that helping individuals derive insights from their data could increase the frequency of engaging in beneficial health behaviors. 
In the field of mobile health research \cite{steinhubl2015emerging, guillodo2020clinical}, traditional methodologies have predominantly centered around specialized, predictive models for defined classification tasks, such as predicting heart rate\cite{parak2014evaluation}, energy expenditure \cite{fuller2020reliability}, blood pressure \cite{arakawa2018recent}, and other vital signs, or classifying diseases using machine learning models tailored to specific predictive purposes such as atrial fibrillation detection \cite{nemati2016monitoring} and improving objective rehabilitation monitoring \cite{bonato2005advances}. More recently, LLMs have been shown be an effective base model to ground physiological and behavior time-series data and make meaningful inferences with zero-shot inference and few-shot learning across a wide variety of personal health tasks \cite{liu2023large,kim2024health}. 
In general, these methods use textual representations of sensor data to inform health metrics or predict health states. In contrast, our work with \ourmodel employs generative AI to expand model utility from only predicting health states to also providing coherent, contextual, and potentially prescriptive outputs that depend on complex health behaviors. While traditional models operate within the confines of specific, often binary or multinomial, outcome prediction, \ourmodel seeks to interpret and generate recommendations based on health behaviors, providing a more interactive and interpretive utility. This evolution from predictive modeling to generative reasoning set out our contribution in bridging quantitative data interpretation with qualitative, contextually-rich output, facilitating a better experience of digital health interaction and personal health data utilization.
\section{Discussion}

We developed an LLM fine-tuned from Gemini (\ourmodel) to perform a variety of tasks relevant to setting and achieving individual personal health goals. Our study shows that \ourmodel is capable of integrating passively-acquired objective data from wearable devices into personalized insights, potential causes for observed behaviors, and recommendations to improve sleep hygiene and fitness outcomes. After fine-tuning from the highly capable \basemodel, which already displays aggregate performance approaching that of experts in fitness, \ourmodel demonstrated significantly improved use of domain knowledge and personalization of relevant user information for sleep insights. Consistent with its strong performance on those long-form case studies, we showed that \ourmodel can accurately answer technical multiple choice questions in the domains of sleep and fitness, and contextualize the benchmark performance of \ourmodel in these datasets with performance of multiple experts in the same tasks. Finally, we demonstrated the ability of \ourmodel to use a multimodal encoder that natively integrates high-resolution time-series health behavior data as input tokens to predict subjective outcomes in sleep with performance on par with specialized models to predict the same outcomes.

Open-ended long-form case studies, while challenging to evaluate in an automated fashion, represent key use cases for applications of LLMs to personal health features on wearable devices. Here we created \numcasestudies case studies drawn from a pool of consented research participants to assess sleep quality and fitness readiness for a workout, and coupled the case studies with rigorous evaluation rubrics. We observed that the average performance over all case study responses was very high for all of human experts, \basemodel, and \ourmodel, underscoring the strong knowledge and reasoning capabilities of the Gemini model family. The improvements of \ourmodel for predicting sleep insight and etiology sections of case studies, and drawing from relevant user and domain knowledge, likely stem from the improved contextualization of relevant sleep features for these tasks. The relatively lower performance gain observed for sleep recommendations may be driven in part by both the increased ambiguity of the task, which led to wider variety in training example content, and the increased challenge of requiring consistency across multiple model generations to provide an accurate recommendation that referenced real data.

The reduced performance of \ourmodel compared to \basemodel and human experts in the training load section of fitness case studies may be partially explained by the data generation process, in which multiple models were used to create candidate case study responses. Expert rater comments suggest that some lower quality statements with incorrect use of domain and user data may be present in the training data, particularly with respect to rest periods (not shown). Moreover, since the fitness case studies incorporate sleep quality as one input, there is potential for further improvement within the fitness vertical by integrating full sleep case studies into the sleep section of fitness for a more detailed view of an individual’s rest status. Further exploration of these hypotheses is an area of active ongoing investigation.

Additionally, we developed methods for automated evaluation of case studies and demonstrated their ability to be used as scalable proxy measures for expert human evaluation of LLM performance, which is useful for model optimization. We showed that our best AutoEval models ranked study response sources similarly to human experts and that these models obtained similar agreement measures with expert raters when compared with inter-rater concordance metrics. By parallelizing automatic evaluation across model replicas, we observed a significant speed up compared to humans with respect to rating time.

Subjective PROs are receiving greater attention in health management~\cite{marshall2006impact,dobrozsi2015patient,lavallee2016incorporating}. Understanding the subjective experience of an individual can be useful for determining a comprehensive and personalized action plan. However, predicting subjective PROs from sensor information alone is difficult, with specialized models developed specifically for those tasks only reaching AUCs in the 0.55-0.75 range. We note that predictive power can be affected by the choice of binarization threshold for low prevalence outcomes. By only considering the binarization of the most extreme positive value against the rest of the outcomes we may underestimate the ability to stratify PROs across samples; however, the appropriate threshold choice is dependent upon context and the intended application of the prediction model. Overall performance notwithstanding, here we showed that \ourmodel could predict subjective PROs from sensor information alone with performance on par with that of specialized discriminative models, and leveraging the multimodal capabilities of Gemini was critical to achieving this performance. While the relative merits of broad generalist models compared to specialized models is an ongoing debate~\cite{yossigregblogpost}, enabling \ourmodel to predict subjective wellbeing opens the possibility of integrating this capability and information directly into user-facing interactions. Additionally, by building the predictive capability directly into \ourmodel, we might expect to benefit from positive transfer learning, in which the model is able to infer additional out-of-distribution outcomes for which it was not specifically trained, as shown previously~\cite{yang2024advancing}.

Our work has several limitations. First, the distribution of case study rubric ratings were skewed quite high, making differentiation across models and expert responses challenging. While some case study sections and evaluation rubric principles did show significant differentiation, further training of expert raters to increase inter-rater reliability or adjudicating existing responses could increase signal strength of model performance. Second, owing to inter-rater variability, we chose to have each expert rate all responses for a given case study. While this made direct comparison of candidate responses straightforward, it introduced the potential for experts to identify expert vs model responses based on style or other non-material factors, and thus introduce conscious or unconscious biases into ratings. Third, we observed that despite improvements in referencing and integrating user data into insights, confabulations or incorrect referencing of user data still occasionally occurred. Addressing and preventing these issues will be critical to ensure the safe and effective deployment of these technologies into user-facing features. Promising progress is being made through active research on agentic workflows that critique and correct candidate responses~\cite{merrill2024transforming}. Fourth, the case studies were sampled broadly across demographics (sleep) or to identify common patterns in active individuals (fitness), but may not be a representative sample of the population nor exhaustively explore the sleep and fitness concerns affecting individuals.
Fifth, our exploration of multimodal encoding of sensor data explored a small fraction of the design space owing to the relatively small dataset with paired outcome data and our purposeful restriction to samples with nearly complete sensor data. Further exploration of self-supervised pre-training on raw waveforms and granularly aggregated sensor features may yield richer representations of individuals that can be effectively purposed toward personal health outcome predictions \cite{abbaspourazad2023large} that expand beyond just sleep metrics and address challenges arising from a sparse and heterogeneous mix of available sensor features. We anticipate that future large datasets with paired outcome data will enable non-linear interactions across features to be learned effectively to improve predictive power. Sixth, a primary overarching goal for developing models specific to personal health is to be able to improve long-term health outcomes through effective behavior change and maintenance of healthy habits. Neither of these tasks is explicitly evaluated here, and remain important areas for future work. Finally, while the performance of \ourmodel on the tasks presented here is encouraging, we caution that much work remains to be done to ensure LLMs are reliable, safe, and equitable in personal health applications. Further reducing confabulations, considering an individual's unique health circumstances not captured by sensor information alone, and ensuring alignment of the training data with real-world distributions are a subset of important research areas that warrant further attention.

Despite the above limitations, we have demonstrated here that the Gemini family of models are imbued with substantial health knowledge, and we can effectively fine-tune \basemodel to improve performance across multiple outcomes relevant for personal health. The results from this study represent an important step toward LLMs that deliver personalized information and recommendations that support individuals to achieve their health goals. 
\subsubsection*{Acknowledgements}
We thank the Fitbit research community participants for making this research possible. We thank the sleep and fitness experts who developed case study responses and evaluated candidate model responses for their dedication, effort, and detailed feedback on multiple model iterations. Contributing sleep experts include Ben Graef, Timothy Wong, Thuan Dang, Suzanne Gorovoy, Narayan Krishnamurthy, and Michelle Jonelis. Contributing fitness experts include Jarod Spraggins, Allison Hetrick, Jonas Hannon, Max Knight, Nolan Dozier, Laura Grissom, and Justin Leach. We thank Hulya Emir-Farinas, Farhad Hormozdiari, and Joëlle Barral for feedback and discussions that significantly improved the work. We also thank Sami Lachgar, Lauren Winer, Maggie Shiels, Lee Gardner, Noa Tal, Annisah Um’rani, Oba Adewunmi, and Archit Mathur for their valuable insights, technical support, and feedback during our research. 

\subsubsection*{Data Availability}
Case study inputs and associated expert and model responses for the holdout evaluation set will be made publicly available upon peer-reviewed publication.

\subsubsection*{Competing Interests}
This study was funded by Google LLC. All authors are employees of Alphabet and may own stock as part of the standard compensation package.

\vspace{-0.5cm}

\newpage
\bibliographystyle{abbrv}
\bibliography{main}

\newpage
\renewcommand{\thetable}{\thesection.\arabic{table}}
\renewcommand{\thefigure}{\thesection.\arabic{figure}}
\appendix
\label{sec:appendix}
\newpage

\renewcommand \thepart{}
\renewcommand \partname{}
\renewcommand{\ptcfont}{\fontfamily{cmss}}
\renewcommand{\ptcSfont}{\normalsize\fontfamily{cmss}} 
\renewcommand{\ptcSSfont}{\normalsize\fontfamily{cmss}} 

\addcontentsline{toc}{section}{Appendix} 
\part{Appendices} 
\parttoc 

\newpage
\setcounter{table}{0}
\setcounter{figure}{0}
\newpage
\section{Coaching Recommendations Case Studies}

\subsection{Case study dataset creation}

\subsubsection{Additional details on creation of sleep case studies}
\label{apendix: additional_sleep_case_study}

Our study utilized de-identified data from individuals who provided consent for use of their data for research purposes. For sleep case studies, in order to ensure a representative sample across different demographics (age and gender), we considered 64 different demographic groups, determined by a combination of 32 different age buckets (13-20 years old, 20-80 years old with each group within this range spanning two years, 80 years old and above) and 2 gender buckets (male, female).

\textbf{Daily Sleep Metrics:} The daily sleep metrics contain up to 29 days of daily sleep metrics. The metrics are: date, day of the week, sleep score (0-100), light sleep (hh:mm), REM sleep (hh:mm), deep sleep (hh:mm), sleep duration (hh:mm), fall asleep time, wake time after sleep onset (hh:mm), sleep efficiency, fraction of sleep goal, number of times the individual woke up, heart rate (bpm), nap duration (min), number of naps, and wake up time. See Table \ref{table:prompt_sleep_case_studies_sleep_logs} for an example.

\textbf{Aggregated Daily Sleep Statistics:} Generally, these statistics included an aggregated metric (e.g. average, median, standard deviation, count) over all the days, the percentile that the aggregated metric is in as compared to other individuals within the same demographic group, minimum value over all the days, maximum value over all the days, as well as 5th and 95th percentiles of the aggregated metric as compared to other individuals within the same demographic group. In some instances, such as bedtime, the metrics were computed separately for all days, weekdays only, and weekends only to understand weekday versus weekend patterns. See Table \ref{table:prompt_sleep_case_studies_sleep_summary} for an example.

\subsubsection{Additional details on creation of fitness case studies}
\label{apendix: additional_fitness_case_study}

For fitness case studies, the individuals from de-identified cohort who provided consent for use of their data for research purposes, were sampled. In order to ensure the fitness case studies contain sufficient activity for interesting training readiness analysis, we sampled individuals who had data for at least 16 days with minimum mean active zone minutes of 45 minutes and with at least 2 logged exercises. In addition, we considered periods of days that contained noticeable changes in heart rate variability, resting heart rate, respiratory rate, sleep, and periods with runs. The experts considered the following data in their analysis:

\textbf{Demographics:} (age, gender, height, weight, body mass index). The experts considered the demographics data and commented on whether any precautions should be taken when recommending a fitness program.

\textbf{Training Load:} The experts were provided with a detailed table capturing daily metrics over the past 30 days, including day of the week, date, minutes spent in fat-burn, cardio, and peak zones, training impulse (TRIMP), and number of steps (Table \ref{table:prompt_fitness_case_studies_daily_activity}). Additionally, we provided aggregated statistical analyses such as means, ranges, acute TRIMP (7-day total training load), chronic TRIMP (28-day average acute training load), Acute-Chronic Workload Ratio (ACWR), and metrics specific to each exercise entry (Table \ref{table:prompt_fitness_case_studies_agg_daily_metrics}). For ease of analysis, in addition to the table, daily TRIMP values were visualized in a barplot (Figure \ref{fig:casestudyexample_fitness}A).

\textbf{Sleep Metrics:} The experts assessed the individual's sleep as it relates to fitness recovery. A table of daily sleep metrics such as bedtime, wake time, sleep time, awake time, deep sleep, REM sleep, and sleep score was given to the experts for analysis (Table \ref{table:prompt_fitness_case_studies_sleep_metrics}). For ease of analysis, some of the daily sleep metrics were also visualized as a graph (Figure \ref{fig:casestudyexample_fitness}B) They were also given aggregated metrics including means, standard deviations, and z-scores indicating the difference in metrics between the most recent 3 days and the past 28 days to identify recent trends.

\textbf{Health Metrics:} A table and a graph of daily resting heart rate, heart rate variability (HRV), and respiratory rate over the past 30 days was given to the experts to assess recovery and stress (Figure \ref{fig:casestudyexample_fitness}C, Table \ref{table:prompt_fitness_case_studies_health_metrics_month}). The experts were also given aggregate metrics such as means, standard deviations, ranges, and z-scores indicating the difference in metrics between the most recent day and the past 28 days (Table \ref{table:prompt_fitness_case_studies_agg_health_metrics}).

To simulate feedback from the user about their subjective state, the experts were also given synthetically (LLM) generated subjective readiness to workout (e.g. "feeling fatigued") and muscle soreness (e.g. "manageable soreness"). For examples, see Tables \ref{table:prompt_fitness_case_studies_subjective_readiness} and \ref{table:prompt_fitness_case_studies_muscle_soreness}.

\begin{figure*}[thp]
  \centering
    \includegraphics[width=\textwidth]{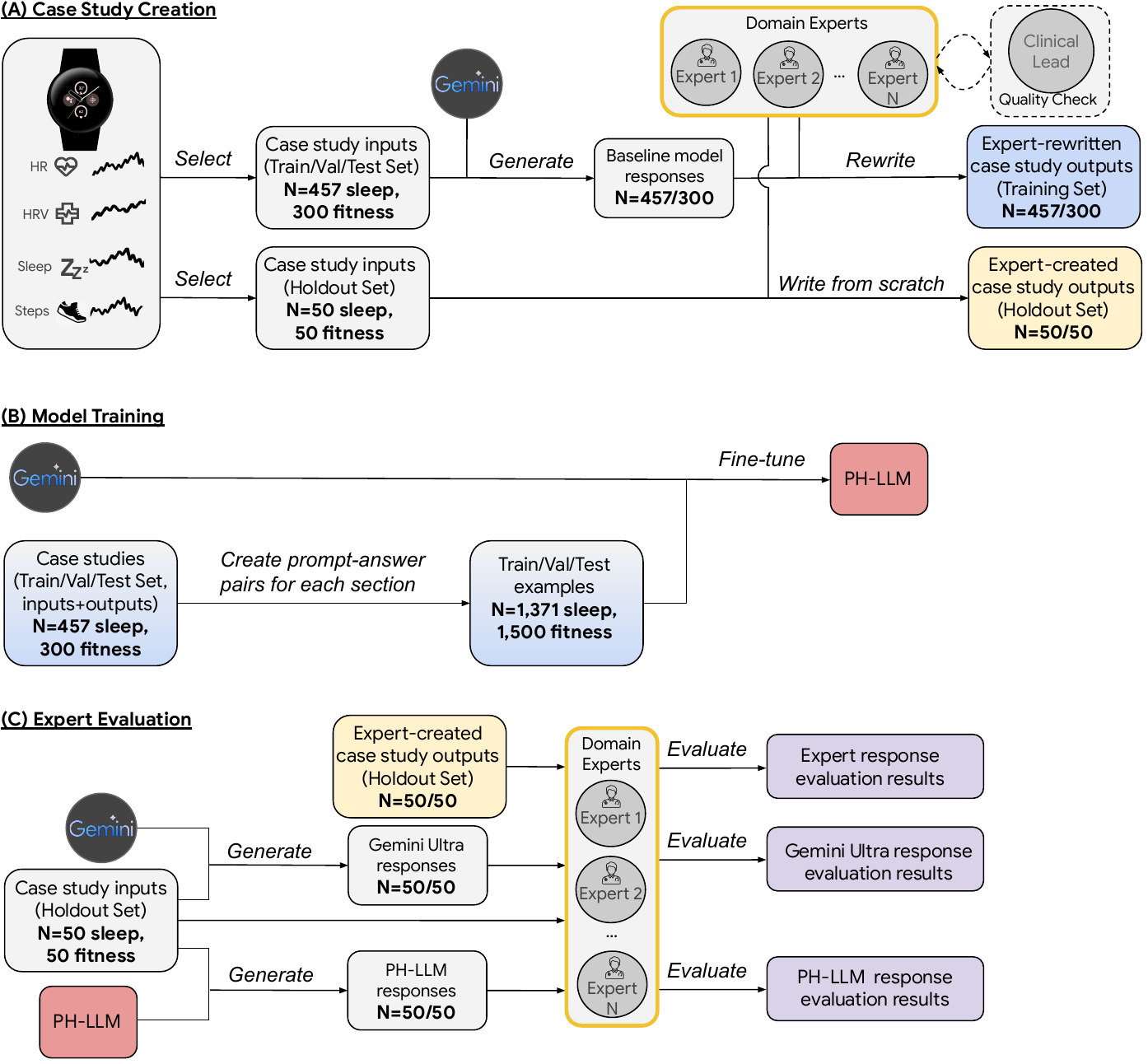}
    \caption{\textbf{Case Study Creation, Curation, and Evaluation Workflow.} Case studies were selected from a large set of anonymized, consented production data. \textbf{(A)} Two sets of case studies were generated. To facilitate rapid development of high-quality answers, the train/validation/test set of case studies had candidate responses generated by Gemini, which were then edited and rewritten by domain experts. To enable comparison of human and model-derived responses, the holdout set had responses written solely by the domain experts. \textbf{(B)} For model training, each case study was split into multiple prompt/answer pairs based on how many sections the case study had (N=3 for sleep with insights, etiology, and recommendations sections, N=5 for fitness with demographics, training load, sleep metrics, health metrics, and assessment sections, see Section~\ref{section:methods:phllm_training_text} for details) and \basemodel underwent full fine-tuning using those examples. \textbf{(C)} Expert evaluation was performed independently on the holdout dataset by the same set of domain experts responsible for generating the expert responses. For each case study in the holdout set, an expert who did not write the corresponding expert response graded all three candidate responses (expert-written response, \basemodel response, \ourmodel response).}
    \label{figure:casestudyworkflow}
  \end{figure*}

\begin{figure*}[htp]
    \centering
    \includegraphics[width=\textwidth]{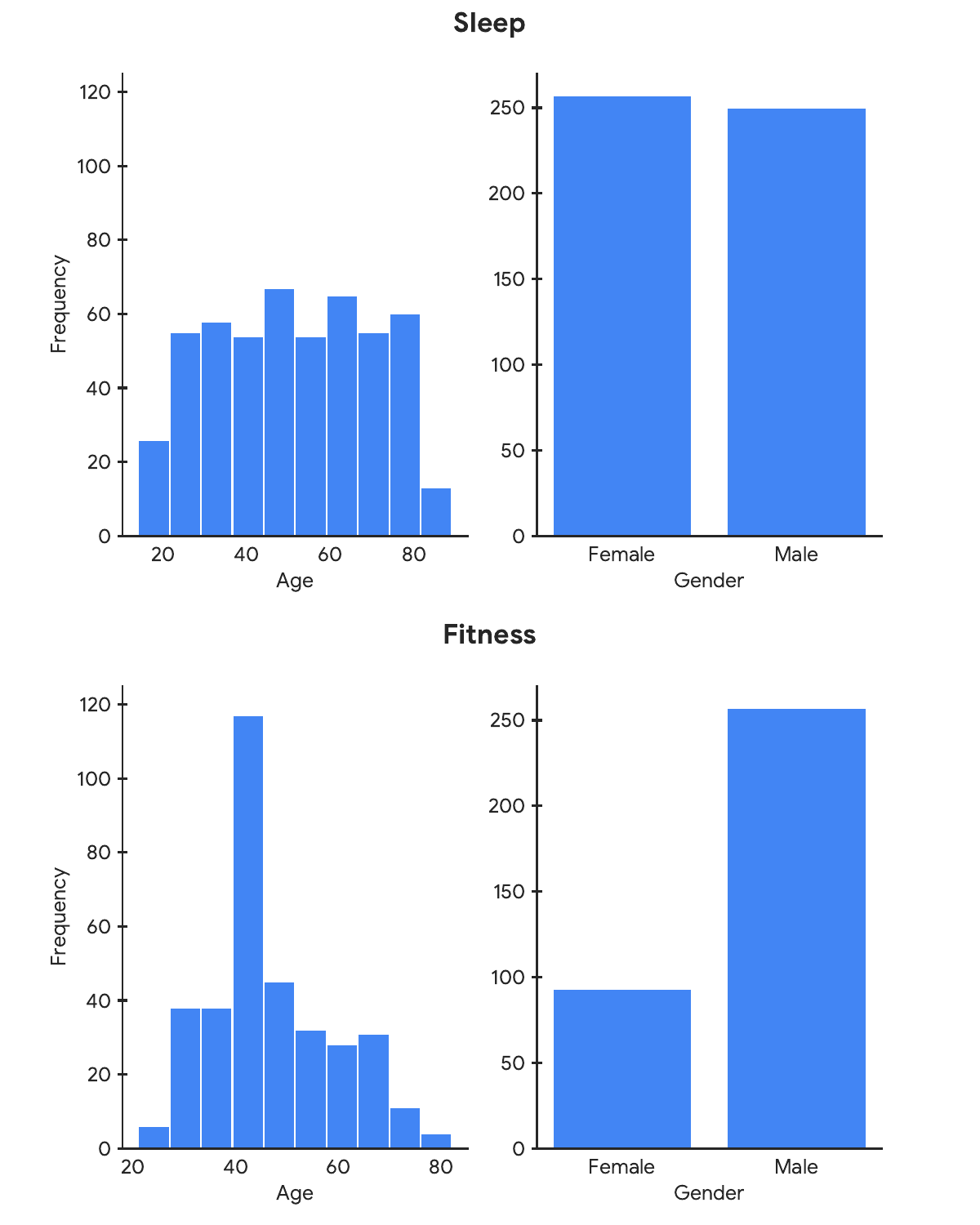}
    \caption{\textbf{Distribution of case study ages and genders across the sleep and fitness verticals.}}
    \label{figure:case_study_stats}
\end{figure*}

\newpage
\begin{table}[htp]
\captionsetup{labelformat=simple, justification=raggedright, singlelinecheck=false, labelsep=pipe} 
\caption{\textbf{Prompt for sleep case studies insights section.}}
\label{table:prompt_sleep_case_studies_insights} 
\end{table}

\begin{tcolorbox}[title={Prompt for sleep case studies insights section}]

You are a sleep medicine expert. You are given the following sleep data.\newline
The user is <gender>, <age> years old.\newline
\textbf{Sleep logs:}\newline
<sleep logs table (see Table \ref{table:prompt_sleep_case_studies_sleep_logs} for an example)>
\newline
\newline
\textbf{Sleep Summary}:
\newline
<sleep summary (see Table \ref{table:prompt_sleep_case_studies_sleep_summary} for an example)>
\newline

List the most important insights. Identify all of the patterns of data that are likely out of the preferred range. Make sure to consider various sleep health dimensions: Routine, Sleep Quality, Alertness, Timing, Efficiency, and Duration. Add a heading for each dimension. Optionally (only do this if extremely important) add a heading called Other for anything else that doesn’t fit the above categories. For Routine, consider the average bedtime, wake time, midsleep point and standard deviations of these, focus on the consistency of the routine, not timing. For Sleep Quality, consider light sleep duration, deep sleep duration, REM sleep duration,  sleep score, restlessness score, time to quality sleep, and wake time after sleep onset. For Alertness, consider the number of naps and nap length. For Timing, consider midsleep point, bedtime, wake time, make any comments on weekend vs. workday. For Efficiency, consider sleep efficiency, wake time after sleep onset, and time to quality sleep, describe how they compare to similar users. For Duration, consider average sleep duration, weekend vs. workday sleep durations and standard deviations, describe how they compare to similar users. When determining whether a metric is normal or abnormal, always provide the corresponding percentile. Avoid generic statements. Avoid incorrect knowledge, inconsistencies and contradictions. Don’t mention “the user”. Talk like you’re speaking directly to someone. Be concise.
\newline
\# Sleep insights report
\end{tcolorbox}

\newpage
\begin{table}[h]
\captionsetup{labelformat=simple, justification=raggedright, singlelinecheck=false, labelsep=pipe} 
\caption{\textbf{Prompt for sleep case studies etiology section.}}
\label{table:prompt_sleep_case_studies_etiology} 
\end{table}

\begin{tcolorbox}[title={Prompt for sleep case studies etiology section}]
You are a sleep medicine expert. You are given the following sleep data.\newline
The user is <gender>, <age> years old.\newline
\textbf{Sleep Summary}:
\newline
<sleep summary (see Table \ref{table:prompt_sleep_case_studies_sleep_summary} for an example)>
\newline
\newline
Based on the data, we can get the following insights:\newline
<insights response>\newline
What are the underlying causes? Make sure to consider the following causes: Circadian rhythm, Homeostatic drive, Psychophysiologic hyperarousal, and Extrinsic factors. Order the causes from most to least relevant. Identify the likelihood of the causes (e.g. unlikely, possible, very likely). Cite relevant data and insights, for example, “consistently low sleep efficiency despite normal sleep durations suggests low homeostatic drive”. Avoid diagnosing health conditions. Avoid providing recommendations. Avoid generic statements. Avoid incorrect knowledge, inconsistencies and contradictions. Don’t mention “the user”. Talk like you’re speaking directly to someone. Be concise.
\newline
\# Causes report
\end{tcolorbox}

\begin{table}[h]
\captionsetup{labelformat=simple, justification=raggedright, singlelinecheck=false, labelsep=pipe} 
\caption{\textbf{Prompt for sleep case studies recommendations section.}}
\label{table:prompt_sleep_case_studies_recs} 
\end{table}

\begin{tcolorbox}[title={Prompt for sleep case studies recommendations section}]
You are a sleep medicine expert. You are given the following sleep data.\newline
The user is <gender>, <age> years old.\newline
\textbf{Sleep Summary}:
\newline
<sleep summary (see Table \ref{table:prompt_sleep_case_studies_sleep_summary} for an example)>
\newline
\newline
Based on the data, we can get the following insights:\newline
<insights response>\newline
Causes:\newline
<etiology response>\newline
What recommendation(s) can you provide to help this user improve their sleep? Tie recommendations to the very likely and possible causes, for example, “Recommendations to address Circadian rhythm”. Tie recommendations to user’s sleep data such as average bedtime, average wake time, and number of naps, and recommend a goal bedtime and wake time based on their data. The recommendations should be time-bound, for example for the next week or the next month. Write one short question to ask the user in order to better understand their sleep. Avoid assumptions regarding the trainee's lifestyle or behavioral choices. Avoid generic statements. Avoid incorrect knowledge, inconsistencies and contradictions. Don’t mention “the user”. Talk like you’re speaking directly to someone. Be concise.
\newline
\# Recommendations report
\end{tcolorbox}

\newpage
\begin{table}[h]
\captionsetup{labelformat=simple, justification=raggedright, singlelinecheck=false, labelsep=pipe} 
\caption{\textbf{Abridged example of sleep logs table for a particular individual used in sleep case studies.} For brevity, only seven days are shown.}
\label{table:prompt_sleep_case_studies_sleep_logs} 
\end{table}
\begin{tcolorbox}[width=1.25\linewidth,center, ,title={Abridged example of sleep logs table for a particular individual used in sleep case studies}]
\fontsize{20pt}{20pt}\selectfont

\begin{adjustbox}{width=\linewidth,center}
\begin{tabular}{llllllllllllllll}
Date       & Day of Week & Sleep Score & Light Sleep (hh:mm) & REM Sleep (hh:mm) & Deep Sleep (hh:mm) & Sleep Duration (hh:mm) & Fall Asleep Time & Wake after Sleep Onset (hh:mm) & Efficiency & Fraction of Sleep Goal & Wakeup Count & Heart Rate (bpm) & Nap Duration (min) & Naps & Wake Time \\
<year-month-day> & Thursday    & 71.0                                                                                                                                        & 04:24      & 00:59         & 00:40        & 06:04            & 00:05              & 00:04          & 0.88 & 0.76 & 4.0  & 58.0 & 0.0   & 0.0 & 06:13 \\
<year-month-day> & Friday      & 72.0                                                                                                                                        & 03:13      & 01:07         & 01:03        & 05:24            & 00:38              & 00:08          & 0.85 & 0.68 & 8.0  & 58.0 & 88.0  & 1.0 & 06:10 \\
<year-month-day> & Saturday    & 87.0                                                                                                                                        & 05:08      & 01:51         & 02:00        & 09:00            & 03:02              & 00:10          & 0.87 & 1.12 & 9.0  & 58.0 & 0.0   & 0.0 & 12:12 \\
<year-month-day> & Sunday      & 83.0                                                                                                                                        & 05:16      & 01:49         & 01:41        & 08:47            & 03:54              & 00:15          & 0.86 & 1.10 & 15.0 & 58.0 & 0.0   & 0.0 & 12:56 \\
<year-month-day> & Monday      & 68.0                                                                                                                                        & 04:21      & 00:50         & 00:42        & 05:54            & 00:07              & 00:08          & 0.85 & 0.74 & 8.0  & 58.0 & 0.0   & 0.0 & 06:09 \\
<year-month-day> & Monday      & 64.0                                                                                                                                        & 01:29      & 00:27         & 00:51        & 02:48            & 16:10              & 00:05          & 0.85 & 0.35 & 5.0  & 58.0 & 0.0   & 0.0 & 19:03 \\
<year-month-day> & Tuesday     & 70.0                                                                                                                                        & 01:18      & 00:43         & 00:50        & 02:52            & 03:42              & 00:02          & 0.87 & 0.36 & 2.0  & 59.0 & 0.0   & 0.0 & 06:36 \\
<year-month-day> & Wednesday   & 72.0                                                                                                                                        & 03:19      & 01:14         & 01:02        & 05:36            & 00:17              & 00:09          & 0.83 & 0.70 & 9.0  & 58.0 & 0.0   & 0.0 & 06:02 \\
<year-month-day> & Wednesday   & 71.0                                                                                                                                        & 01:41      & 00:43         & 00:35        & 03:00            & 16:22              & 00:00          & 0.86 & 0.38 & 0.0  & 58.0 & 0.0   & 0.0 & 19:22
\end{tabular}
\end{adjustbox}
\end{tcolorbox}

\begin{table}[h]
\captionsetup{labelformat=simple, justification=raggedright, singlelinecheck=false, labelsep=pipe} 
\caption{\textbf{Abridged example of sleep summary for a particular individual used in sleep case studies.} Stratified features report overall statistics as well as stratified by workday vs weekend, and include bedtime, wake time, midsleep point, sleep duration, and sleep score. Unstratified features include time to quality sleep, wake time after sleep onset, sleep efficiency, light sleep duration, deep sleep duration, REM sleep duration, and restlessness score. Nap length and total number of naps are also reported.}
\label{table:prompt_sleep_case_studies_sleep_summary} 
\end{table}
\begin{tcolorbox}[title={Abridged example of sleep summary for a particular individual used in sleep case studies.}]
\fontsize{7pt}{7pt}\selectfont
Average bedtime is 00:26\newline
Average bedtime is in the 65th percentile\newline
Earliest bedtime is 16:10\newline
Latest bedtime is 06:22\newline
Bottom 5th percentile of similar users' average bedtimes is 21:25\newline
Top 95th percentile of similar users' average bedtimes is 03:07\newline

Bedtime standard deviation is 03:34\newline
Bedtime standard deviation is in the 94th percentile\newline

Average bedtime on the weekend is 01:35\newline
Average bedtime on the weekend is in the 72nd percentile\newline
Earliest bedtime on the weekend is 16:58\newline
Latest bedtime on the weekend is 06:22\newline
Bottom 5th percentile of similar users' average bedtimes on the weekend is 21:45\newline
Top 95th percentile of similar users' average bedtimes on the weekend is 03:28\newline

Bedtime standard deviation on the weekend is 03:46\newline
Bedtime standard deviation on the weekend is in the 92nd percentile\newline

Average bedtime on a workday is 23:58\newline
Average bedtime on a workday is in the 60th percentile\newline
Earliest bedtime on a workday is 16:10\newline
Latest bedtime on a workday is 06:22\newline
Bottom 5th percentile of similar users' average bedtimes on a workday is 21:10\newline
Top 95th percentile of similar users' average bedtimes on a workday is 03:14\newline

Bedtime standard deviation on a workday is 03:28\newline
Bedtime standard deviation on a workday is in the 94th percentile\newline

Median bedtime on a workday is 00:38\newline
Median bedtime on a workday is in the 73rd percentile\newline
Bottom 5th percentile of similar users' median bedtimes on a workday is 21:04\newline
Top 95th percentile of similar users' median bedtimes on a workday is 03:23\newline

...

Average time to quality sleep is 00:33\newline
Average time to quality sleep is in the 92nd percentile\newline
Shortest time to quality sleep is 00:04\newline
Longest time to quality sleep is 01:23\newline
Bottom 5th percentile of similar users' average times to quality sleep is 00:13\newline
Top 95th percentile of similar users' average times to quality sleep is 00:35\newline

...

Average nap length is 129\newline
Average nap length is in the 92nd percentile\newline

Total number of naps is 4
\end{tcolorbox}

\newpage
\begin{table}[h]
\captionsetup{labelformat=simple, justification=raggedright, singlelinecheck=false, labelsep=pipe} 
\caption{\textbf{Prompt for fitness case studies demographics section.}}
\label{table:prompt_fitness_case_studies_demographics} 
\end{table}

\begin{tcolorbox}[title={Prompt for fitness studies demographics section}]
You are a NSCA and ACSM board-certified fitness trainer who specializes in athlete training performance and recovery.\newline
Age: <age> \newline
Height: <height>\newline
Weight: <weight>\newline
BMI: <BMI>\newline
Gender: <gender>\newline\newline
Are there any special precautions that should be taken into account when recommending a fitness program to avoid injury? Comment if the trainee has exceptional demographics (e.g. very old, very high BMI, very low BMI) that require special considerations. Write a single sentence. Avoid mentioning diseases.
\end{tcolorbox}

\newpage
\begin{table}[h]
\captionsetup{labelformat=simple, justification=raggedright, singlelinecheck=false, labelsep=pipe} 
\caption{\textbf{Prompt for fitness case studies training load section.}}
\label{table:prompt_fitness_case_studies_training_load} 
\end{table}

\begin{tcolorbox}[title={Prompt for fitness case studies training load section}]
The following section shows some of the trainee's recent activity metrics including the active zone minutes: Fat burn zone (50\% heart rate reserve), Cardio zone (70\% heart rate reserve), and Peak zone (85\% heart rate reserve.)\newline
\textbf{Daily activity metrics:}\newline\newline
<table of daily activity metrics (see Table \ref{table:prompt_fitness_case_studies_daily_activity} for an example)>
\newline
Today is <day of the week> <year-month-day>.\newline\newline\newline
\textbf{Here are some aggregate statistics for the last 30 days:}\newline
<aggregate statistics of daily activity metrics (see Table \ref{table:prompt_fitness_case_studies_agg_daily_metrics} for an example)>\newline\newline
Analyze the trainee's recent activity metrics, aggregate statistics for the last 30 days, and most recent exercise logs. Assess the following: Training Load Trends, Intensity, Duration, Frequency, Rest Periods, Acute-Chronic-Workload Ratio (ACWR), Recent Activity Levels, and Significant Workouts. For Training Load Trends, consider mean moderate activity per day, mean vigorous activity per day, comment on balance between moderate and vigorous activity. For Intensity, consider the most recent exercise logs, assess time in fat-burn zone (moderate intensity), time in cardio zone (vigorous intensity), time in peak zone (peak intensity), and state whether the workouts overall reached each zone, consider the daily activity metrics and assess the TRIMP values. For Duration, consider the most recent exercise logs and list the lowest and highest duration as a range. For Frequency, consider the most recent exercise logs, and check on which days of the week there is a workout. For Rest Periods, consider the daily activity metrics table and see if some days have very low to zero TRIMP - these are also rest periods, comment on the number of rest days and which days of the week. For Acute-Chronic-Workload Ratio, consider acute TRIMP, chronic TRIMP, see if acute TRIMP is higher than chronic TRIMP and state what it means in terms of training load, consider Acute-Chronic Workload Ratio (ACWR) and state what it means for recovery. ACWR values above 1.5 reflect a significant increase in training load and may result in a higher risk of injury. ACWR values of less than 0.7 indicate that the trainee has had a significant decrease in training load and may be at risk of detraining. For Recent Activity Levels and Significant Workouts, consider the most recent exercise logs and note any recent significant workouts that are related to changes in the training load metrics, consider the daily activity metrics and highlight days with highest TRIMP and explain their importance.\newline\newline
Note: Remember to avoid readiness assessments, avoid recommendations, avoid making up data, and stay directly aligned with the provided data.\newline
- Base all observations and insights on the provided data.\newline
- Avoid generic advice.\newline
- Refrain from making up data or giving general advice not rooted in the data.\newline
- Avoid assumptions regarding the trainee's lifestyle or behavioral choices.\newline
- Do not elaborate on anything not contained within the data tables.\newline
- Do not compute or reference complex mathematical calculations like correlation coefficients.\newline
- When explaining the numerical difference, refrain from inventing any calculations if you are not certain about them.\newline
- Use markdown to structure the response.\newline
- Use an observation/insight format:\newline
* **Observation:** A factual observation from the data.\newline  
* **Insight:** The implication of the observation in the context of the user's health.\newline
- Group the observation/insights into appropriate sections.\newline
\# Training load report
\end{tcolorbox}

\newpage
\begin{table}[h]
\captionsetup{labelformat=simple, justification=raggedright, singlelinecheck=false, labelsep=pipe} 
\caption{\textbf{Prompt for fitness case studies sleep section.}}
\label{table:prompt_fitness_case_studies_sleep} 
\end{table}

\begin{tcolorbox}[title={Prompt for fitness case studies sleep section}]
These are the trainee's \textbf{recent sleep metrics:}\newline\newline
<table of sleep metrics for fitness case studies (see Table \ref{table:prompt_fitness_case_studies_sleep_metrics} for an example)>
\newline
Today is <day of the week> <year-month-day>.\newline\newline
\textbf{Here are some aggregate statistics for the last 30 days:}\newline
<aggregate statistics of sleep metrics (see Table \ref{table:prompt_fitness_case_studies_agg_sleep_metrics} for an example)>\newline\newline
- Assess the following aspects of trainee's sleep based on metrics:\newline  
* Sleep Schedule: bedtimes and wake-times\newline
* Sleep Duration: sleep duration metrics\newline
* Sleep Quality: sleep score. Excellent sleep score is 90 to 100. Good sleep score is 80 to 89. Fair sleep score is 60 to 79. Poor sleep score is less than 60.\newline
* Today’s Sleep: Comment on today's values and compare them to the aggregate statistics for the last 30 days. Make this comment only if sleep duration Z-score or sleep score Z-score is less than -2, comment that this indicates significantly worse recent sleep in the last 3 days compared to the monthly average sleep duration and low final readiness assessment is recommended . Make this comment only if sleep duration Z-score or sleep score Z-score is more than 2, comment that this indicates significantly improved recent sleep in the last 3 days compared to the monthly average sleep duration.\newline\newline
- Base all observations and insights on the provided data.\newline
- Avoid generic advice.\newline
- Refrain from making up data or giving general advice not rooted in the data.\newline
- Avoid assumptions regarding the trainee's lifestyle or behavioral choices.\newline
- Do not elaborate on anything not contained within the data tables.\newline
- Do not compute or reference complex mathematical calculations like correlation coefficients.\newline
- When explaining the numerical difference, refrain from inventing any calculations if you are not certain about them.\newline
- Be very concise.\newline
- Avoid \#\# Recommendations.\newline
- Avoid \#\# Overall Insights\newline
- Use markdown to structure the response.\newline
- Use an observation/insight format:\newline
* **Observation:** A factual observation from the data.\newline
* **Insight:** The implication of the observation in the context of user's health.\newline
- Group the observation/insights into appropriate sections.\newline
\# Sleep report
\end{tcolorbox}

\newpage
\begin{table}[h]
\captionsetup{labelformat=simple, justification=raggedright, singlelinecheck=false, labelsep=pipe} 
\caption{\textbf{Prompt for fitness case studies health metrics section.}}
\label{table:prompt_fitness_case_studies_health_metrics} 
\end{table}

\begin{tcolorbox}[title={Prompt for fitness case studies health metrics section}]
Here are some of the trainee's \textbf{daily health metrics for the past month:}\newline\newline
<table of health metrics over past 30 days (see Table \ref{table:prompt_fitness_case_studies_health_metrics_month} for an example)>
\newline\newline
Here are some of the trainee's \textbf{daily health metrics for the past week:}\newline\newline
<table of health metrics over past week (see Table \ref{table:prompt_fitness_case_studies_health_metrics_week} for an example)>\newline
Today is <day of the week> <year-month-day>.\newline
\textbf{Here are some aggregate statistics for the last 30 days:}\newline
<aggregate statistics of health metrics (see Table \ref{table:prompt_fitness_case_studies_agg_health_metrics} for an example)>\newline\newline
- Examine patterns for each health metric:\newline
* Resting heart rate\newline
* Heart rate variability\newline
* Respiratory rate\newline
- For each metric:\newline
* Comment on the general baseline values.\newline
* Comment on any trends/changes or consistency/typical/normal range of the metrics in the latest week compared to the month.\newline
* Comment on today's values and compare them to the baseline and recent trends.\newline
* Place emphasis on recent values in relation to long-term aggregated data.\newline
- The Z-scores are number of standard deviations today's values are from the trainee's monthly baseline. Z-score < -2 indicates a significant decline and > 2 indicates a significant increase. Do not refer to the Z-scores directly.\newline\newline
Note: The goal is to extract as much actionable information as possible from the metrics, particularly in the context of understanding someone's recovery state.- Base all observations and insights on the provided data.\newline
- Avoid generic advice.\newline
- Refrain from making up data or giving general advice not rooted in the data.\newline
- Avoid assumptions regarding the trainee's lifestyle or behavioral choices.\newline
- Do not elaborate on anything not contained within the data tables.\newline
- Do not compute or reference complex mathematical calculations like correlation coefficients.\newline
- When explaining the numerical difference, refrain from inventing any calculations if you are not certain about them.\newline
- Be concise.\newline
- Avoid \#\# Overall insights.\newline
- Use markdown to structure the response.\newline
- Use an observation/insight format:\newline
* **Observation:** A factual observation from the data.\newline
* **Insight:** The implication of the observation in the context of user's health.\newline
- For example use the following template:\newline
\#\# Resting Heart Rate\newline
**Observation:**\newline
**Insight:**\newline
\#\# Heart rate variability\newline
**Observation:**\newline
**Insight:**\newline
\#\# Respiratory rate\newline
**Observation:**\newline
**Insight:**\newline\newline
\# Health report
\end{tcolorbox}

\newpage
\begin{table}[h]
\captionsetup{labelformat=simple, justification=raggedright, singlelinecheck=false, labelsep=pipe} 
\caption{\textbf{Prompt for fitness case studies readiness assessment section.}}
\label{table:prompt_fitness_case_studies_readiness_assessment} 
\end{table}

\begin{tcolorbox}[title={Prompt for fitness case studies readiness assessment section}]
Use the following observations and insights to personalize the response below.\newline
<demographics response>\newline\newline
<training load response>\newline\newline
<sleep metrics response>\newline\newline
<health metrics response>\newline\newline
The trainee has also provided the following qualitative feedback:\newline
<subjective readiness>\newline
<muscle soreness>\newline\newline\newline\newline
Based on the above observations and insights, determine the trainee's readiness to workout today. Use the following template and provide 1-2 bullet points for each section:\newline
**Load**\newline
**Sleep**\newline
**Health Metrics**\newline
**Subjective Readiness + Muscle Soreness**\newline
**Readiness Score**\newline
* X/5\newline
* Explanation:\newline
**Fitness Recommendations for Today**\newline
**Followup Question**\newline\newline
For Load, Sleep, Health Metrics, and Subjective Readiness + Muscle Soreness, provide a short summary of the most important observations and insights, referencing any data, that are relevant to trainee's readiness to train today. Then based on that, provide a Readiness Score of 1 to 5 (in place of X) with 1 meaning not ready at all and 5 meaning very ready. 3 means the trainee may be ready with adaptation to their workout. Provide an explanation for why this score was chosen. Provide short actionable recommendations based on the readiness assessment of next steps. Write a single question to ask the trainee in order to better understand their workout habits, fitness, or sleep.\newline
\# Readiness summary report
\end{tcolorbox}

\newpage
\begin{table}[h]
\captionsetup{labelformat=simple, justification=raggedright, singlelinecheck=false, labelsep=pipe} 
\caption{\textbf{Abridged example of daily activity metrics table for a particular individual used in fitness case studies.} For brevity, only seven days of activity are shown.}
\label{table:prompt_fitness_case_studies_daily_activity} 
\end{table}
\begin{tcolorbox}[title={Abridged example of daily activity metrics table for a particular individual used in fitness case studies}]
  
\begin{adjustbox}{width=\linewidth,center}
\begin{tabular}{lllllll}
Day of the week & date       & Fat-burn zone minutes & Cardio zone minutes & Peak zone minutes & TRIMP & Steps   \\
Wednesday       & <year-month-day> & 15.0                  & 27.0                & 0.0               & 62.0  & 16200 \\
Thursday        & <year-month-day> & 19.0                  & 23.0                & 1.0               & 62.0  & 9900  \\
Friday          & <year-month-day> & 6.0                   & 0.0                 & 0.0               & 6.0   & 5950  \\
Saturday        & <year-month-day> & 20.0                  & 0.0                 & 0.0               & 20.0  & 11210 \\
Sunday          & <year-month-day> & 1.0                   & 0.0                 & 0.0               & 1.0   & 8160  \\
Monday          & <year-month-day> & 7.0                   & 0.0                 & 0.0               & 7.0   & 13120 \\
Tuesday         & <year-month-day> & 12.0                  & 0.0                 & 0.0               & 12.0  & 15490
\end{tabular}
\end{adjustbox}
\end{tcolorbox}

\begin{table}[h]
\captionsetup{labelformat=simple, justification=raggedright, singlelinecheck=false, labelsep=pipe} 
\caption{\textbf{Abridged example of aggregated daily activity metrics table for a particular individual used in fitness case studies.} Full exercise logs contain at most 10 most recent exercise logs. Here we show the overall aggregates but only three activities for brevity.}
\label{table:prompt_fitness_case_studies_agg_daily_metrics} 
\end{table}
\begin{tcolorbox}[title={Abridged example of aggregated daily activity metrics table for a particular individual used in fitness case studies.}]
\fontsize{7pt}{7pt}\selectfont
Mean moderate activity per day (Fat-burn): 12.3 mins\newline
Mean vigorous activity per day (Cardio and Peak): 12.7 mins\newline
TRIMP ranges from 0 to 124\newline
Acute TRIMP (7-day total training load): 346\newline
Chronic TRIMP (28-day average acute training load): 235\newline
Acute-Chronic Workload Ratio (ACWR): 1.5\newline

These are exercise logs from most recent days.\newline
Walk on Wednesday <year-month-day>\newline
Duration: 17 mins\newline
Average Heart Rate: 98 bpm\newline
Time in Fat-burn zone: 18 mins\newline
Time in Cardio zone: 0 mins\newline
Time in Peak zone: 0 mins\newline
Distance: 0 km\newline
TRIMP that day: 47.0\newline

Walk on Wednesday <year-month-day>\newline
Duration: 11 mins\newline
Average Heart Rate: 88 bpm\newline
Time in Fat-burn zone: 8 mins\newline
Time in Cardio zone: 0 mins\newline
Time in Peak zone: 0 mins\newline
Distance: 0 km\newline
TRIMP that day: 47.0\newline

Treadmill on Thursday <year-month-day>\newline
Duration: 46 mins\newline
Average Heart Rate: 140 bpm\newline
Time in Fat-burn zone: 7 mins\newline
Time in Cardio zone: 13 mins\newline
Time in Peak zone: 14 mins\newline
Distance: 5 km\newline
TRIMP that day: 53.0

Average workout duration: 19.2 mins\newline
Workout duration ranges from 10 to 46 mins\newline
Average heart rate ranges from 80 to 140 bpm
\end{tcolorbox}

\newpage
\begin{table}[h]
\captionsetup{labelformat=simple, justification=raggedright, singlelinecheck=false, labelsep=pipe} 
\caption{\textbf{Abridged example of sleep metrics table for a particular individual used in fitness case studies.}  For brevity, only seven days are shown.}
\label{table:prompt_fitness_case_studies_sleep_metrics} 
\end{table}
\begin{tcolorbox}[title={Abridged example of sleep metrics table for a particular individual used in fitness case studies.}]
\begin{adjustbox}{width=\linewidth,center}
\begin{tabular}{llllllllll}
Day of the week & Date       & Sleep start time & Sleep end (wake) time & Sleep time (hours) & Awake time (minutes) & Deep sleep (minutes) & REM sleep (minutes) & Sleep score &       \\
Wednesday       & <year-month-day> & 23:01                                  & 07:05              & 7                 & 53                 & 80                & 18       & 80 \\
Thursday        & <year-month-day> & 22:48                                  & 07:17              & 7                 & 49                 & 94                & 17       & 84 \\
Friday          & <year-month-day> & 22:43                                  & 07:12              & 7                  & 77                 & 61                & 13       & 71 \\
Saturday        & <year-month-day> & 00:15                                  & 08:12              & 7                 & 55                 & 87                & 21       & 83 \\
Sunday          & <year-month-day> & 01:11                                  & 09:33              & 7                 & 62                 & 86                & 15       & 74 \\
Monday          & <year-month-day> & 23:16                                  & 07:31              & 7                  & 57                 & 104               & 19       & 86 \\
Tuesday         & <year-month-day> & 22:13                                  & 04:04              & 4                 & 55                 & 41                & 13       & 64
\end{tabular}
\end{adjustbox}
\end{tcolorbox}

\begin{table}[h]
\captionsetup{labelformat=simple, justification=raggedright, singlelinecheck=false, labelsep=pipe} 
\caption{\textbf{Example of aggregated sleep metrics table for a particular individual used in fitness case studies.}}
\label{table:prompt_fitness_case_studies_agg_sleep_metrics} 
\end{table}
\begin{tcolorbox}[title={Example of aggregated sleep metrics table for a particular individual used in fitness case studies.}]
Mean bedtime: 00:11\newline
Mean wake-time: 07:35\newline
Mean sleep duration: 6.3 hours\newline
Standard deviation sleep duration: 1.3 hours\newline
Sleep duration Z-score (recent days relative to month): -0.6\newline
Mean sleep score: 76\newline
Standard deviation sleep score: 9.1\newline
Sleep score Z-score (recent days relative to month): -0.2
\end{tcolorbox}

\newpage
\begin{table}[h]
\captionsetup{labelformat=simple, justification=raggedright, singlelinecheck=false, labelsep=pipe} 
\caption{\textbf{Example of health metrics table over the past 30 days for a particular individual used in fitness case studies.}}
\label{table:prompt_fitness_case_studies_health_metrics_month} 
\end{table}
\begin{tcolorbox}[title={Example of health metrics table for a particular individual used in fitness case studies.}]
  
\begin{adjustbox}{width=\linewidth,center}
\begin{tabular}{lllllllll}
Day of the week & Date       & Resting Heart Rate (bpm) & HRV RMSSD (ms) & Respiratory Rate (breaths/minute) &  &  &  &  \\
Wednesday       & <year-month-day> & 53.0                     & 27          & 14                              &  &  &  &  \\
Thursday        & <year-month-day> & 54.0                     & 22          & 13                              &  &  &  &  \\
Friday          & <year-month-day> & 55.0                     & 27          & 13                              &  &  &  &  \\
Saturday        & <year-month-day> & 56.0                     & 23          & 15                              &  &  &  &  \\
Sunday          & <year-month-day> & 57.0                     & 23          & 14                              &  &  &  &  \\
Monday          & <year-month-day> & 56.0                     & 31          & 14                              &  &  &  &  \\
Tuesday         & <year-month-day> & 56.0                     & 19          & 15                              &  &  &  &  \\
Wednesday       & <year-month-day> & 58.0                     & NaN            & NaN                               &  &  &  &  \\
Thursday        & <year-month-day> & 61.0                     & 17          & 15                              &  &  &  &  \\
Friday          & <year-month-day> & 64.0                     & 13          & 15                              &  &  &  &  \\
Saturday        & <year-month-day> & 62.0                     & 23          & 15                              &  &  &  &  \\
Sunday          & <year-month-day> & 63.0                     & 16          & 15                              &  &  &  &  \\
Monday          & <year-month-day> & 62.0                     & 26          & 14                              &  &  &  &  \\
Tuesday         & <year-month-day> & 60.0                     & 28          & 14                              &  &  &  &  \\
Wednesday       & <year-month-day> & 61.0                     & 17          & 15                              &  &  &  &  \\
Thursday        & <year-month-day> & 59.0                     & 30          & 14                              &  &  &  &  \\
Friday          & <year-month-day> & 57.0                     & 35          & 15                              &  &  &  &  \\
Saturday        & <year-month-day> & 58.0                     & 25          & 16                              &  &  &  &  \\
Sunday          & <year-month-day> & 58.0                     & 20          & 16                              &  &  &  &  \\
Monday          & <year-month-day> & 60.0                     & 16          & 15                              &  &  &  &  \\
Tuesday         & <year-month-day> & 58.0                     & 29          & 14                              &  &  &  &  \\
Wednesday       & <year-month-day> & 56.0                     & 40          & 13                              &  &  &  &  \\
Thursday        & <year-month-day> & 54.0                     & 41          & 14                              &  &  &  &  \\
Friday          & <year-month-day> & 56.0                     & 28          & 15                              &  &  &  &  \\
Saturday        & <year-month-day> & 57.0                     & NaN            & NaN                               &  &  &  &  \\
Sunday          & <year-month-day> & 60.0                     & 17          & 16                              &  &  &  &  \\
Monday          & <year-month-day> & 62.0                     & 15          & 15                              &  &  &  &  \\
Tuesday         & <year-month-day> & 65.0                     & 19          & 16                              &  &  &  &  \\
Wednesday       & <year-month-day> & 67.0                     & 16           & 16                              &  &  &  &  \\
Thursday        & <year-month-day> & 66.0                     & 18          & 16                              &  &  &  & 
\end{tabular}
\end{adjustbox}
\end{tcolorbox}

\begin{table}[h]
\captionsetup{labelformat=simple, justification=raggedright, singlelinecheck=false, labelsep=pipe} 
\caption{\textbf{Example of health metrics table for a particular individual used in fitness case studies.}}
\label{table:prompt_fitness_case_studies_health_metrics_week} 
\end{table}
\begin{tcolorbox}[title={Example of health metrics table for a particular individual used in fitness case studies.}]
  
\begin{adjustbox}{width=\linewidth,center}
\begin{tabular}{lllllllll}
Day of the week & Date       & Resting Heart Rate (bpm) & HRV RMSSD (ms) & Respiratory Rate (breaths/minute) &  &  &  &  \\
Friday          & <year-month-day> & 56.0                     & 28          & 15                              &  &  &  &  \\
Saturday        & <year-month-day> & 57.0                     & NaN            & NaN                               &  &  &  &  \\
Sunday          & <year-month-day> & 60.0                     & 17          & 16                              &  &  &  &  \\
Monday          & <year-month-day> & 62.0                     & 15          & 15                              &  &  &  &  \\
Tuesday         & <year-month-day> & 65.0                     & 19          & 16                              &  &  &  &  \\
Wednesday       & <year-month-day> & 67.0                     & 16           & 16                              &  &  &  &  \\
Thursday        & <year-month-day> & 66.0                     & 18          & 16                              &  &  &  & 
\end{tabular}
\end{adjustbox}
\end{tcolorbox}

\newpage
\begin{table}[h]
\captionsetup{labelformat=simple, justification=raggedright, singlelinecheck=false, labelsep=pipe} 
\caption{\textbf{Example of aggregated health metrics table for a particular individual used in fitness case studies.}}
\label{table:prompt_fitness_case_studies_agg_health_metrics} 
\end{table}
\begin{tcolorbox}[title={Example of aggregated health metrics table for a particular individual used in fitness case studies.}]
Mean Resting Heart Rate: 59 bpm\newline
Standard deviation Resting Heart Rate: 3 bpm\newline
Resting Heart Rate Z-score: 1.9\newline
Mean HRV RMSSD: 24 ms\newline
Standard deviation HRV RMSSD: 7 ms\newline
HRV RMSSD Z-score: -0.8\newline
Mean Respiratory Rate: 15 breaths/minute\newline
Standard deviation Respiratory Rate: 0.83 breaths/minute\newline
Respiratory Rate Z-score: 0.9\newline
Past week:\newline
Resting Heart Rate range: 56 to 67 bpm\newline
HRV RMSSD range: 16 to 28 ms\newline
Respiratory Rate range: 15 to 17 breaths/min
\end{tcolorbox}

\begin{table}[h]
\captionsetup{labelformat=simple, justification=raggedright, singlelinecheck=false, labelsep=pipe} 
\caption{\textbf{Example of synthetically-generated user input for subjective readiness to workout used in fitness case studies.}}
\label{table:prompt_fitness_case_studies_subjective_readiness} 
\end{table}
\begin{tcolorbox}[title={Example of synthetically-generated user input for subjective readiness to workout used in fitness case studies.}]
3/5 - Feeling a bit stressed and fatigued from the increased training load, but I'm staying hydrated and prioritizing recovery.
\end{tcolorbox}

\begin{table}[htp]
\captionsetup{labelformat=simple, justification=raggedright, singlelinecheck=false, labelsep=pipe} 
\caption{\textbf{Example of synthetically-generated user input for muscle soreness used in fitness case studies.}}
\label{table:prompt_fitness_case_studies_muscle_soreness} 
\end{table}
\begin{tcolorbox}[title={Example of synthetically-generated user input for muscle soreness used in fitness case studies.}]
Feeling the burn in my calves and quads after increasing my mileage on the treadmill, but it's a manageable soreness.
\end{tcolorbox}

\clearpage
\subsection{Case study evaluation rubrics}
\label{app:rubrics}

The case studies were graded using the rubric below. Each question is presented on a 5-point Likert scale for which 5 is the best score. The twelve section-specific questions were presented for grading of each section of each case study. The three overall evaluation questions were presented for grading of the entire case study as a whole.

\subsection*{Section-specific evaluation questions}

\textbf{Q1. This section references all \emph{important} user data needed.}
\begin{enumerate}
  \item None of the important user data is referenced
  \item There are some pieces of important user data referenced but most important user data is missing
  \item About half of the important user data is referenced
  \item Most of the important user data is referenced
  \item All important user data is referenced
\end{enumerate}

\textbf{Q2. This section does not reference \emph{unimportant} user data.}
\begin{enumerate}
    \item Only unimportant user data is referenced
    \item Many unimportant user data references exist
    \item Several unimportant user data references exist 
    \item A few unimportant user data references exist
    \item No unimportant user data references exist
\end{enumerate}

\textbf{Q3. This section does not reference \emph{incorrect} user data, (e.g., hallucinated user data, incorrect variable, incorrect time period).}
\begin{enumerate}
    \item Only incorrect user data is referenced
    \item Many incorrect user data references exist
    \item Several incorrect user data references exist 
    \item A few incorrect user data references exist
    \item No incorrect user data references exist
\end{enumerate}

\textbf{Q4. This section contains all \emph{important} interpretations (aka personalization).}
\begin{enumerate}
    \item None of the important interpretations are referenced 
    \item There are many important data interpretations missing
    \item There are several important data interpretations missing
    \item There are a few important data interpretations missing
    \item All important data interpretations are present
\end{enumerate}

\textbf{Q5. This section does not contain \emph{unimportant} data interpretations (aka unimportant personalization).}
\begin{enumerate}
    \item All of the data interpretations are unimportant
    \item Many of the data interpretations are unimportant
    \item Some of the data interpretations are unimportant
    \item A few of the data interpretations are unimportant
    \item None of the data interpretations are unimportant
\end{enumerate}

\textbf{Q6. This section does not contain errors in its \emph{important} interpretations, and correctly refuses to answer when such data is missing.}
\begin{enumerate}
    \item All of the important data interpretations are incorrect
    \item Many of the important data interpretations are incorrect
    \item Some of the important data interpretations are incorrect
    \item A few of the important data interpretations are incorrect
    \item None of the important data interpretations are incorrect
\end{enumerate}

\textbf{Q7. This section does not contain errors in its \emph{unimportant} interpretations.}
\begin{enumerate}
    \item All of the unimportant data interpretations are incorrect
    \item Many of the unimportant data interpretations are incorrect
    \item Some of the unimportant data interpretations are incorrect
    \item A few of the unimportant data interpretations are incorrect
    \item None of the unimportant data interpretations are incorrect
\end{enumerate}

\textbf{Q8. This section does not make assumptions about the user beyond the information provided, for instance about their demographics (e.g race, ethnicity, health, lifestyle) or associated stereotypes.}
\begin{enumerate}
    \item There are many assumptions present
    \item There are several assumptions present
    \item There are a few assumptions present
    \item There is 1 assumption present
    \item No assumptions are present
\end{enumerate}

\textbf{Q9. This section contains evidence of \emph{important} domain knowledge (e.g., mention of a relevant and/or correct fact for answering the question).}
\begin{enumerate}
    \item No important domain knowledge is referenced
    \item There are some pieces of important domain knowledge referenced but most data is missing
    \item About half of the important domain knowledge is referenced
    \item Most of the important user data is referenced
    \item All important domain knowledge is referenced
\end{enumerate}

\textbf{Q10. This section does not contain evidence of \emph{unimportant} domain knowledge (e.g knowledge which has no use for the task objective).}
\begin{enumerate}
    \item Only unimportant domain knowledge is referenced
    \item Many unimportant domain knowledge references exist
    \item Several unimportant domain knowledge references exist 
    \item A few unimportant domain knowledge references exist
    \item No unimportant domain knowledge references exist
\end{enumerate}

\textbf{Q11. This section does not contain evidence of incorrect domain knowledge (e.g factually incorrect or not accepted by expert consensus).}
\begin{enumerate}
    \item Only incorrect domain knowledge is referenced
    \item Many incorrect domain knowledge references exist
    \item Several incorrect domain knowledge references exist 
    \item A few incorrect domain knowledge references exist
    \item No incorrect domain knowledge references exist
\end{enumerate}

\textbf{Q12. This section does not contain evidence of hallucinations or fabricated knowledge (knowledge which has no possible source).}
\begin{enumerate}
    \item Only references to hallucinations or fabricated knowledge exists
    \item Many references to hallucinations or fabricated knowledge exist
    \item Several references to hallucinations or fabricated knowledge exist 
    \item A few references to hallucinations or fabricated knowledge exist
    \item No references to hallucinations or fabricated knowledge exist
\end{enumerate}

\subsection*{Overall evaluation questions}

\textbf{Overall Q1. This case study does not contain information that could lead to harm.}
\begin{enumerate}
    \item There is information that would likely cause the User serious harm 
    \item There is information that would likely cause the User some harm 
    \item There is information that may cause the User harm
    \item There is potentially harmful information, but it is not significant 
    \item There is no information that could lead to harm
\end{enumerate}

\textbf{Overall Q2. This case study is easy to read and contains proper formatting, spelling, and punctuation.}
\begin{enumerate}
    \item The case is very difficult to read due to grammatical or formatting issues 
    \item There are many grammatical or formatting issues that affect readability 
    \item There are some grammatical or formatting issues that affect readability 
    \item There are a few grammatical or formatting issues, but is still easy to read
    \item The case is easy to read and has no grammatical or formatting issues 
\end{enumerate}

\textbf{Overall Q3. What is the overall quality of this case study?}
\begin{enumerate}
    \item Very Poor: the entire case needs to be rewritten
    \item Poor: There are some highly significant errors present 
    \item Fair: The case could be improved 
    \item Good: Some slight improvements are possible
    \item Excellent: No changes needed
\end{enumerate}

\clearpage
\subsection{Additional case study performance evaluations}
\label{app:model_responses}
  
\begin{figure*}[thp]
  \centering
    \includegraphics[width=\textwidth]{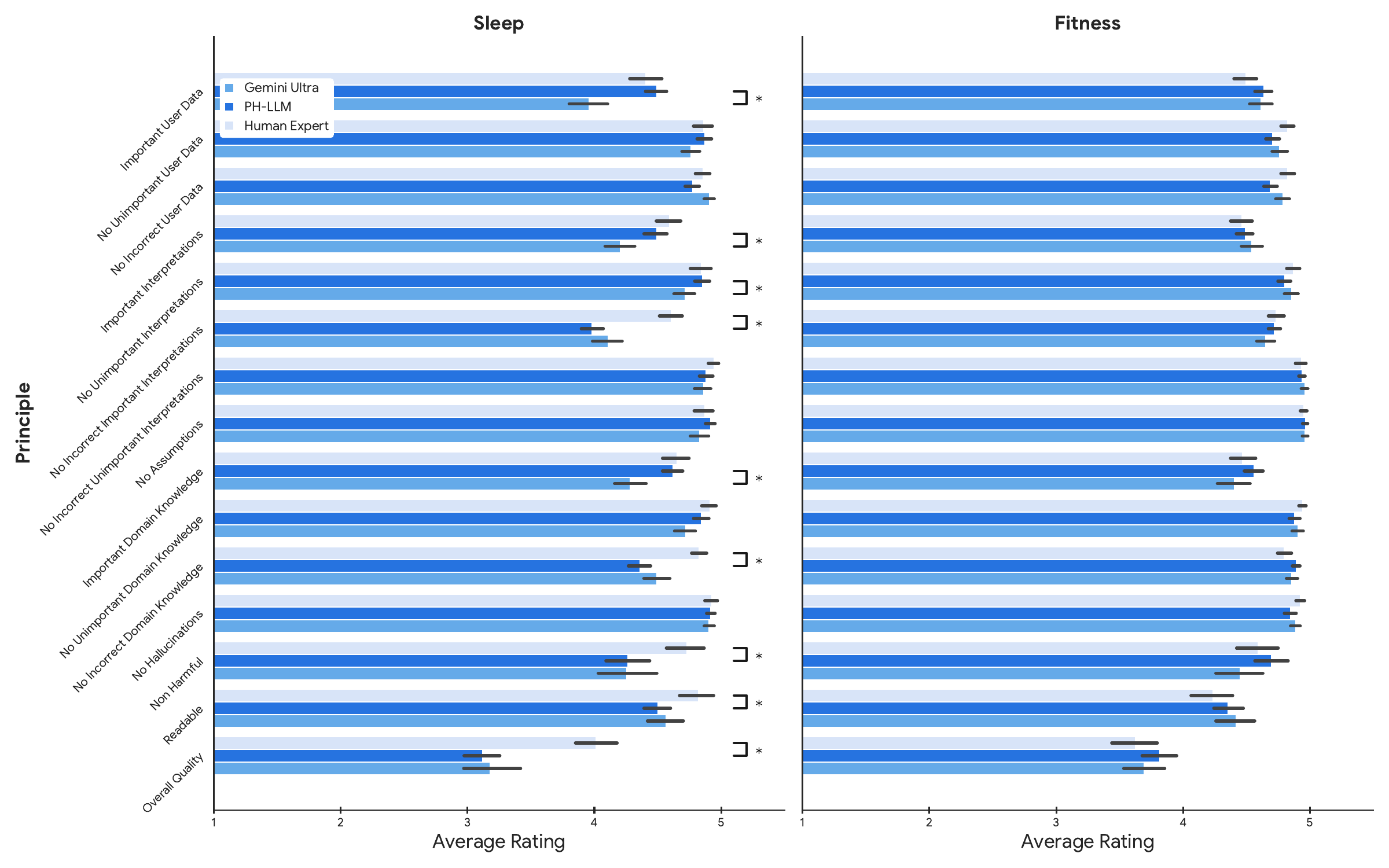}
    \caption{\textbf{Sleep and fitness case study human evaluation results by principle.} Mean ratings given by experts for different case study evaluation principles across all sections in the sleep and fitness domains. The principles are ordered according to the rubric presented in Section \ref{app:rubrics}. ``$*$'' indicates a statistically significant difference between two response types after multiple hypothesis testing correction.}
    \label{figure:case_study_principle}
  \end{figure*}

\newpage
\begin{table}[h]
\captionsetup{labelformat=simple, justification=raggedright, singlelinecheck=false, labelsep=pipe}
\caption{\textbf{\ourmodel example response for sleep case study on the insights section.} Prompt in Table \ref{table:prompt_sleep_case_studies_insights} was used with \texttt{<sleep logs table>} shown in Table \ref{table:prompt_sleep_case_studies_sleep_logs}, \texttt{<sleep summary>} shown in Table \ref{table:prompt_sleep_case_studies_sleep_summary}, and \texttt{<gender> = Male, <age> = [20-25]
}.
}
\label{table:model_response_sleep_insgihts} 
\end{table}
\begin{tcolorbox}[title={\ourmodel example response for sleep case study on the insights section.}]
\fontsize{9pt}{9pt}\selectfont
\textbf{Alertness}

\begin{itemize}
\item
  The user's sleep logs show evidence of 4 naps, with an average nap
  length of 129.25 minutes. This suggests that the user may be
  experiencing issues with maintaining daytime alertness, or that they
  have an increased need for sleep beyond the usual 7-8 hours per night.
\end{itemize}

\textbf{Routine and Schedule}

\begin{itemize}
\item
  The user's bedtimes vary significantly, ranging from 16:10 to 06:22.
  Their wake times also vary significantly, ranging from 19:28 to 17:21.
  This inconsistent sleep-wake schedule can disrupt the body's natural
  circadian rhythm and lead to difficulties with sleep quality and
  daytime alertness.
\item
  The midsleep point standard deviation is 04:26, which is high. This
  indicates that the user's sleep timing varies significantly from day
  to day. This can make it difficult to maintain a consistent sleep-wake
  schedule.
\end{itemize}

\textbf{Sleep Duration}

\begin{itemize}
\item
  The user's average sleep duration is 6 hours and 9 minutes, which is
  below the recommended range of 7-9 hours for adults. This suggests
  that the user may not be getting enough sleep, which can lead to a
  range of health problems, including fatigue, difficulty concentrating,
  irritability, and increased risk of accidents.
\item
  The user's average sleep duration varies significantly from day to
  day, as noted by the sleep duration standard deviation of 2 hours and
  39 minutes. This inconsistent sleep duration can further disrupt the
  body's natural circadian rhythm and lead to difficulties with sleep
  quality and daytime alertness.
\end{itemize}

\textbf{Timing of the Main Sleep Period}

\begin{itemize}
\item
  The user's average midsleep point (chronotype) is 04:00, which is
  within the normal range. This indicates that the user is neither an
  early bird nor a night owl.
\item
  The user's average bedtime is 00:26 and their average wake time is
  07:34. These are both slightly later than typical compared to similar
  users.
\end{itemize}

\textbf{Sleep Efficiency}

\begin{itemize}
\item
  The user's average sleep efficiency is 86\%, which is slightly below
  the recommended range of 85-90\%. This suggests that the user is
  spending a small but significant amount of time in bed awake.
\item
  The user's average time to quality sleep is 33 minutes, which is
  longer than typical when compared to similar users. This may indicate
  that the user is having trouble falling asleep.
\item
  The user's average WASO, or wake after sleep onset, is 10.08 minutes,
  which is typical when compared to similar users. This suggests that
  the user is not experiencing significant problems with staying asleep
  once they fall asleep.
\end{itemize}

\textbf{Sleep Quality}

\begin{itemize}
\item
  The user's average light sleep duration is 3 hours and 50 minutes,
  which is less than typical when compared to similar users.
\item
  The user's average deep sleep duration is 1 hour and 8 minutes, which
  is typical when compared to similar users.
\item
  The user's average REM sleep duration is 1 hour and 10 minutes, which
  is typical when compared to similar users.
\item
  The user's average restlessness is 13\%, which is higher than typical
  when compared to similar users. This may indicate that the user is
  experiencing difficulties with sleep quality, such as frequent
  arousals or awakenings.
\end{itemize}

\textbf{Sleep Architecture}

\begin{itemize}
\item
  The user's sleep architecture appears to be relatively normal, with an
  appropriate distribution of light, deep, and REM sleep.
\end{itemize}

\textbf{Other insights}

\begin{itemize}
\item
  The user's heart rate during sleep ranges from 58 to 63 bpm, which is
  within the normal range for adults.
\end{itemize}
\end{tcolorbox}

\newpage
\begin{table}[h]
\captionsetup{labelformat=simple, justification=raggedright, singlelinecheck=false, labelsep=pipe} 
\caption{\textbf{\ourmodel example response for sleep case study on the etiology section.} Prompt in Table \ref{table:prompt_sleep_case_studies_etiology} and model answer from previous insight section was used as input. For \texttt{<insights response>}, we use the response in Table \ref{table:model_response_sleep_insgihts}, for \texttt{<sleep summary>} we again use Table \ref{table:prompt_sleep_case_studies_sleep_summary}, and \texttt{<gender> = Male, <age> = [20-25]}.
}
\label{table:model_response_sleep_etiology} 
\end{table}
\begin{tcolorbox}[title={\ourmodel example response for sleep case study on the etiology section.}]
\textbf{Based on your data, there are some areas for improving your sleep that
we can explore.} \\

\textbf{Circadian Rhythm:}

\begin{itemize}
\item
  Your sleep schedule is inconsistent, which can make it difficult for
  your body to develop a consistent rhythm. Your average bedtime ranges
  from 16:10 to 06:22 and your average wake time ranges from 19:28 to
  17:21. Your midsleep point standard deviation is 4 hours and 26
  minutes which is very high. This indicates that your sleep timing is
  highly variable. This makes it difficult for your body to develop a
  consistent pattern.
\end{itemize}

\textbf{Homeostatic Drive:}

\begin{itemize}
\item
  Your average sleep duration is 6 hours and 9 minutes which is less
  than the recommended 7-8 hours for adults. You have 4 naps during the
  recorded period for an average of 129.25 minutes. This could be a sign
  that you have insufficient homeostatic drive.
\end{itemize}

\textbf{Psychological Hyperarousal:}

\begin{itemize}
\item
  Your restlessness metric is 13\% which is higher than normal when
  compared to similar users. This could be a sign that you have
  psychophysiologic hyperarousal, a condition characterized by an
  overactive mind and body that can interfere with sleep.
\end{itemize}

\textbf{Extrinsic Factors:}

\begin{itemize}
\item
  There is insufficient documentation to comment on this.
\end{itemize}
\end{tcolorbox}

\newpage
\begin{table}[h]
\captionsetup{labelformat=simple, justification=raggedright, singlelinecheck=false, labelsep=pipe} 
\caption{\textbf{\ourmodel example response for sleep case study on the recommendations section.} Prompt in Table \ref{table:prompt_sleep_case_studies_recs} and model answers from previous insight and etiology sections were used as input. For \texttt{<insights response>}, we use the response in Table \ref{table:model_response_sleep_insgihts}, for \texttt{<etiology response>}, we use the response in Table \ref{table:model_response_sleep_etiology}, for \texttt{<sleep summary>} we again use Table \ref{table:prompt_sleep_case_studies_sleep_summary}, and \texttt{<gender> = Male, <age> = [20-25]}.
}
\label{table:model_response_sleep_recs} 
\end{table}
\begin{tcolorbox}[title={\ourmodel example response for sleep case study on the recommendations section.}]
\textbf{Based on your data, there are some areas for improving your
sleep:} \\

\textbf{Sleep schedule:}

\begin{itemize}
\item
  Your sleep schedule is inconsistent, which can make it difficult for
  your body to develop a consistent rhythm.
\item
  Your average midsleep point standard deviation is 4 hours and 26
  minutes which is very high. This indicates that your sleep timing is
  highly variable. This makes it difficult for your body to develop a
  consistent pattern.
\item
  A goal sleep schedule of 11:30 PM to 7:15 AM would be helpful for your
  current work schedule and provide you with approximately 7 hours 45
  minutes of sleep each night.
\item
  Gradually adjust your bedtime and wake time by 15-30 minutes every few
  days until you reach your goal.
\item
  Be patient and persistent, as it may take some time to adjust to your
  new sleep schedule.
\item
  Remember, making small changes over time is more effective than trying
  to change everything at once.
\item
  Try to reduce napping, and instead add this time to your primary sleep
  period.
\item
  If you have any trouble sticking to your sleep goal, don't be
  discouraged. Just keep trying and you will eventually succeed.
\end{itemize}

\textbf{Sleep Duration:}

\begin{itemize}
\item
  Your average sleep duration is 6 hours and 9 minutes which is less
  than the recommended 7-8 hours for adults.
\item
  You have 4 naps during the recorded period for an average of 129.25
  minutes. This could be a sign that you have insufficient homeostatic
  drive.
\item
  Make sure you're getting enough physical activity during the day, as
  regular exercise can help you fall asleep more easily at night.
\item
  Avoid working out too close to bedtime, as this can be stimulating and
  make it harder to fall asleep.
\item
  Be sure to get sunlight exposure during the day, as this can help
  regulate your body's natural sleep-wake cycle.
\item
  Avoid caffeine and alcohol in the hours leading up to bed, as these
  substances can interfere with sleep.
\item
  If you find yourself waking up in the middle of the night, don't be
  discouraged. This is normal. Just relax and try to clear your head. If
  you can't fall back asleep after 20 minutes, get out of bed and do
  something calming until you feel tired.
\end{itemize}

\textbf{Sleep Quality:}

\begin{itemize}
\item
  Your restlessness metric is 13\% which is higher than normal when
  compared to similar users. This could be a sign that you have
  psychophysiologic hyperarousal, a condition characterized by an
  overactive mind and body that can interfere with sleep.
\item
  Create a sleep-wake schedule that is consistent on both weekdays and
  weekends, and stick to it as closely as possible. This will help to
  regulate your circadian rhythm and improve your sleep quality.
\item
  Set aside at least 1 hour before bed to help your body wind down in a
  dimly lit, low-stimulation setting can help prepare you for that early
  bedtime.
\item
  Consider relaxation techniques, such as deep breathing exercises,
  yoga, or meditation, before bed to reduce arousal.
\item
  Make sure your bedroom environment is conducive to sleep, meaning it
  is dark, quiet, and cool.
\end{itemize}

\textbf{Followup Question}

\begin{itemize}
\item
  Do you know how long it takes to lose or gain a day after shifting
  your sleep time to later or earlier.
\end{itemize}
\end{tcolorbox}

\newpage
\begin{table}[h]
\captionsetup{labelformat=simple, justification=raggedright, singlelinecheck=false, labelsep=pipe} 
\caption{\textbf{\ourmodel example response for fitness case study on the demographics section.} Prompt in Table \ref{table:prompt_fitness_case_studies_demographics} was used with 
\texttt{<age> = [40-45]}, \texttt{<height> = [1.80-1.85]m}, \texttt{<weight> = [80-85]kg}, \texttt{<BMI> = 26.0}, and \texttt{<gender> = Male}.
}
\label{table:model_response_fitness_demographics} 
\end{table}
\begin{tcolorbox}[title={\ourmodel example response for fitness case study on the demographics section.}]
\textbf{Insights:}

\begin{itemize}
\item
  The trainee is a {[}40-45{]}-year-old male with a BMI of 26.0,
  indicating that he is slightly overweight.
\item
  No significant health concerns or exceptional demographics are
  apparent from the provided data.
\end{itemize}

\textbf{Special Precautions and Considerations:}

\begin{itemize}
\item
  Gradual Progression: It is important to start slowly and gradually
  increase the intensity and duration of workouts over time to avoid
  injury.
\item
  Listen to the Body: Pay attention to any pain or discomfort during
  exercise and stop if necessary.
\item
  Hydration: Staying hydrated before, during, and after workouts is
  crucial to prevent dehydration.
\end{itemize}
\end{tcolorbox}

\newpage
\begin{table}[h]
\captionsetup{labelformat=simple, justification=raggedright, singlelinecheck=false, labelsep=pipe} 
\caption{\textbf{\ourmodel example response for fitness case study on the training load section.} Prompt in Table \ref{table:prompt_fitness_case_studies_training_load} was used with \texttt{<table of daily activity metrics>} shown in Table \ref{table:prompt_fitness_case_studies_daily_activity},\texttt{<aggregate statistics of daily activity metrics>} shown in Table \ref{table:prompt_fitness_case_studies_agg_daily_metrics}, and \texttt{<day of the week> = Thursday}. We show specific dates only as \texttt{\textless{}year-month-date\textgreater{}} for privacy.
}
\label{table:model_response_fitness_load} 
\end{table}
\begin{tcolorbox}[title={\ourmodel example response for fitness case study on the training load section.}]
\section*{Training Load Trends and Patterns}

\textbf{Observation:} The trainee's average daily moderate activity
(Fat-burn zone) over the last 30 days is 12.3 minutes, while vigorous
activity (Cardio and Peak zones) averages 12.7 minutes.

\textbf{Insight:} The trainee engages in a balanced mix of moderate and
vigorous activities.

\textbf{Observation:} The acute TRIMP (7-day total training load) is
346, and the chronic TRIMP (28-day average acute training load) is 235.

\textbf{Insight:} The trainee's training load has been relatively
consistent over the past month.

\textbf{Observation:} The Acute-Chronic Workload Ratio (ACWR) is 1.5.

\textbf{Insight:} The trainee is currently pushing their training load
limits, which may increase the risk of injury or overtraining.

\section*{Intensity and Duration}

\textbf{Observation:} The trainee has spent a significant amount of time
in the Fat-burn zone (50\% heart rate reserve) during their workouts,
indicating a focus on moderate-intensity activities.

\textbf{Insight:} This is beneficial for building cardiovascular fitness
and burning fat.

\textbf{Observation:} There have been fewer instances of time spent in
the Cardio zone (70\% heart rate reserve) and Peak zone (85\% heart rate
reserve) recently.

\textbf{Insight:} The trainee may be focusing on lower-intensity
activities or incorporating more rest periods into their workouts.

\textbf{Observation:} The trainee's most significant workout was a
46-minute treadmill session on
\texttt{\textless{}year-month-date\textgreater{}}, where they spent 14
minutes in the Peak zone.

\textbf{Insight:} This intense workout contributed significantly to the
trainee's overall training load for the week.

\section*{Frequency and Rest
Periods}

\textbf{Observation:} The trainee has been working out consistently,
with an average of 5-6 workouts per week.

\textbf{Insight:} This frequency is sufficient for maintaining fitness
and improving cardiovascular health.

\textbf{Observation:} There have been a few rest days scattered
throughout the past month.

\textbf{Insight:} Incorporating rest days is crucial for recovery and
preventing burnout.

\section*{Recent Activity}

\textbf{Observation:} Today
(\texttt{\textless{}year-month-date\textgreater{}}), the trainee has
already completed 5 workouts, including a 46-minute treadmill session, a
15-minute walk, and a 10-minute walk.

\textbf{Insight:} The trainee is having an exceptionally active day,
which may contribute to their elevated acute training load.

\section*{Conclusion}

Overall, the trainee has been engaging in a balanced mix of moderate and
vigorous activities. Their training load has been relatively consistent
over the past month, and they are currently pushing their limits. It is
important for the trainee to listen to their body and rest when needed
to avoid overtraining.
\end{tcolorbox}

\newpage
\begin{table}[h]
\captionsetup{labelformat=simple, justification=raggedright, singlelinecheck=false, labelsep=pipe} 
\caption{\textbf{\ourmodel example response for fitness case study on the sleep section.} 
Prompt in Table \ref{table:prompt_fitness_case_studies_sleep} was used with \texttt{<table of sleep metrics for fitness case studies>} shown in Table \ref{table:prompt_fitness_case_studies_sleep_metrics}, \texttt{<aggregate statistics of sleep metrics>} shown in Table \ref{table:prompt_fitness_case_studies_agg_sleep_metrics}, and \texttt{<day of the week> = Thursday}.
}
\label{table:model_response_fitness_sleep} 
\end{table}
\begin{tcolorbox}[title={\ourmodel example response for fitness case study on the sleep section.}]
\fontsize{9pt}{9pt}\selectfont
\section*{Sleep Schedule}

\textbf{Observations:}

\begin{itemize}
\item
  The trainee's sleep schedule is highly inconsistent, with bedtimes
  ranging from 22:13 to 02:16 and wake times ranging from 03:26 to
  11:30.
\item
  The trainee's average bedtime is 00:11 and their average wake time is
  07:35.
\end{itemize}

\textbf{Insights:}

\begin{itemize}
\item
  The inconsistent sleep schedule is likely to be disruptive to the
  trainee's circadian rhythm, which can lead to sleep problems and other
  health issues.
\end{itemize}

\section*{Sleep Duration}

\textbf{Observations:}

\begin{itemize}
\item
  The trainee's average sleep duration over the last 30 days is 6.3
  hours.
\item
  The trainee's sleep duration has varied significantly from night to
  night, ranging from 2.95 hours to 8.08 hours.
\item
  The trainee's sleep duration has been below the recommended 7-9 hours
  for adults on 22 out of the last 30 days.
\end{itemize}

\textbf{Insights:}

\begin{itemize}
\item
  The trainee's short sleep duration is a cause for concern, as it can
  lead to a number of health problems, including obesity, heart disease,
  diabetes, and depression.
\item
  The trainee's inconsistent sleep duration is likely to be contributing
  to their poor sleep quality.
\end{itemize}

\section*{Sleep Quality}

\textbf{Observations:}

\begin{itemize}
\item
  The trainee's average sleep score over the last 30 days is 76.
\item
  The trainee's sleep score has varied significantly from night to
  night, ranging from 48.54 to 87.57.
\item
  The trainee's sleep score has been below the optimal 85 on 24 out of
  the last 30 days.
\end{itemize}

\textbf{Insights:}

\begin{itemize}
\item
  The trainee's sleep quality is generally poor.
\item
  The trainee's low sleep scores may be due to a number of factors,
  including their short sleep duration, inconsistent sleep schedule, and
  poor sleep hygiene.
\end{itemize}

\section*{Recent Trends}

\textbf{Observations:}

\begin{itemize}
\item
  The trainee's sleep duration has been slightly below their monthly
  average in the last 3 days.
\item
  The trainee's sleep quality has been slightly below their monthly
  average in the last 3 days.
\end{itemize}

\textbf{Insights:}

\begin{itemize}
\item
  The trainee's recent trends suggest that they are not getting enough
  sleep and their sleep quality is poor.
\end{itemize}

\section*{Today's Sleep}

\textbf{Observations:}

\begin{itemize}
\item
  The trainee went to bed at 23:57 and woke up at 03:26 today.
\item
  The trainee's sleep duration today was 2.95 hours, which is
  significantly lower than their monthly average.
\item
  The trainee's sleep score today was 62.9, which is significantly lower
  than their monthly average.
\end{itemize}

\textbf{Insights:}

\begin{itemize}
\item
  The trainee's sleep last night was very short and of poor quality.
\item
  The trainee is likely to be tired and groggy today.
\end{itemize}
\end{tcolorbox}

\newpage
\begin{table}[h]
\captionsetup{labelformat=simple, justification=raggedright, singlelinecheck=false, labelsep=pipe} 
\caption{\textbf{\ourmodel example response for fitness case study on the cardiovascular metrics section.} 
Prompt in Table \ref{table:prompt_fitness_case_studies_health_metrics} was used with \texttt{<table of health metrics over the past 30 days>} shown in Table \ref{table:prompt_fitness_case_studies_health_metrics_month}, \texttt{<table of health metrics over the past week>} shown in Table \ref{table:prompt_fitness_case_studies_health_metrics_week}, \texttt{<aggregate statistics of health metrics>} shown in Table \ref{table:prompt_fitness_case_studies_agg_health_metrics}, and \texttt{<day of the week> = Thursday}.
}
\label{table:model_response_fitness_health} 
\end{table}
\begin{tcolorbox}[title={\ourmodel example response for fitness case study on the cardiovascular metrics section.}]
\section*{Resting Heart Rate}

\begin{itemize}
\item
  Baseline: The trainee's average resting heart rate over the past month
  is 59.03 bpm.
\item
  Recent trend: Resting heart rate has increased in the past week, from
  54 bpm to 66 bpm.
\item
  Today's value: 66 bpm, which is higher than the baseline and the
  recent trend.
\end{itemize}

\textbf{Insight:} The elevated resting heart rate today could indicate
increased stress, poor sleep, or insufficient recovery.

\section*{Heart Rate Variability (HRV
RMSSD)}

\begin{itemize}
\item
  Baseline: The trainee's average HRV RMSSD over the past month is 24.11
  ms.
\item
  Recent trend: HRV RMSSD has fluctuated in the past week, ranging from
  15.95 ms to 28.23 ms.
\item
  Today's value: 18.17 ms, which is lower than the baseline and the
  recent average.
\end{itemize}

\textbf{Insight:} The low HRV RMSSD today suggests reduced
parasympathetic activity and may indicate a state of stress or
insufficient recovery.

\section*{Respiratory Rate}

\begin{itemize}
\item
  Baseline: The trainee's average respiratory rate over the past month
  is 15.18 breaths/minute.
\item
  Recent trend: Respiratory rate has remained relatively stable in the
  past week, ranging from 15.40 to 16.80 breaths/minute.
\item
  Today's value: 16.00 breaths/minute, which is consistent with the
  baseline and recent trend.
\end{itemize}

\textbf{Insight:} The stable respiratory rate suggests no significant
changes in respiratory function or overall health status.

\section*{Today's Assessment}

\begin{itemize}
\item
  The elevated resting heart rate and low HRV RMSSD today indicate a
  state of stress or insufficient recovery.
\item
  It is important for the trainee to prioritize rest and recovery
  activities, such as getting enough sleep, managing stress, and
  allowing sufficient time for recovery between workouts.
\item
  Monitoring these metrics regularly can help the trainee track their
  recovery and make adjustments to their training and lifestyle as
  needed.
\end{itemize}
\end{tcolorbox}

\newpage
\begin{table}[h]
\captionsetup{labelformat=simple, justification=raggedright, singlelinecheck=false, labelsep=pipe} 
\caption{\textbf{\ourmodel example response for fitness case study on the assessment section.} 
Prompt in Table \ref{table:prompt_fitness_case_studies_readiness_assessment} and model answers for previous sections were used as input. For \texttt{<demographics response>}, we use the response in Table \ref{table:model_response_fitness_demographics}, for \texttt{<training load response>} we use Table \ref{table:model_response_fitness_load}, for \texttt{<sleep metrics response>}, we use Table \ref{table:model_response_fitness_sleep}, for \texttt{<health metrics response>}, we use Table \ref{table:model_response_fitness_health}. We also incorporate synthetically generated qualitative feedback from the user with \texttt{<subjective readiness>} shown in Table \ref{table:prompt_fitness_case_studies_subjective_readiness} and \texttt{<muscle soreness>} shown in Table \ref{table:prompt_fitness_case_studies_muscle_soreness}.
}
\label{table:model_response_fitness_recommendation} 
\end{table}
\begin{tcolorbox}[title={\ourmodel example response for fitness case study on the assessment section.}]
\textbf{Important insights}

\textbf{Load}

\begin{itemize}
\item
  The trainee's acute training load is 346 and their ACWR is 1.5. This
  is consistent with the last month. This indicates a slightly high ACWR
  which may put the trainee at risk for over training.
\end{itemize}

\textbf{Sleep}

\begin{itemize}
\item
  The trainee's average sleep duration is 6.3 hours with a sleep score
  of 76. This is below the recommended 7-9 hours. The trainee has been
  below the recommended amount of sleep for 22 of the last 30 days.
  Today's sleep was 2.95 hours and a score of 62.9 which is
  significantly lower than the monthly averages.
\item
  The trainee's sleep schedule is inconsistent.
\end{itemize}

\textbf{Health Metrics}

\begin{itemize}
\item
  The trainee's resting heart rate is 66 bpm which is elevated compared
  to the monthly average of 59.03 bpm. This could indicate stress, poor
  sleep, or insufficient recovery.
\item
  The trainee's HRV is 18.17 ms which is decreased compared to the
  monthly average of 24.11 ms. This could indicate stress, poor sleep,
  or insufficient recovery.
\end{itemize}

\textbf{Subjective Readiness + Muscle Soreness}

\begin{itemize}
\item
  The trainee is feeling a bit stressed and fatigued but is managing by
  hydrating and prioritizing recovery.
\item
  The trainee is feeling some soreness in their legs but it is
  manageable.
\end{itemize}

\textbf{Readiness Score}

\begin{itemize}
\item
  2/5
\item
  Explanation: The trainee's load and metrics indicate a need for
  recovery. The trainee's sleep has been poor and inconsistent as well.
  The trainee's subjective assessment also suggests a need for recovery.
\end{itemize}

\textbf{Fitness Recommendations for Today}

\begin{itemize}
\item
  The trainee should prioritize recovery today by taking a rest day or
  doing a light activity.
\item
  The trainee should focus on getting a good night's sleep by going to
  bed and waking up at the same time each day, creating a relaxing
  bedtime routine, and making sure their bedroom is dark, quiet, and
  cool.
\item
  The trainee should manage stress by incorporating stress-reducing
  activities into their day, such as exercise, yoga, or meditation.
\item
  The trainee should ensure they are hydrating adequately by drinking
  plenty of water throughout the day.
\end{itemize}

\textbf{Followup Questions}

\begin{itemize}
\item
  What happened to sleep last night?
\item
  What caused the trainee to feel more stressed and fatigued?
\end{itemize}
\end{tcolorbox}

\newpage
\subsection{Inter-rater expert agreement and rating speeds across primary and secondary raters}
\label{section:appendix:case_studies:primary_vs_secondary}

\begin{figure}[ht]
  \centering
  \includegraphics[width=0.7\textwidth]{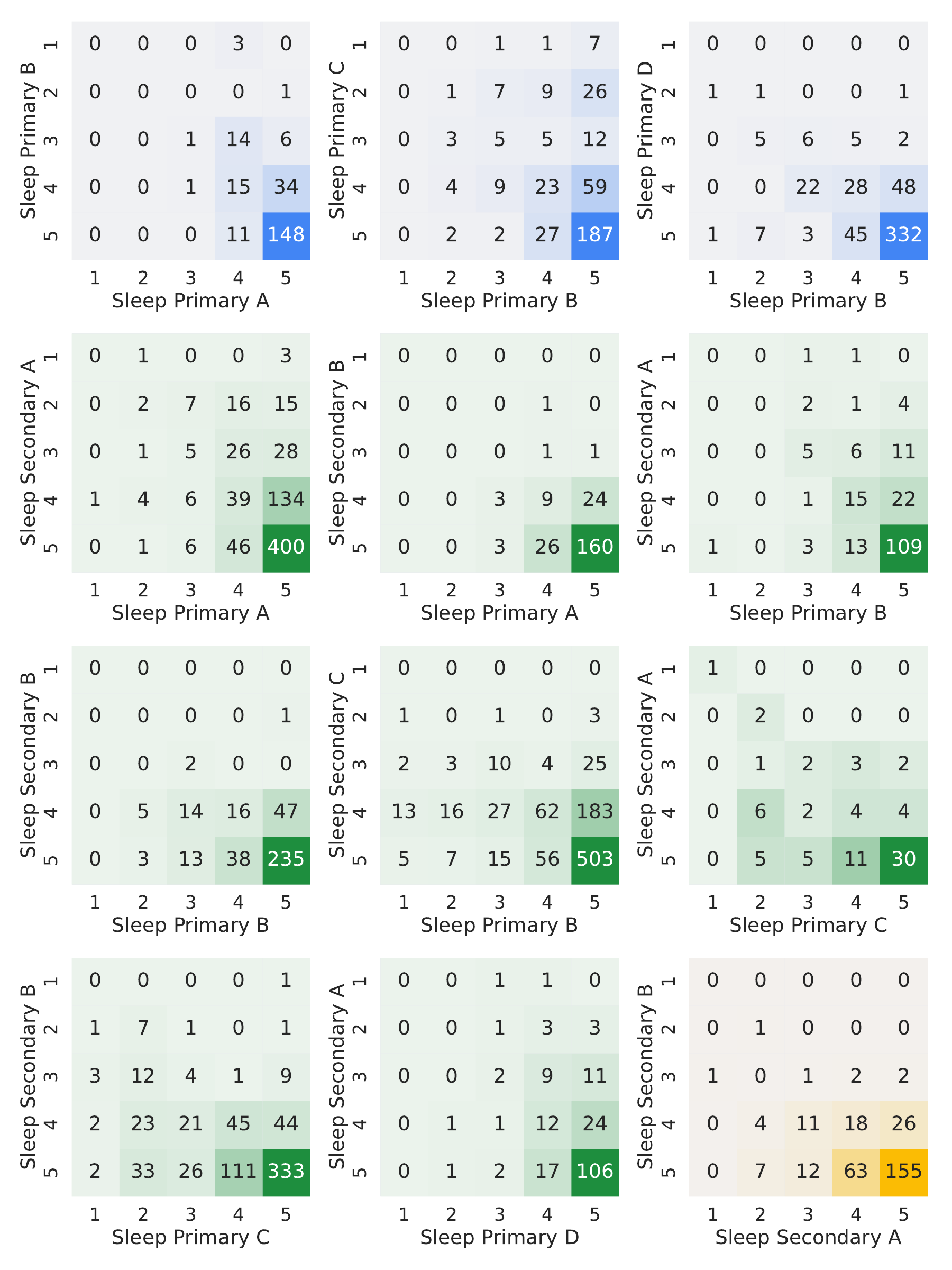}
  \caption{\textbf{Contingency tables showing pairwise rating agreement between raters in the sleep vertical.} Counts are aggregated across all case studies, sections, and principles for each case study for which multiple ratings are available. Blue, primary vs primary raters. Green, primary vs secondary raters. Yellow, secondary vs secondary raters.}
  \label{figure:rater_contingency_sleep_v2_all}
\end{figure}

\begin{figure}[ht]
  \centering
  \includegraphics[width=0.7\textwidth]{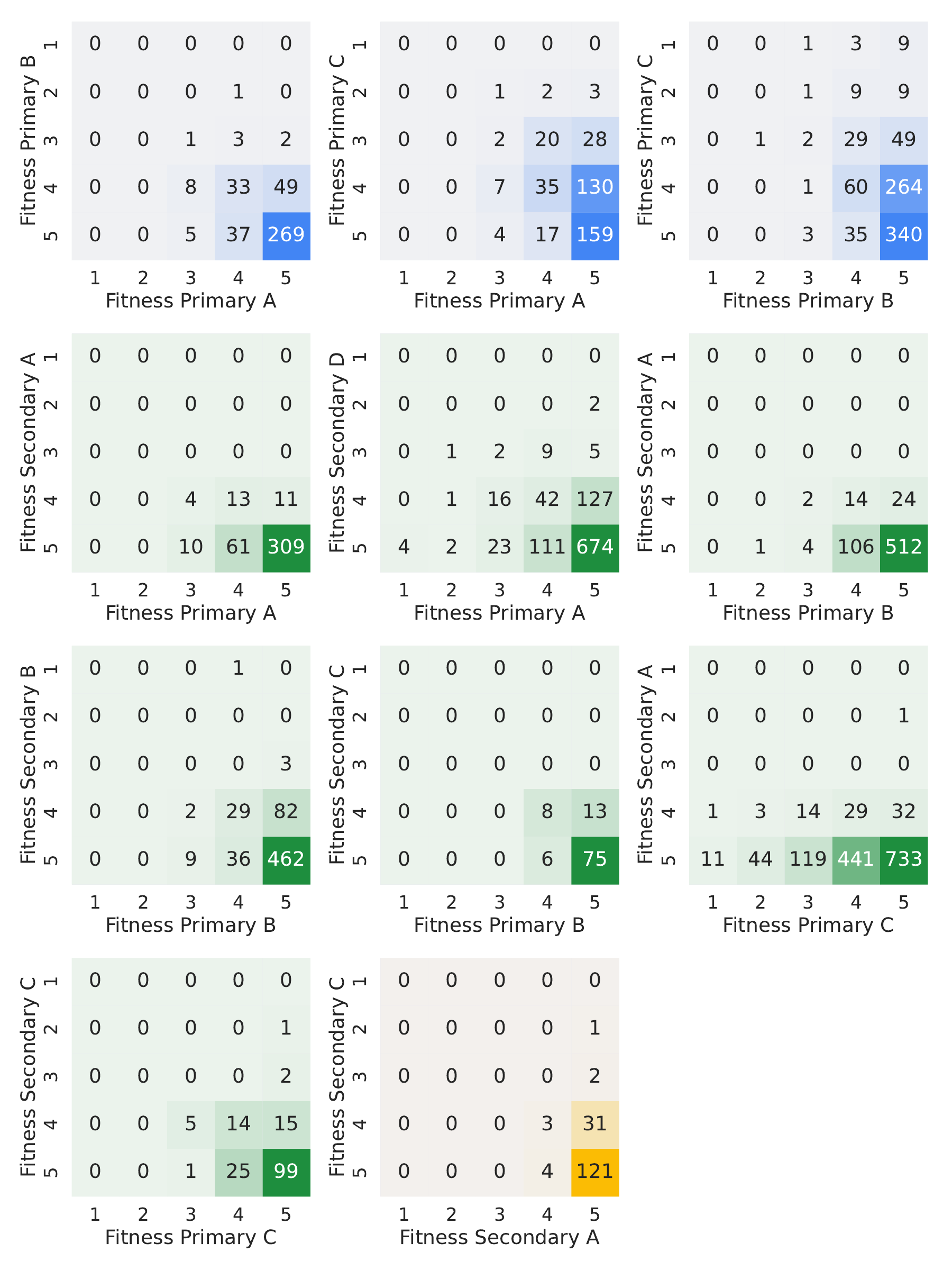}
  \caption{\textbf{Contingency tables showing pairwise rating agreement between raters in the fitness vertical.} Counts are aggregated across all case studies, sections, and principles for each case study for which multiple ratings are available. Blue, primary vs primary raters. Green, primary vs secondary raters. Yellow, secondary vs secondary raters.}
  \label{figure:rater_contingency_readiness_v2_all}
\end{figure}

\begin{figure}[ht]
  \centering
  \includegraphics[width=\textwidth]{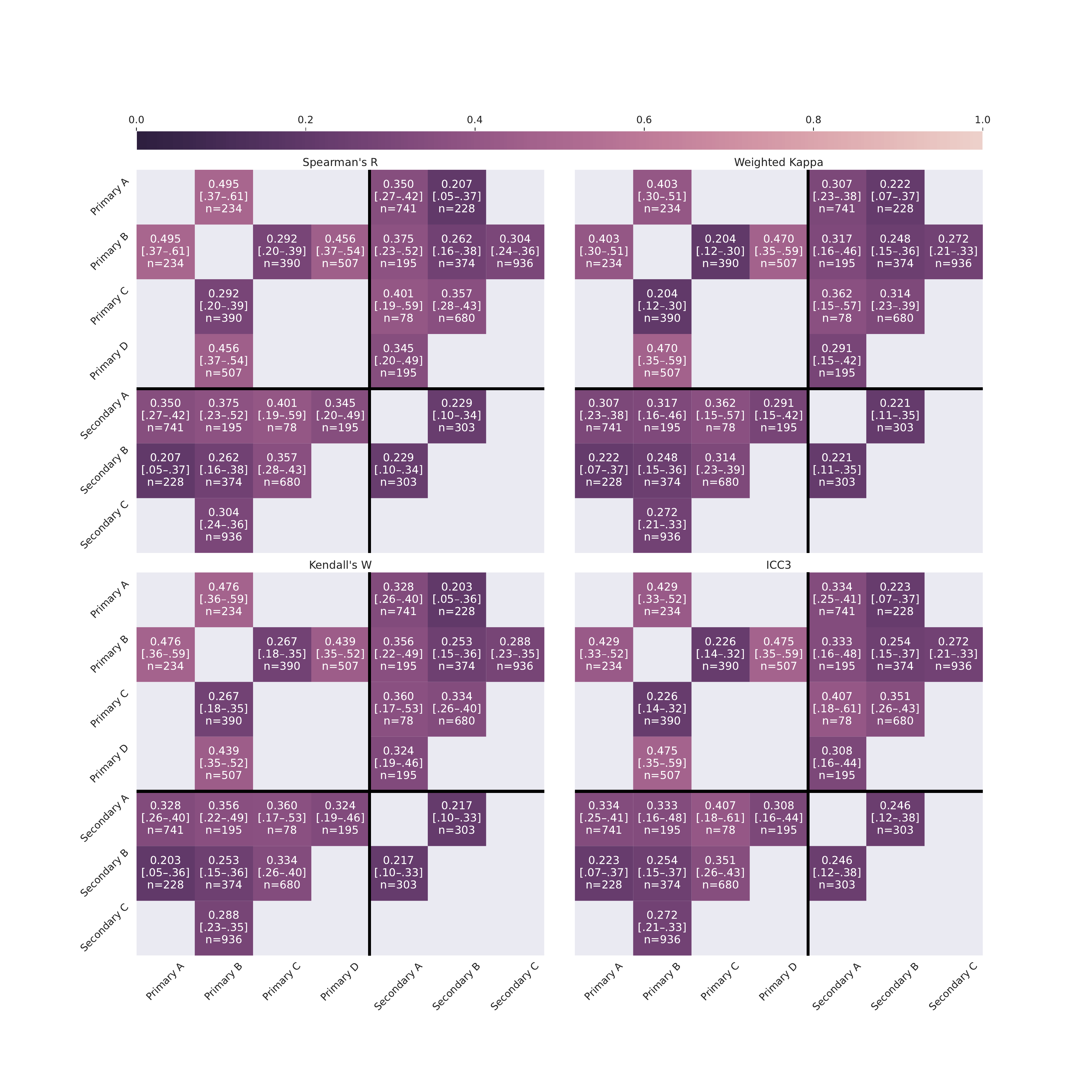}
  \caption{\textbf{Pairwise Spearman's rank correlation, Weighted Cohen's Kappa, Kendall's Coefficient of Concordance (Kendall's W), and Interclass correlation (ICC(3,1)) measuring concordance between primary and secondary raters in the sleep vertical.} Metrics were computed using all ratings for each principle and section across case studies rated by more than one rater. The number of overlapping ratings is denoted by $n$. Mean metrics and 95\% confidence intervals derived from 1,000 bootstrapping iterations are reported for each pair.}
  \label{figure:sleep_rater_agreement_metrics}
\end{figure}

\begin{figure}[ht]
  \centering
  \includegraphics[width=\textwidth]{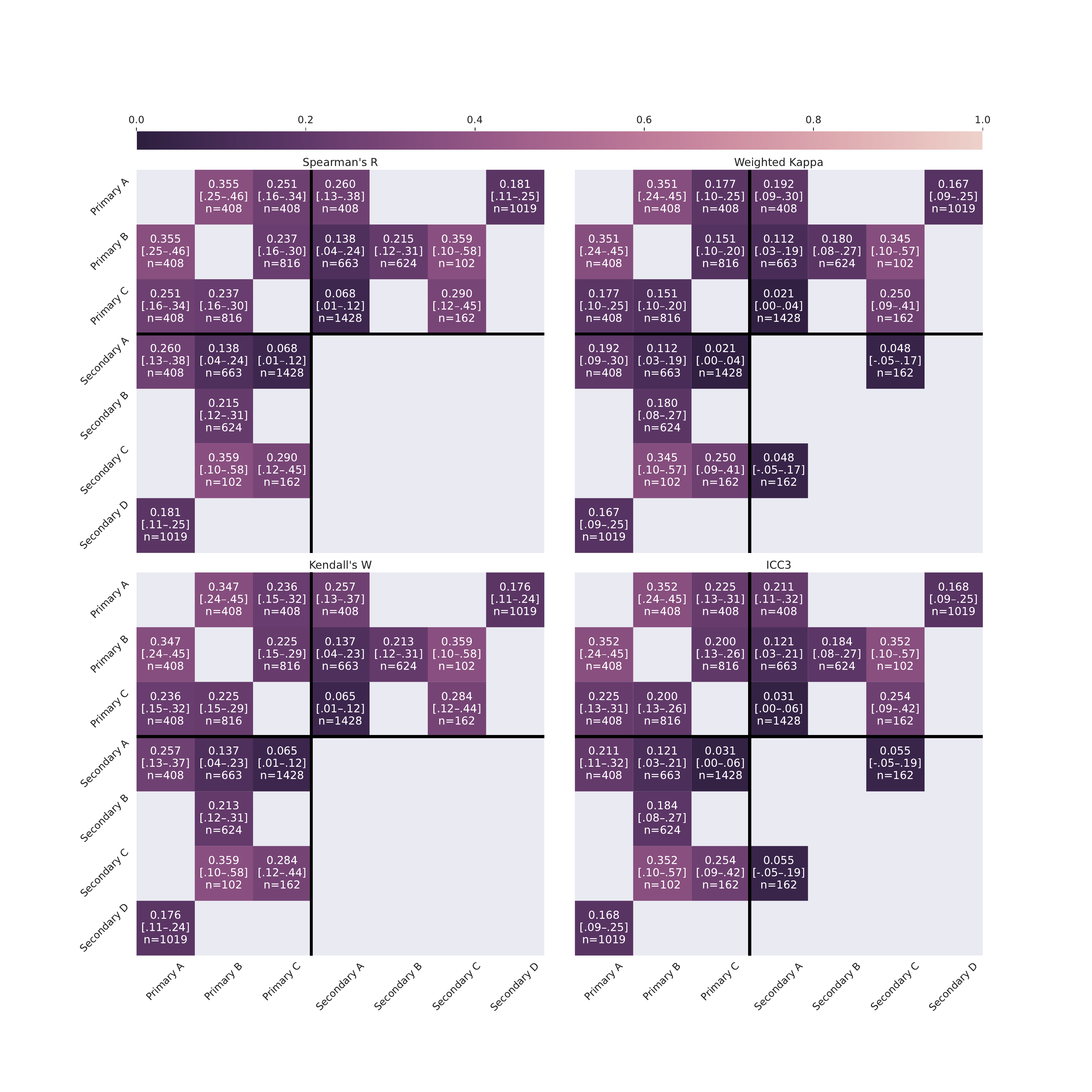}
  \caption{\textbf{Pairwise Spearman's rank correlation, Weighted Cohen's Kappa, Kendall's Coefficient of Concordance (Kendall's W), and Interclass correlation (ICC(3,1)) measuring concordance between primary and secondary raters in the fitness vertical.} Metrics were computed using all ratings for each principle and section across case studies rated by more than one rater. The number of overlapping ratings is denoted by $n$. Mean metrics and 95\% confidence intervals derived from 1,000 bootstrapping iterations are reported for each pair.}
  \label{figure:readiness_rater_agreement_metrics}
\end{figure}

In order to assess agreement between raters and analyze differences between primary and secondary rater groups, we introduced a small amount of overlap across case study rating assignments within each vertical, resulting in anywhere from 78 to 1,428 paired ratings within a subset of raters. Computing pairwise Spearman's rank correlation, Weighted Cohen’s Kappa, Kendall’s Coefficient of Concordance (Kendall’s W), and  Interclass correlation (ICC(3,1)) measures, we generally observe that primary raters tend to have higher measures of agreement with one another than with secondary raters. However, due to low sample sizes, this difference is not significant (see \Cref{figure:rater_contingency_sleep_v2_all,figure:rater_contingency_readiness_v2_all} for contingency tables and \Cref{figure:readiness_rater_agreement_metrics,figure:sleep_rater_agreement_metrics} for agreement measures). We also measured the amount of time it took for each rater to rate all sections and principles for a given case study in minutes. We find that primary raters rate significantly faster than secondary raters in both verticals (\Cref{table:auto_eval_time_to_rate}).

\clearpage
\newpage
\subsection{Automatic evaluation of case studies}

\begin{table}[h]
\captionsetup{labelformat=simple, justification=raggedright, singlelinecheck=false, labelsep=pipe} 
\caption{\textbf{Generic prompting format for AutoEval model fine-tuning and inference.} Here, the ``\{VERTICAL\}'' placeholder is either sleep or fitness and ``\{TAG\_LABEL\}'' is the case study section being rated (e.g., ``Etiology''). ``\{OBJECTIVE\}'', ``\{PRINCIPLE\_CRITERIA\}'', and ``\{OPTION\_N\}'' refer to the evaluation criteria for the given section and principle as outlined in \Cref{section:methods:expert_rating}. The ``\{RATING\}'' and ``\{RATING\_TEXT\_DESC\}'' placeholders are replaced with the rating assigned by the expert raters. Finally, ``\{CASE\_STUDY\_DATA\}'' is replaced with a high level overview of the case study while ``\{ASSISTANT\_TEXT\}'' is replaced with the case study response from an expert or model.}
\label{table:autoeval_prompt_format} 
\end{table}
\begin{tcolorbox}[title={Generic prompting format for AutoEval model fine-tuning and inference}]
\textbf{Prompt:} \\
You are a \{VERTICAL\} expert rater assessing the quality of health suggestions made by another personal health expert using the criteria below. \\
\newline
We use the following definitions when describing evaluation objectives: \\
\newline
- Personalization: The act of taking into account the users data to describe relationships, causes, or to add/implement domain knowledge. \\
- Important Data: Useful for accomplishing the objective. \\
- Unimportant Data: Not useful for accomplishing the objective. \\
- Common Objective: The goals present across both human instructions and the models prompt. Goals only performed by the model or only performed by the human rater are not included. \\
\newline
You are rating the quality of a \{VERTICAL\} ``\{TAG\_LABEL\}'' response. The objective is: ``\{OBJECTIVE\}''. \\
\newline
The other \{VERTICAL\} expert was provided the following user data and information: \\
\newline
\`{}\`{}\`{} \\
\{CASE\_STUDY\_DATA\} \\
\`{}\`{}\`{} \\
\newline
The response from the other \{VERTICAL\} expert is: \\
\newline
\`{}\`{}\`{} \\
\{ASSISTANT\_TEXT\} \\
\`{}\`{}\`{}\\
\newline
Grade this response using the following criteria and Likert scale statements: \\
\newline
Criteria: \{PRINCIPLE\_CRITERIA\} \\
\newline
1. \{OPTION\_1\} \\
2. \{OPTION\_2\} \\
3. \{OPTION\_3\} \\
4. \{OPTION\_4\} \\
5. \{OPTION\_5\} \\
\newline
State only the numeric score and option text when providing your rating. The formatting of your response must match that of the Likert scale statement. \\
\newline
\textbf{Target:} \\
\{RATING\}. \{RATING\_TEXT\_DESC\} \\
\end{tcolorbox}

\newpage
\begin{table}[h]
\captionsetup{labelformat=simple, justification=raggedright, singlelinecheck=false, labelsep=pipe} 
\caption{\textbf{Prompting format for AutoEval model fine-tuning and inference in the sleep vertical for the ``Recommendations'' section and the ``No incorrect domain knowledge'' principle.} Here, ``\{CASE\_STUDY\_DATA\}'' is replaced with a high level overview of the case study while ``\{ASSISTANT\_TEXT\}'' is replaced with the case study response from an expert or model.}
\label{table:autoeval_prompt_format_example} 
\end{table}
\begin{tcolorbox}[title={Prompting format for AutoEval model fine-tuning and inference in the sleep vertical for the ``Recommendations'' section and the ``No incorrect domain knowledge'' principle}]
\textbf{Prompt:} \\
You are a sleep expert rater assessing the quality of health suggestions made by another personal health expert using the criteria below. \\
\newline
We use the following definitions when describing evaluation objectives: \\
\newline
- Personalization: The act of taking into account the users data to describe relationships, causes, or to add/implement domain knowledge. \\
- Important Data: Useful for accomplishing the objective. \\
- Unimportant Data: Not useful for accomplishing the objective. \\
- Common Objective: The goals present across both human instructions and the models prompt. Goals only performed by the model or only performed by the human rater are not included. \\
\newline
You are rating the quality of a sleep ``Recommendations'' response. The objective is: ``Provide recommendations to the user that can help them improve their sleep by addressing potential causes identified in the Etiology section. Avoid providing generic recommendations that are not personalized. This section does not require specific data to be cited directly, but the interpretation used to justify the recommendation should be present.''. \\
\newline
The other sleep expert was provided the following user data and information: \\
\newline
\`{}\`{}\`{} \\
\{CASE\_STUDY\_DATA\} \\
\`{}\`{}\`{} \\
\newline
The response from the other sleep expert is: \\
\newline
\`{}\`{}\`{} \\
\{ASSISTANT\_TEXT\} \\
\`{}\`{}\`{} \\
\newline
Grade this response using the following criteria and Likert scale statements: \\
\newline
Criteria: This section does not contain evidence of incorrect domain knowledge (e.g., factually incorrect or not accepted by expert consensus). \\
\newline
1. Only incorrect domain knowledge is referenced. \\
2. Many incorrect domain knowledge references exist. \\
3. Several incorrect domain knowledge references exist. \\
4. A few incorrect domain knowledge references exist. \\
5. No incorrect domain knowledge references exist. \\
\newline
State only the numeric score and option text when providing your rating. The formatting of your response must match that of the Likert scale statement. \\
\newline
\textbf{Target:} \\
5. No incorrect domain knowledge references exist.
\end{tcolorbox}

\begin{table}[htp]
\setlength{\tabcolsep}{3pt}
\caption{\textbf{AutoEval model performance in the validation set across verticals.} AutoEval model rating predictions are compared with ground truth human ratings from the validation dataset. Here, ``Gemini Pro'' denotes an untuned baseline, ``Primary'' denotes models tuned on only one expert's ratings, and ``All'' denotes models tuned on all ratings for the given vertical. Spearman's rank correlation, Kendall's Coefficient of Concordance (Kendall's W), and Weighted Cohen's Kappa measurements were computed using all ratings for each principle and section. Mean metrics and 95\% confidence intervals derived from 1,000 bootstrapping iterations are reported for each pair. Using paired bootstrapping, we find that all tuned AutoEval models significantly outperform the untuned baseline across metrics. However, due to low sample size, differences between tuned AutoEval models are not statistically significant.}
\label{table:autoeval_validation_performance}
\centering
\adjustbox{max width=\textwidth}{
\begin{tabular}{lrrrr}
\toprule
\textbf{AutoEval Model} & \textbf{Spearman's R} & \textbf{Kendall's W} & \textbf{Weighted Kappa} \\
\midrule
\textbf{Fitness} &  &  &  \\
Gemini Pro & 0.205 (0.152--0.256) & 0.198 (0.147--0.248) & 0.203 (0.152--0.256) \\
All & 0.280 (0.228--0.329) & 0.274 (0.224--0.322) & 0.152 (0.114--0.195) \\
Fitness Primary B & 0.284 (0.230--0.336) & 0.278 (0.225--0.329) & 0.142 (0.106--0.182) \\
Fitness Primary C & 0.305 (0.256--0.352) & 0.291 (0.245--0.335) & 0.320 (0.270--0.369) \\
\midrule
\textbf{Sleep} &  &  &  \\
Gemini Pro & 0.242 (0.185--0.299) & 0.230 (0.177--0.283) & 0.223 (0.166--0.280) \\
Sleep Primary D & 0.333 (0.279--0.384) & 0.316 (0.265--0.365) & 0.321 (0.256--0.389) \\
All & 0.341 (0.288--0.395) & 0.325 (0.273--0.378) & 0.347 (0.272--0.424) \\
Sleep Primary C & 0.368 (0.320--0.417) & 0.343 (0.298--0.388) & 0.389 (0.338--0.443) \\
\bottomrule
\end{tabular}
}
\end{table}

 \begin{figure}[htp]
  \centering
    \includegraphics[width=\columnwidth]{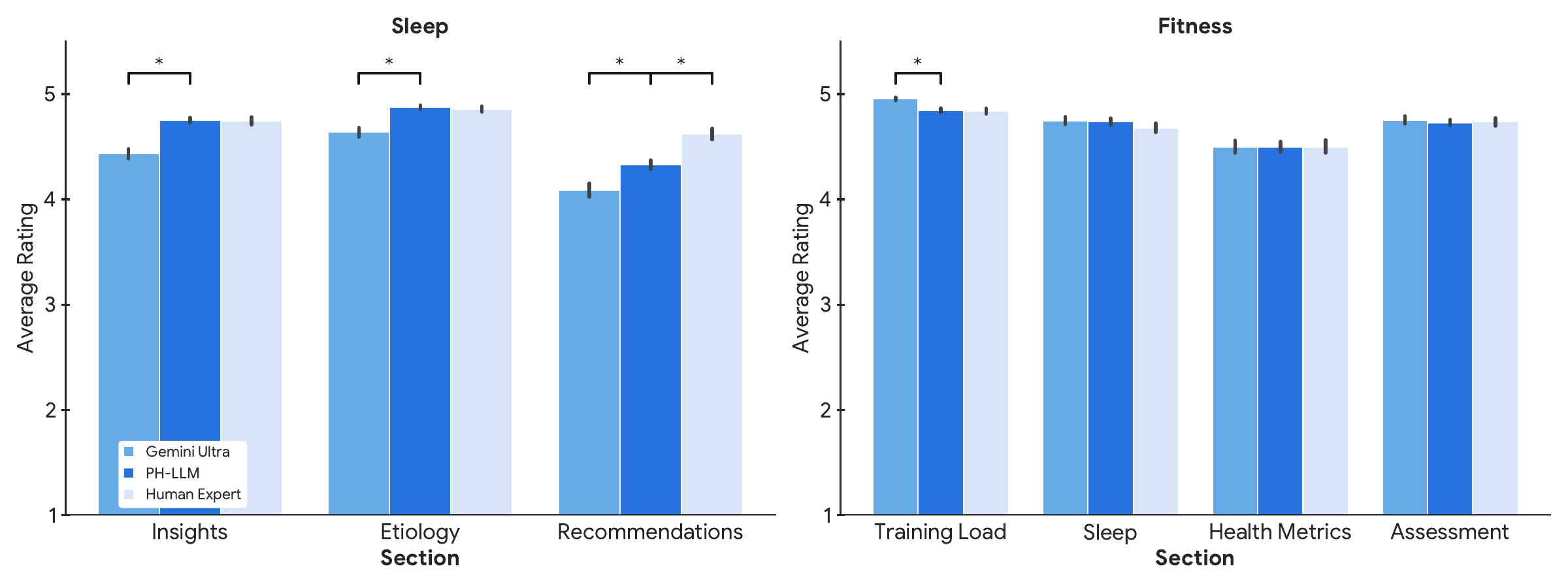}
    \caption{\textbf{Case Study AutoEval Evaluation Results.} Mean ratings given by AutoEval models tuned using ratings from high variance raters for the case study subsections across the \textbf{(Left)} sleep and \textbf{(Right)} fitness domains. ``$*$'' indicates a  statistically significant difference between two response types after multiple hypothesis testing correction.}
    \label{fig:autoeval_ratings_high_var_raters}
\end{figure}

\begin{figure}[htp]
  \centering
    \includegraphics[width=\columnwidth]{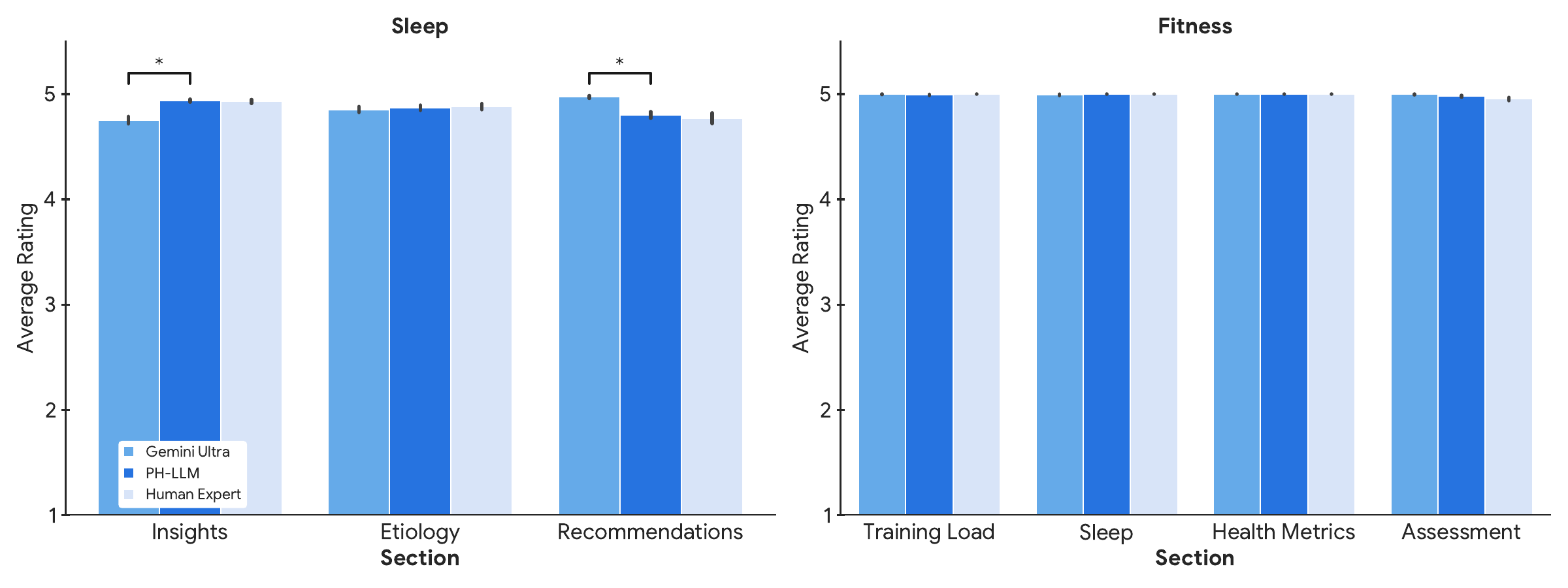}
    \caption{\textbf{Case Study AutoEval Evaluation Results.} Mean ratings given by AutoEval models tuned using ratings from low variance raters for the case study subsections across the \textbf{(Left)} sleep and \textbf{(Right)} fitness domains. ``$*$'' indicates a  statistically significant difference between two response types after multiple hypothesis testing correction.}
    \label{fig:autoeval_ratings_low_var_raters}
\end{figure}

\begin{table}[ht]
\caption{\textbf{Average time (m) taken to rate all responses for a single case across all sections and principles}. We consider primary raters, secondary raters, a single model, and a fully parallelized set of models, where ``fully parallelized'' means serving a model replica for each combination of sections and principles. The mean time-to-rate and the corresponding 95\% confidence intervals were calculated across 1,000 bootstrapping iterations.}
\label{table:auto_eval_time_to_rate}
\centering
\begin{tabular}{lll}
\toprule
\textbf{Rater} & \textbf{Fitness} & \textbf{Sleep} \\
\midrule
Primary raters & 44.7 (38.6--51.2) & 26.8 (25.2--28.6) \\
Secondary raters & 87.7 (62.4--115.9) & 45.7 (34.3--60.9) \\
Model (single) & 27.5 (26.9--28.1) & 24.7 (23.7--25.6) \\
Model (Fully parallelized) & ~~0.367 (0.361--0.373) & ~~0.411 (0.402--0.421) \\
\bottomrule
\end{tabular}
\end{table}

\clearpage
\setcounter{table}{0}
\setcounter{figure}{0}
\section{Professional Examinations}
\label{app:mcqs}

\subsection{Additional ablation experiments}
\begin{table}[htp]
\caption{\textbf{Effects of chain-of-thought prompting and self-consistency on \ourmodel Accuracy for Sleep and Fitness Professional Exams.} CoT=Chain-of-Thought, SC=Self-Consistency.}
\centering
\label{table:mcq_ablation_cot_sc}
\begin{tabular}{llcc}
\toprule[1.5pt]
\textbf{Domain} & \textbf{Prompt} & SC & No SC \\
\midrule
\multirow{2}{*}{Sleep} & CoT & \textbf{\textcolor{forestgreen}{79\%}} & 78\% \\[1.2pt]
      & No CoT & \textbf{\textcolor{forestgreen}{79\%}} & \textbf{\textcolor{forestgreen}{79\%}} \\[1.2pt]
\midrule
\multirow{2}{*}{Fitness} & CoT & \textbf{\textcolor{forestgreen}{88\%}} & 84\% \\[1.2pt]
& No CoT & \textbf{\textcolor{forestgreen}{87\%}} & 85\% \\[1.2pt]
\bottomrule[1.5pt]
\end{tabular}
\end{table}

\begin{table}[htp]
\caption{\textbf{Effects of Few-Shot prompting on \ourmodel Accuracy for Sleep and Fitness Professional Exams.}}
\centering
\label{table:mcq_ablation_few_shot}
\begin{tabular}{llc}
\toprule[1.5pt]
\textbf{Domain} & \textbf{Prompt} & SC + CoT \\
\midrule
\multirow{2}{*}{Sleep} & Few-Shot & \textbf{\textcolor{forestgreen}{79\%}}  \\[1.2pt]
      & Zero-Shot &75\% \\[1.2pt]
\midrule
\multirow{2}{*}{Fitness} & Few-Shot & \textbf{\textcolor{forestgreen}{88\%}} \\[1.2pt]
& Zero-Shot & 87\% \\[1.2pt]
\bottomrule[1.5pt]
\end{tabular}
\end{table}

\subsection{Comparison of Professional Exam Performance on Additional Models.}
\label{appendix:additional_models_mcqs_perf}
 \begin{figure}[h!]
  \centering
    \includegraphics[width=\columnwidth]{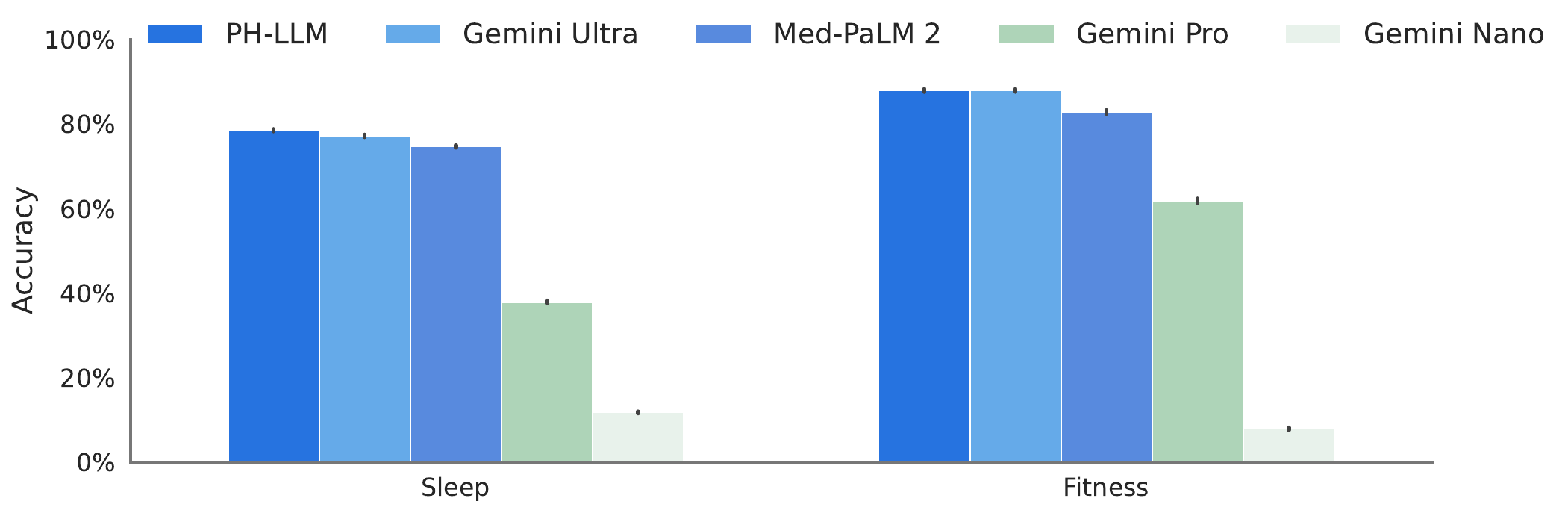}
    \caption{\textbf{Overall performance of professional exams across \ourmodel, different Gemini models, and Med-PaLM 2.} All Gemini model sizes are based on the Gemini 1.0 model family.}
    \label{fig:mcq_scaling_comp}
\end{figure}

 \begin{figure}[htp]
  \centering
    \includegraphics[width=\columnwidth]{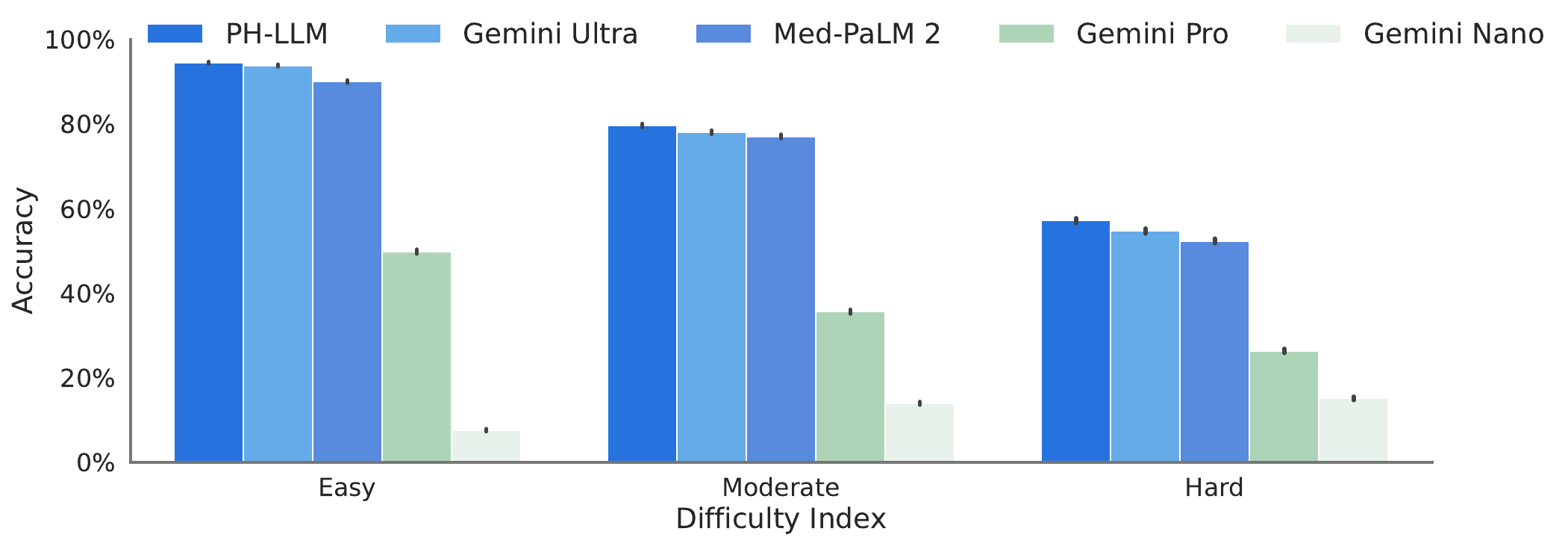}
    \caption{\textbf{Breakdown performance of sleep professional exams across \ourmodel, different Gemini models, and Med-PaLM 2.} All Gemini model sizes are based on the Gemini 1.0 model family.}
    \label{fig:mcq_scaling_comp_stratified}
\end{figure}

\clearpage
\subsection{Prompts Used for Professional Exams}
\begin{table}[h]
\captionsetup{labelformat=simple, justification=raggedright, singlelinecheck=false, labelsep=pipe} 
\caption{\textbf{Prompt for Multiple-Choice Questions with Chain-of-Thought}}
\label{table:prompt_mcq_cot} 
\end{table}
\begin{tcolorbox}[title={Prompt for Multiple-Choice Questions with Chain-of-Thought}]
\textbf{Question:} \color{black} 
\\
Instructions: The following are multiple choice questions about \{domain\} knowledge. Solve them in a step-by-step fashion, starting by summarizing the available information. Output a single option from $\text{\{mcq\_options\}}$  as the final answer and enclosed by xml tags <answer></answer>.

Here are two examples:

$\text{\#\#}$ Question: A 26-year-old female presents asking about jet lag. She has no past medical history, lives on the East Coast, and travels frequently to the West Coast for business. The person's career involves planning evening events, and she reports significant sleepiness at these events that impairs her ability to perform her job. She wants to know how she can adapt to Pacific Standard Time (PST) before she travels. What treatment plan will help this patient adapt to PST prior to travel?

(A) Light in evening and later bedtime 1 day before traveling

(B) Light in morning and earlier wake time 3 days before traveling

(C) Light in evening and later bedtime 3 days before traveling

(D) Light in morning and earlier wake time 1 month before traveling

(E) Light in evening and later bedtime 1 month before traveling

Explanation: Let's solve this step-by-step, referring to authoritative sources as needed. The West Coast is 3 timezones behind the East Coast. Since she plans evening events, she needs to shift her schedule to stay up 3 hours later. Adding light in the evening will disrupt melatonin production, delaying sleepiness. Transitioning timezones typically takes one day per timezone.
\newline
Answer: <answer>(C)</answer>
\newline
$\text{\#\#}$ Question: What is a difference in the clinical features of obstructive sleep apnea (OSA) in older adults compared to younger adults?

(A) Increased prevalence of OSA among older adults occurs after age 65.

(B) Clinical symptoms associated with OSA (e.g. excessive daytime
sleepiness) are less common and less severe in older adults than in younger adults.

(C) Increased risk of cardiopulmonary diseases is greater among elderly than among younger individuals.

(D) Excess body weight, snoring, and witnessed apneas more consistently indicate OSA in older adults than in younger individuals.

(E) There are no significant OSA differences between older and younger adults.

Explanation: Let's solve this step-by-step, referring to authoritative sources as needed. Compared to younger patients with the same apnea hypopnea index, OSA in older patients is associated with less sleepiness (Morrell et al 2012). This observation has led some to suggest that OSA in the elderly may represent a distinct physiological phenotype.
Answer: <answer>(B)</answer>
\newline
$\text{\#\#}$ Question: $\text{\{mcq\_question\}}$
\newline
Explanation: Let us solve this step-by-step, referring to authoritative sources as needed. 
\end{tcolorbox}

\newpage
\begin{table}[h]
\captionsetup{labelformat=simple, justification=raggedright, singlelinecheck=false, labelsep=pipe} 
\caption{\textbf{Prompt for Multiple-Choice Questions without Chain-of-Thought}}
\label{table:prompt_mcq_no_cot} 
\end{table}
\begin{tcolorbox}[title={Prompt for Multiple-Choice Questions without Chain-of-Thought}]
\textbf{Question:} \color{black} 
\\
Instructions: The following are multiple choice questions about \{domain\} knowledge. Output a single option from $\text{\{mcq\_options\}}$  as the final answer and enclosed by xml tags <answer></answer>.

Here are two examples:

$\text{\#\#}$ Question: A 26-year-old female presents asking about jet lag. She has no past medical history, lives on the East Coast, and travels frequently to the West Coast for business. The person's career involves planning evening events, and she reports significant sleepiness at these events that impairs her ability to perform her job. She wants to know how she can adapt to Pacific Standard Time (PST) before she travels. What treatment plan will help this patient adapt to PST prior to travel?

(A) Light in evening and later bedtime 1 day before traveling

(B) Light in morning and earlier wake time 3 days before traveling

(C) Light in evening and later bedtime 3 days before traveling

(D) Light in morning and earlier wake time 1 month before traveling

(E) Light in evening and later bedtime 1 month before traveling

Answer: <answer>(C)</answer>
\newline
$\text{\#\#}$ Question: What is a difference in the clinical features of obstructive sleep apnea (OSA) in older adults compared to younger adults?

(A) Increased prevalence of OSA among older adults occurs after age 65.

(B) Clinical symptoms associated with OSA (e.g. excessive daytime
sleepiness) are less common and less severe in older adults than in younger adults.

(C) Increased risk of cardiopulmonary diseases is greater among elderly than among younger individuals.

(D) Excess body weight, snoring, and witnessed apneas more consistently indicate OSA in older adults than in younger individuals.

(E) There are no significant OSA differences between older and younger adults.

Answer: <answer>(B)</answer>
\newline
$\text{\#\#}$ Question: $\text{\{mcq\_question\}}$
\end{tcolorbox}

\begin{table}[h]
\captionsetup{labelformat=simple, justification=raggedright, singlelinecheck=false, labelsep=pipe} 
\caption{\textbf{Prompt for Multiple-Choice Questions with Chain-of-Thought and Zero-Shot}}
\label{table:prompt_mcq_no_cot_zero_shot} 
\end{table}
\begin{tcolorbox}[title={Prompt for Multiple-Choice Questions with Chain-of-Thought and Zero-Shot}]
\textbf{Question:} \color{black} 
\newline
Instructions: The following are multiple choice questions about $\text{\{domain\}}$ knowledge. Solve them in a step-by-step fashion, starting by summarizing the available information. Output a single option from $\text{\{mcq\_options\}}$ as the final answer and enclosed by xml tags <answer></answer> such as <answer>(A)</answer>. \newline
$\text{\#\#}$ Question: $\text{\{mcq\_question\}}$

\end{tcolorbox}

\newpage

\subsection{Annotated examples of professional exam questions and responses}
\begin{table}[h]
\captionsetup{labelformat=simple, justification=raggedright, singlelinecheck=false, labelsep=pipe} 
\caption{\textbf{Examples correctly answered by \ourmodel.}}
\label{table:mcq_correct_examples} 
\end{table}
\begin{tcolorbox}[title={Examples correctly answered by \ourmodel }]
\textbf{\underline{Sleep Question:}} \color{black} 
\\
A 26-year-old woman inquires about managing jet lag. She has no medical history and resides on the East Coast, frequently traveling to the West Coast for work. Her job involves organizing evening events, and she experiences considerable drowsiness during these events, hindering her job performance. She seeks advice on how to adjust to Pacific Standard Time (PST) before her trips. What strategy will assist her in acclimating to PST before she travels?

(A) Light in evening and later bedtime 1 day before traveling

(B) Light in morning and earlier wake time 3 days before traveling 

(C) Light in evening and later bedtime 3 days before traveling

(D) Light in morning and earlier wake time 1 month before traveling 

(E) Light in evening and later bedtime 1 month before traveling

\textbf{Model Response:} 

The West Coast is 3 timezones behind the East Coast. Since she plans evening events, she needs to shift her schedule to stay up 3 hours later. Adding light in the evening will disrupt melatonin production, delaying sleepiness. Transitioning timezones typically takes one day per timezone. Therefore, she needs to start the treatment 3 days before traveling. Answer: <answer>(C)</answer>.
\newline
\textcolor{forestgreen}{\textbf{{Expert Comment:}} "The model identifies the differential in circadian phase (transitioning 3 time zones) and the needed direction of the phase shift (a delay), incorporates multiple essential concepts about circadian physiology (e.g., light's effect on melatonin based on timing), and establishes an appropriate plan for adjustment."}

\vspace{\baselineskip}
\textbf{\underline{Fitness Question:}} \color{black} 
\\
When a male gymnast performs a push-up, which muscle action occurs in the erector spinae muscle group?

(A) Concentric 

(B) Eccentric 

(C) Isometric 

(D) Myocentric

\textbf{Model Response:} 

The erector spinae muscle group is located in the back and helps to extend the spine. When a gymnast performs a push-up, the erector spinae muscles contract to help stabilize the spine and keep it in a neutral position. This type of muscle action, where the muscle contracts and remains the same length, is called an isometric contraction.Answer: <answer>(C)</answer>.
\newline
\textcolor{forestgreen}{\textbf{Expert Comment:} "Not only is this a good question the model got correct, its rationale and understanding of the muscle group and contraction type are also clear. "}

\end{tcolorbox}

\newpage
\begin{table}[h]
\captionsetup{labelformat=simple, justification=raggedright, singlelinecheck=false, labelsep=pipe} 
\caption{\textbf{Examples incorrectly answered by \ourmodel.}}
\label{table:mcq_incorrect_examples} 
\end{table}
\begin{tcolorbox}[title={Examples incorrectly answered by \ourmodel }]
\textbf{\underline{Sleep Question:}} \color{black} 
\\
A patient undergoes polysomnography. During quiet wakefulness, no posterior alpha rhythm is observed when the eyes are closed. Which of the following statements is correct?

(A) With no clear posterior dominant alpha rhythm, sleep stage N1 cannot be definitively determined. 

(B) If average background frequency during wakefulness is 8Hz, stage N1 can be defined only if activity decreases to $\leq$ 6Hz. 

(C) Stage N1 can be scored if EEG demonstrates sharply contoured waves that are most prominent over central region and last <0.5 seconds. 

(D) Sleep stage N1 can be scored with rapid eye movement. 

(E) Sleep stage N1 can be scored with a brief series of waves at 12-14Hz, lasting 2 seconds and most prominent in central derivation.
\newline
\textbf{Model Response:} 

The absence of alpha rhythm during quiet wakefulness suggests that the patient does not have a clear posterior dominant alpha rhythm. This makes it difficult to definitively determine sleep stage N1, as the presence of alpha rhythm is one of the key criteria for scoring N1. Answer: <answer>(A)</answer>
\newline
\textbf{Correct Answer: } (C)
\newline
\textcolor{forestgreen}{\textbf{Expert Comment:}"This incorrect answer highlights 2 primary issues: 1. A lack of appropriate factual knowledge and/or application of the alternate criteria outlined in The AASM Manual for the Scoring of Sleep and Associated Events and 2. A lack of “common sense” reasoning that recognizes that the ~10\% of people without a posterior dominant rhythm also have N1 sleep (implying there must be other ways of scoring it)."}

\vspace{\baselineskip}

\textbf{\underline{Fitness Question:}} \color{black} 
\\
What shoulder movements and planes of action are involved in a tennis backhand?

(A) Adduction/frontal 

(B) Abduction/frontal 

(C) Adduction/transverse 

(D) Abduction/transverse

\textbf{Model Response:} 
The tennis backhand involves horizontal adduction, which is movement toward the body in the transverse plane. Answer: <answer>(C)</answer>
\newline
\textbf{Correct Answer: } (D)
\newline
\textcolor{forestgreen}{\textbf{Expert Comment: }"This indicates that the model is working out the mechanics. It gets the planes of movement, and terms of movement,  but missed the 'backhand movement' that changes answer to abduction. It simply missed the most human element of the question."}
\end{tcolorbox}

\clearpage
\setcounter{table}{0}
\setcounter{figure}{0}
\section{Patient-Reported Outcomes}

\subsection{Patient-reported outcome prediction input features}
\begin{table}[h]
\centering
\caption{\textbf{Sensor features used to predict each patient-reported outcome.}}
\label{table:pro_sensor_features} 
\begin{tabular}{ll}
\toprule[1.5pt]
\textbf{Sensor Feature} & \textbf{Definition} \\ 
\midrule
Heart rate variability (rmssd) & Heart rate variability root mean square of successive differences \\
Respiratory rate (rate\_brpm) & Respiratory rate breaths per minute \\
Resting heart rate (rhr\_bpm) & Resting heart rate beats per minute \\
Awake minutes (awake\_minutes) & Awake minutes \\
Deep sleep minutes (deep\_sleep\_minutes) & Deep sleep minutes \\
Sleep duration (duration\_minutes) & Sleep duration minutes \\
Sleep efficiency (efficiency) & The fraction of time in bed that was spent sleeping \\
Overall sleep score (overall\_score) & Overall sleep score \\
Percent of sleep in REM (rem\_sleep\_percent) & Percent of sleep in REM \\
Restlessness (restlessness) & Restlessness \\
Revitalization score (revitalization\_score) & Revitalization score \\
Sleep end time (sleep\_end\_time) & Sleep end time encoded as minutes after midnight \\
Sleep start time (sleep\_start\_time) & Sleep start time encoded as minutes after midnight \\
Sleep time (sleep\_time\_minutes) & Sleep time minutes \\
Awake state minutes (waso\_count\_long\_wakes) & Total number of minutes in awake state after sleep onset \\
Number steps (num\_steps) & Number of steps walked during the day \\
Cardio minutes (cardio\_minutes) & Number of minutes spent in cardio zone during the day \\
Fat burn minutes (fat\_burn\_minutes) & Number of minutes spent in fat burn zone during the day \\
Peak minutes (peak\_minutes) & Number of minutes spent in peak zone during the day \\
Total exercise time (total\_multiplied\_minutes) & Total multiplied minutes of exercise during the day \\
\bottomrule[1.5pt]
\end{tabular}
\end{table}

 \begin{figure}[htp]
  \centering
    \includegraphics[width=0.7\columnwidth]{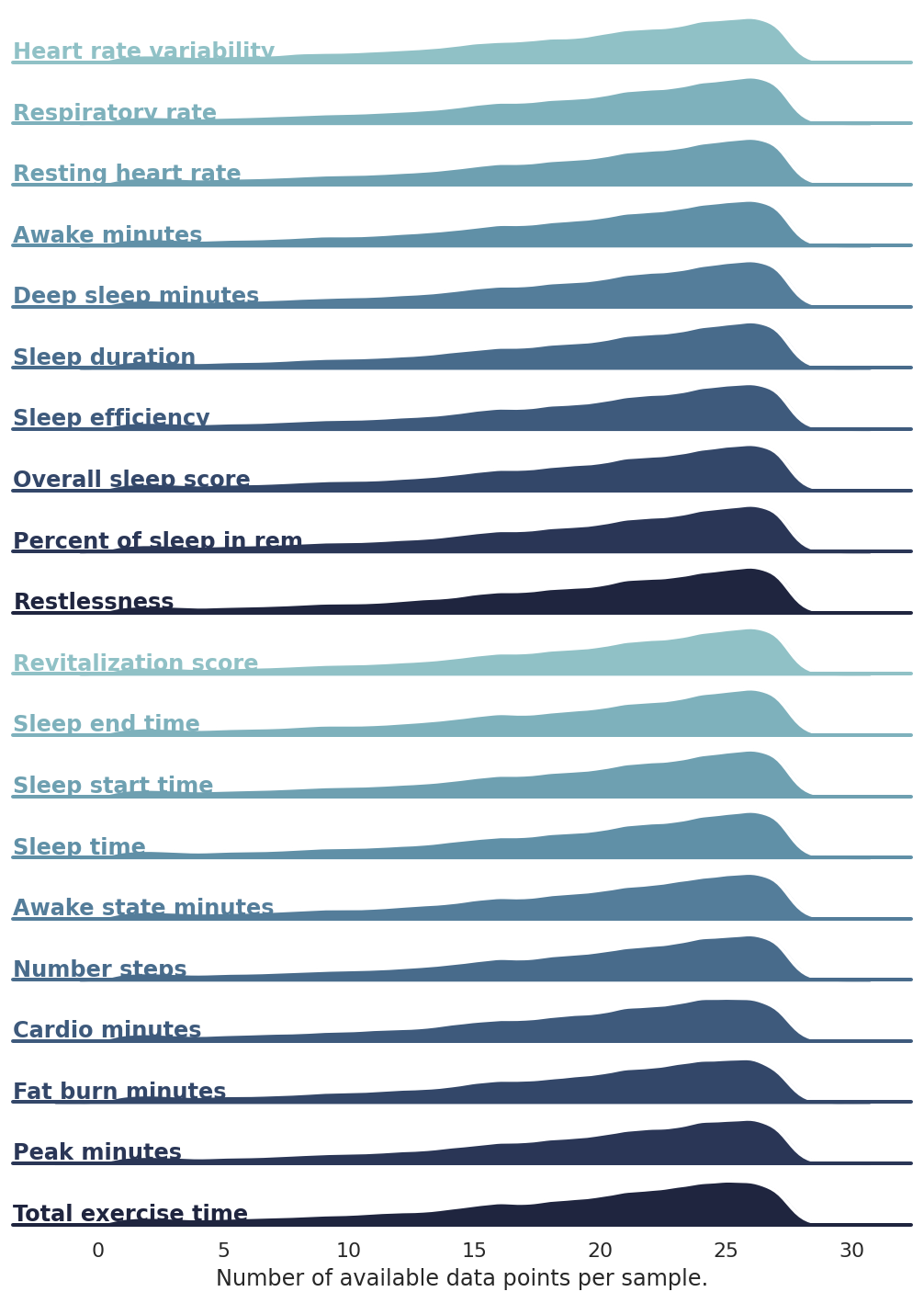}
    \caption{\textbf{Distribution of number of data points available for each sensor.}}
    \label{fig:supp:sensor_distributions}
\end{figure}
\newpage

\subsection{Patient-reported outcome surveys}
\label{sec:pro_surveys}

Each survey is coded so that higher values correspond with greater sleep disturbance or impairment.

\subsubsection*{Sleep Disturbance Survey}

In the past 7 days, my sleep was restless. [Very Restless]
\begin{longtable}{
>{\centering\arraybackslash}p{\thin{}} >{\centering\arraybackslash}p{\thin{}} >{\centering\arraybackslash}p{\thin{}} >{\centering\arraybackslash}p{\thin{}} >{\centering\arraybackslash}p{\thin{}}
}
1. Not at all & 2. A little bit & 3. Somewhat & 4. Quite a bit & 5. Very much \\
\st{} \\ 
\end{longtable}
    
In the past 7 days, I was satisfied with my sleep. [Satisfied]
\begin{longtable}{
>{\centering\arraybackslash}p{\thin{}} >{\centering\arraybackslash}p{\thin{}} >{\centering\arraybackslash}p{\thin{}} >{\centering\arraybackslash}p{\thin{}} >{\centering\arraybackslash}p{\thin{}}
}
5. Not at all & 4. A little bit & 3. Somewhat & 2. Quite a bit & 1. Very much \\
\st{} \\ 
\end{longtable}
    
In the past 7 days, my sleep was refreshing. [Refreshed]
\begin{longtable}{
>{\centering\arraybackslash}p{\thin{}} >{\centering\arraybackslash}p{\thin{}} >{\centering\arraybackslash}p{\thin{}} >{\centering\arraybackslash}p{\thin{}} >{\centering\arraybackslash}p{\thin{}}
}
5. Not at all & 4. A little bit & 3. Somewhat & 2. Quite a bit & 1. Very much \\
\st{} \\
\end{longtable}
    
In the past 7 days, I had difficulty falling asleep. [Trouble falling asleep]
\begin{longtable}{
>{\centering\arraybackslash}p{\thin{}} >{\centering\arraybackslash}p{\thin{}} >{\centering\arraybackslash}p{\thin{}} >{\centering\arraybackslash}p{\thin{}} >{\centering\arraybackslash}p{\thin{}}
}
1. Not at all & 2. A little bit & 3. Somewhat & 4. Quite a bit & 5. Very much \\
\st{} \\
\end{longtable}
    
In the past 7 days, I had trouble staying asleep. [Trouble staying asleep]
\begin{longtable}{
>{\centering\arraybackslash}p{\thin{}} >{\centering\arraybackslash}p{\thin{}} >{\centering\arraybackslash}p{\thin{}} >{\centering\arraybackslash}p{\thin{}} >{\centering\arraybackslash}p{\thin{}}
}
1. Never & 2. Rarely & 3. Sometimes & 4. Often & 5. Always \\
\st{} \\ 
\end{longtable}
    
In the past 7 days, I had trouble sleeping. [Trouble sleeping]
\begin{longtable}{
>{\centering\arraybackslash}p{\thin{}} >{\centering\arraybackslash}p{\thin{}} >{\centering\arraybackslash}p{\thin{}} >{\centering\arraybackslash}p{\thin{}} >{\centering\arraybackslash}p{\thin{}}
}
1. Never & 2. Rarely & 3. Sometimes & 4. Often & 5. Always \\
\st{} \\ 
\end{longtable}

In the past 7 days, I got enough sleep. [Enough sleep]
\begin{longtable}{
>{\centering\arraybackslash}p{\thin{}} >{\centering\arraybackslash}p{\thin{}} >{\centering\arraybackslash}p{\thin{}} >{\centering\arraybackslash}p{\thin{}} >{\centering\arraybackslash}p{\thin{}}
}
5. Never & 4. Rarely & 3. Sometimes & 2. Often & 1. Always \\
\st{} \\ 
\end{longtable}
    
In the past 7 days, my sleep quality was. [Quality]
\begin{longtable}{
>{\centering\arraybackslash}p{\thin{}} >{\centering\arraybackslash}p{\thin{}} >{\centering\arraybackslash}p{\thin{}} >{\centering\arraybackslash}p{\thin{}} >{\centering\arraybackslash}p{\thin{}}
}
5. Very poor & 4. Poor & 3. Fair & 2. Good & 1. Very good \\
\st{} \\ 
\end{longtable}

\subsubsection*{Sleep Impairment Survey}

In the past 7 days, I had a hard time getting things done because I was sleepy. [Trouble being productive]
\begin{longtable}{
>{\centering\arraybackslash}p{\thin{}} >{\centering\arraybackslash}p{\thin{}} >{\centering\arraybackslash}p{\thin{}} >{\centering\arraybackslash}p{\thin{}} >{\centering\arraybackslash}p{\thin{}}
}
1. Not at all & 2. A little bit & 3. Somewhat & 4. Quite a bit & 5. Very much \\
\st{} \\ 
\end{longtable}
    
In the past 7 days, I felt alert when I woke up. [Alert]
\begin{longtable}{
>{\centering\arraybackslash}p{\thin{}} >{\centering\arraybackslash}p{\thin{}} >{\centering\arraybackslash}p{\thin{}} >{\centering\arraybackslash}p{\thin{}} >{\centering\arraybackslash}p{\thin{}}
}
5. Not at all & 4. A little bit & 3. Somewhat & 2. Quite a bit & 1. Very much \\
\st{} \\ 
\end{longtable}
    
In the past 7 days, I felt tired. [Tiredness]
\begin{longtable}{
>{\centering\arraybackslash}p{\thin{}} >{\centering\arraybackslash}p{\thin{}} >{\centering\arraybackslash}p{\thin{}} >{\centering\arraybackslash}p{\thin{}} >{\centering\arraybackslash}p{\thin{}}
}
1. Not at all & 2. A little bit & 3. Somewhat & 4. Quite a bit & 5. Very much \\
\st{} \\ 
\end{longtable}
    
In the past 7 days, I had problems during the day because of poor sleep. [Having problems]
\begin{longtable}{
>{\centering\arraybackslash}p{\thin{}} >{\centering\arraybackslash}p{\thin{}} >{\centering\arraybackslash}p{\thin{}} >{\centering\arraybackslash}p{\thin{}} >{\centering\arraybackslash}p{\thin{}}
}
1. Not at all & 2. A little bit & 3. Somewhat & 4. Quite a bit & 5. Very much \\
\st{} \\ 
\end{longtable}
    
In the past 7 days, I had a hard time concentrating because of poor sleep. [Sleep impairment due to trouble concentrating]
\begin{longtable}{
>{\centering\arraybackslash}p{\thin{}} >{\centering\arraybackslash}p{\thin{}} >{\centering\arraybackslash}p{\thin{}} >{\centering\arraybackslash}p{\thin{}} >{\centering\arraybackslash}p{\thin{}}
}
1. Not at all & 2. A little bit & 3. Somewhat & 4. Quite a bit & 5. Very much \\
\st{} \\ 
\end{longtable}
    
In the past 7 days, I felt irritable because of poor sleep. [Sleep impairment due to irritability]
\begin{longtable}{
>{\centering\arraybackslash}p{\thin{}} >{\centering\arraybackslash}p{\thin{}} >{\centering\arraybackslash}p{\thin{}} >{\centering\arraybackslash}p{\thin{}} >{\centering\arraybackslash}p{\thin{}}
}
 1. Not at all & 2. A little bit & 3. Somewhat & 4. Quite a bit & 5. Very much \\
 \st{} \\ 
 \end{longtable}

In the past 7 days, I was sleepy during the daytime. [Sleepy during daytime]
\begin{longtable}{
>{\centering\arraybackslash}p{\thin{}} >{\centering\arraybackslash}p{\thin{}} >{\centering\arraybackslash}p{\thin{}} >{\centering\arraybackslash}p{\thin{}} >{\centering\arraybackslash}p{\thin{}}
}
1. Not at all & 2. A little bit & 3. Somewhat & 4. Quite a bit & 5. Very much \\
\st{} \\ 
\end{longtable}
    
In the past 7 days, I had trouble staying awake during the day. [Trouble staying awake]
\begin{longtable}{
>{\centering\arraybackslash}p{\thin{}} >{\centering\arraybackslash}p{\thin{}} >{\centering\arraybackslash}p{\thin{}} >{\centering\arraybackslash}p{\thin{}} >{\centering\arraybackslash}p{\thin{}}
}
1. Not at all & 2. A little bit & 3. Somewhat & 4. Quite a bit & 5. Very much \\
\st{} \\ 
\end{longtable}
\newpage

\begin{figure}[thp]
  \centering
    \includegraphics[width=0.8\columnwidth]{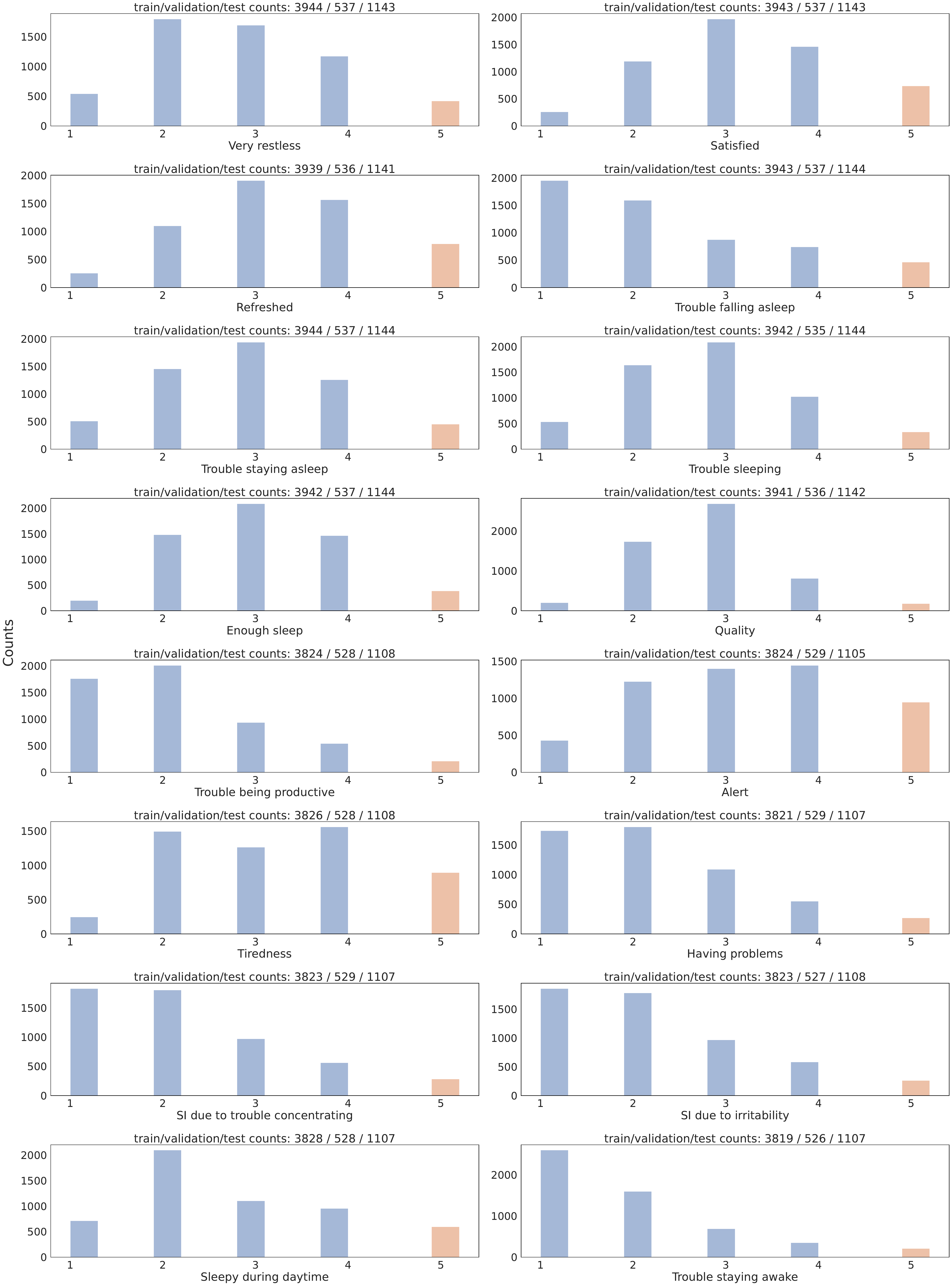}
    \caption{\textbf{Distribution of responses for each survey question.} Survey questions are answered on a Likert scale. Here we show the distribution of responses for each question. The bar highlighted in a darker tone indicates those answers that were labeled as positive cases in the defined binary traits. The training, validation and test set counts are included in the title of each subplot.}
    \label{fig:supp:survey_distributions}
\end{figure}

\newpage
\subsection{Patient-reported outcome prediction performance.}

\begin{figure}[h!]
     \centering
     \begin{subfigure}[b]{0.45\textwidth}
        \centering
        \includegraphics[width=\textwidth]{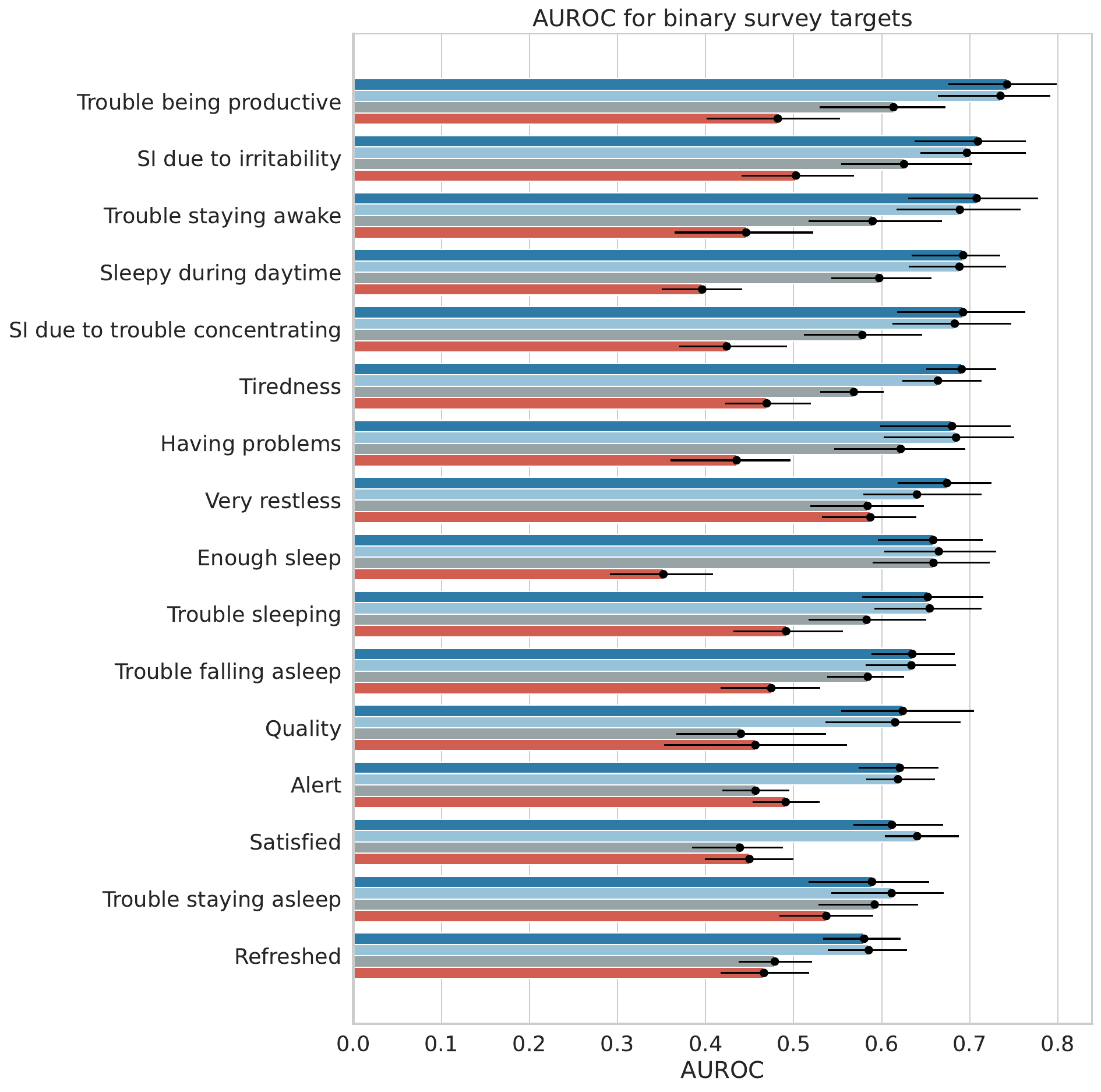}
        \caption{Performance of \ourmodel measured via AUROC.}
        \label{fig:supp:pro_prediction_auroc}
     \end{subfigure}
     \hfill
     \begin{subfigure}[b]{0.45\textwidth}
        \centering
        \includegraphics[width=\textwidth]{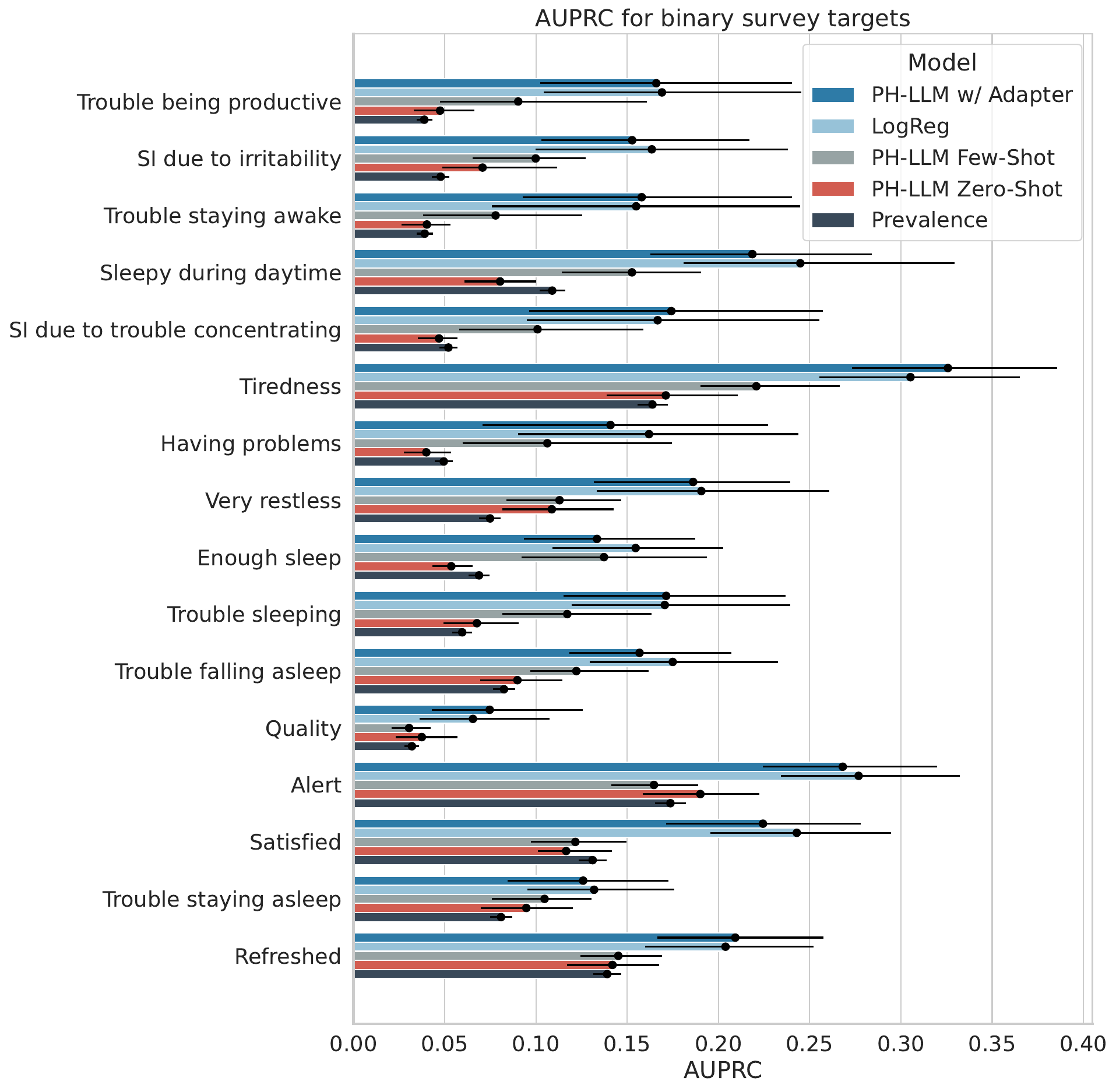}
        \caption{Performance of \ourmodel measured via AUPRC.}
        \label{fig:supp:pro_prediction_auprc}
     \end{subfigure}
    \caption{\textbf{Prediction of Patient-Reported Outcomes.} We evaluated the ability for the \ourmodel to infer subjective patient-reported outcomes using a multimodal adapter and compare to a suite of logistic regression models trained to predict each task independently, as well as \ourmodel using zero- and few-shot text prompting.}
    \hfill
\end{figure}

\newpage
\begin{table}[h]
\centering
\caption{\textbf{Differences in AUROC between \ourmodel using a multimodal adapter and other modeling approaches.} Here, we highlight values in bold where the difference between \ourmodel with adapter and the other approach were \textbf{not} statistically significant. Statistical significance was determined via a paired bootstrap estimator.}
\label{table:supp:pro_roauc_diff} 
\begin{tabular}{llll}
\toprule[1.5pt] 
 & \ourmodel zero-shot & \ourmodel few-shot & Logistic Regression \\
\midrule
Very restless & 0.087 & 0.090 & \textbf{0.034} \\
Satisfied & 0.162 & 0.173 & \textbf{-0.029} \\
Refreshed & 0.114 & 0.101 & \textbf{-0.005} \\
Trouble falling asleep & 0.160 & \textbf{0.051} & \textbf{0.001} \\
Trouble staying asleep & \textbf{0.052} & \textbf{-0.003} & \textbf{-0.022} \\
Trouble sleeping & 0.161 & 0.069 & \textbf{-0.002} \\
Enough sleep & 0.306 & \textbf{-0.000} & \textbf{-0.006} \\
Quality & 0.167 & 0.184 & \textbf{0.009} \\
Trouble being productive & 0.260 & 0.129 & \textbf{0.007} \\
Alert & 0.130 & 0.164 & \textbf{0.002} \\
Tiredness & 0.221 & 0.122 & \textbf{0.027} \\
Having problems & 0.245 & \textbf{0.058} & \textbf{-0.005} \\
SI due to trouble concentrating & 0.268 & 0.114 & \textbf{0.009} \\
SI due to irritability & 0.207 & 0.084 & \textbf{0.012} \\
Sleepy during daytime & 0.296 & 0.095 & \textbf{0.004} \\
Trouble staying awake & 0.262 & 0.118 & \textbf{0.019} \\
\bottomrule[1.5pt]
\end{tabular}
\end{table}

\begin{table}[ht]
\centering
\caption{\textbf{Differences in AUPRC between \ourmodel using a multimodal adapter and other modeling approaches.} Here, we highlight values in bold where the difference between \ourmodel with adapter and the other approach were \textbf{not} statistically significant. Statistical significance was determined via a paired bootstrap estimator.}
\label{table:supp:pro_roauprc_diff} 
\begin{tabular}{llll}
\toprule[1.5pt] 
 & \ourmodel zero-shot & \ourmodel few-shot & Logistic Regression \\
\midrule
Very restless & 0.077 & 0.073 & \textbf{-0.004} \\
Satisfied & 0.108 & 0.103 & \textbf{-0.019} \\
Refreshed & 0.067 & 0.064 & \textbf{0.005} \\
Trouble falling asleep & 0.067 & \textbf{0.035} & \textbf{-0.018} \\
Trouble staying asleep & \textbf{0.031} & \textbf{0.021} & \textbf{-0.006} \\
Trouble sleeping & 0.104 & 0.054 & \textbf{0.001} \\
Enough sleep & 0.080 & \textbf{-0.004} & \textbf{-0.021} \\
Quality & 0.037 & 0.044 & \textbf{0.009} \\
Trouble being productive & 0.118 & 0.076 & \textbf{-0.003} \\
Alert & 0.078 & 0.103 & \textbf{-0.009} \\
Tiredness & 0.155 & 0.105 & \textbf{0.021} \\
Having problems & 0.101 & 0.035 & \textbf{-0.021} \\
SI due to trouble concentrating & 0.127 & 0.073 & \textbf{0.008} \\
SI due to irritability & 0.082 & 0.053 & \textbf{-0.011} \\
Sleepy during daytime & 0.138 & 0.066 & \textbf{-0.026} \\
Trouble staying awake & 0.118 & 0.080 & \textbf{0.003} \\
\bottomrule[1.5pt]
\end{tabular}
\end{table}

\newpage
\subsection{Patient-reported outcome prompt examples}

\begin{table}[ht]
\captionsetup{labelformat=simple, justification=raggedright, singlelinecheck=false, labelsep=pipe} 
\caption{\textbf{Example of prompt given to \ourmodel to score PROs.} Demographic and sensor values are passed as text to the model. The feature to predict (in this example, \texttt{very restless}) can then be scored using the potential completions ``yes.'' or ``no.'' For few-shot prompting we additionally prepend complete examples from the training set to the prompt. When using the multimodal adapter (see Methods), a vector representation of the quantitative data is passed in via a set of learned tokens. Values in the below prompt are synthetic.
}
\label{table:prompt_pro} 
\end{table}
\begin{tcolorbox}[title={Example of prompt given to \ourmodel to score PROs.}]
\color{black} 
Use the information provided to predict ``very restless''.
\\
age: [40-45].
heart rate variability root mean square of successive differences: 33.5.
respiratory rate breaths per minute: 16.5.
resting heart rate beats per minute: 60.0.
awake minutes: 51.0.
deep sleep minutes: 53.0.
sleep duration minutes: 471610.0.
efficiency: 0.85.
overall sleep score: 81.0.
percent of sleep in REM: 16.0.
restlessness: 0.07.
revitalization score: 83.2.
sleep end time: -274.0.
sleep start time: 364.0.
sleep time minutes: 420.8.
total number of minutes in awake state after sleep onset: 7.4.
number of steps walked during the day: 6850.0.
number of minutes spent in cardio zone during the day: 6.7.
number of minutes spent in fat burn zone during the day: 18.9.
number of minutes spent in peak zone during the day: 0.41.
total multiplied minutes of exercise during the day: 45.32.
\\
very restless: yes or no?
\end{tcolorbox}

\end{document}